\documentclass[runningheads]{llncs}
\usepackage{graphicx}
\usepackage{amsmath,amssymb}
\usepackage{color}

\usepackage{rotating}

\usepackage[utf8]{inputenc}
\usepackage[T1]{fontenc}
\usepackage{epsfig}
\usepackage{graphicx}
\usepackage{amsmath}
\usepackage{amssymb}
\usepackage{comment}

  \newcommand{\R}{\mathbb{R}}

  \usepackage{bm}
  
  \newcommand{\mbf}[1]{\mathbf{#1}}
  \newcommand{\vect}[1]{\mbf{#1}}

  \usepackage{xcolor,colortbl}
  
  \definecolor{DarkGray}{gray}{0.85}
  \definecolor{Gray}{gray}{0.90}

  \definecolor{mycolor1}{rgb}{0.00000,0.44700,0.74100}%
  \definecolor{mycolor2}{rgb}{0.85000,0.32500,0.09800}%
  \definecolor{mycolor3}{rgb}{0.92900,0.69400,0.12500}%
  \definecolor{mycolor4}{rgb}{0.49400,0.18400,0.55600}%
  \definecolor{mycolor5}{rgb}{0.46600,0.67400,0.18800}%
  \definecolor{mycolor6}{rgb}{0.30100,0.74500,0.93300}%

  \newcommand{\eg}{\textit{e.g.}}
  \newcommand{\ie}{\textit{i.e.}}

  \usepackage{booktabs}
  \usepackage{multirow}
  \usepackage{tabularx}
  \usepackage{longtable}

  \makeatletter
  \let\MYcaption\@makecaption
  \makeatother

  \usepackage{subcaption}

  \makeatletter
  \let\@makecaption\MYcaption
  \makeatother

  \usepackage{tikz,pgfplots}
  \usetikzlibrary{plotmarks,shapes,arrows}
  \pgfplotsset{compat=newest} 
  \pgfkeys{/pgf/number format/.cd,1000 sep={}}

  \newlength\figureheight
  \newlength\figurewidth

\usepackage[pagebackref=true,breaklinks=true,letterpaper=true,colorlinks,bookmarks=true,hidelinks]{hyperref}

\def\middot{\textperiodcentered~}

\renewcommand{\orcidID}[1]{}

\begin{document}
\pagestyle{headings}
\mainmatter
\def\ECCV18SubNumber{2131}

\title{ADVIO: An Authentic Dataset for \\ Visual-Inertial Odometry}

\titlerunning{ADVIO: An authentic dataset for \\ visual-inertial odometry}
\authorrunning{Cort\'es, Solin, Rahtu, and Kannala}

\author{Santiago Cort\'es\inst{1}\orcidID{0000-0001-7886-7841} \and 
        Arno Solin\inst{1}\orcidID{0000-0002-0958-7886} \and 
        Esa Rahtu\inst{2}\orcidID{0000-0001-8767-0864} \and 
        Juho Kannala\inst{1}\orcidID{0000-0001-5088-4041}}
\institute{Department of Computer Science, Aalto University, Espoo, Finland \\
  \email{\{santiago.cortesreina,arno.solin,juho.kannala\}@aalto.fi}
  \and Tampere University of Technology, Tampere, Finland \\ 
  \email{esa.rahtu@tut.fi}}

\maketitle

\begin{abstract}
  The lack of realistic and open benchmarking datasets for pedestrian visual-inertial odometry has made it hard to pinpoint differences in published methods. Existing datasets either lack a full six degree-of-freedom ground-truth or are limited to small spaces with optical tracking systems. We take advantage of advances in pure inertial navigation, and develop a set of versatile and challenging real-world computer vision benchmark sets for visual-inertial odometry. For this purpose, we have built a test rig equipped with an iPhone, a Google Pixel Android phone, and a Google Tango device. We provide a wide range of raw sensor data that is accessible on almost any modern-day smartphone together with a high-quality ground-truth track. We also compare resulting visual-inertial tracks from Google Tango, ARCore, and Apple ARKit with two recent methods published in academic forums. The data sets cover both indoor and outdoor cases, with stairs, escalators, elevators, office environments, a shopping mall, and metro station.
\end{abstract}

\noindent
\begin{tikzpicture}
  \footnotesize
  \draw [DarkGray,fill=Gray, line width=1pt,rounded corners=2mm]
    (0,0) rectangle
    (\columnwidth,1.2);
  \node at (.5\columnwidth,.6) {
    \parbox{.9\columnwidth}{\raggedright 
      \bf Access data and documentation at: \\
      \url{https://github.com/AaltoVision/ADVIO}
    }
  };
\end{tikzpicture}

\section{Introduction}
Various systems and approaches have recently emerged for tracking the motion of hand-held or wearable mobile devices based on video cameras and inertial measurement units (IMUs). There exist both open published methods (\eg\ \cite{Mourikis+Roumeliotis:2007,Mur-Artal+Tardos:2017,Blosch+Omari+Hutter+Siegwart:2015,Li+Hyung+Mourikis,Schops+Engel+Cremers})
and closed proprietary systems. Recent examples of the latter are ARCore by Google and ARKit by Apple which run on the respective manufacturers' flagship smartphone models. Other examples of mobile devices with built-in visual-inertial odometry are the Google Tango tablet device and Microsoft Hololens augmented reality glasses. The main motivation for developing odometry methods for smart mobile devices is to enable augmented reality applications which require precise real-time tracking of ego-motion. Such applications could have significant value in many areas, like architecture and design, games and entertainment, telepresence, and education and training.

Despite the notable scientific and commercial interest towards visual-inertial odometry, the progress of the field is constrained by the lack of public datasets and benchmarks which would allow fair comparison of proposed solutions and facilitate further developments to push the current boundaries of the state-of-the-art systems. For example, since the performance of each system depends on both the algorithms and sensors used, it is hard to compare methodological advances and algorithmic contributions fairly as the contributing factors from hardware and software may be mixed. In addition, as many existing datasets are either captured in small spaces or utilise significantly better sensor hardware than feasible for low-cost consumer devices, it is difficult to evaluate how the current solutions would scale to medium or long-range odometry, or large-scale simultaneous localization and mapping (SLAM), on smartphones.
 
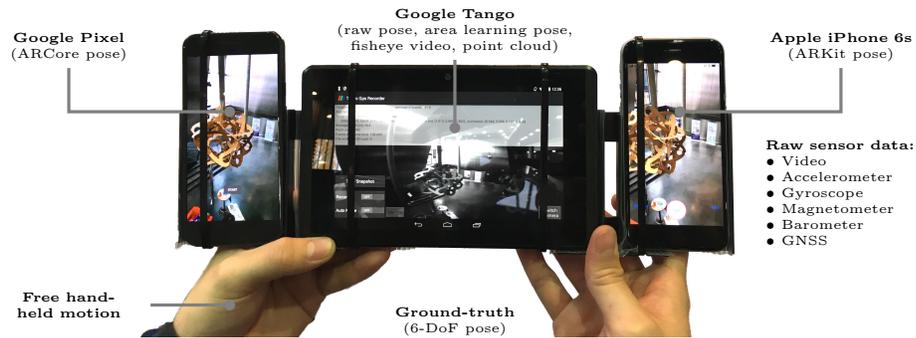
\begin{figure}[!t]
  \centering\fontsize{5}{6}\selectfont
  \begin{tikzpicture}
    \node[anchor=south west,inner sep=0] (image) at (0,0) %
      {\includegraphics[width=.7\columnwidth,trim=0 0 0 2.5cm,clip]{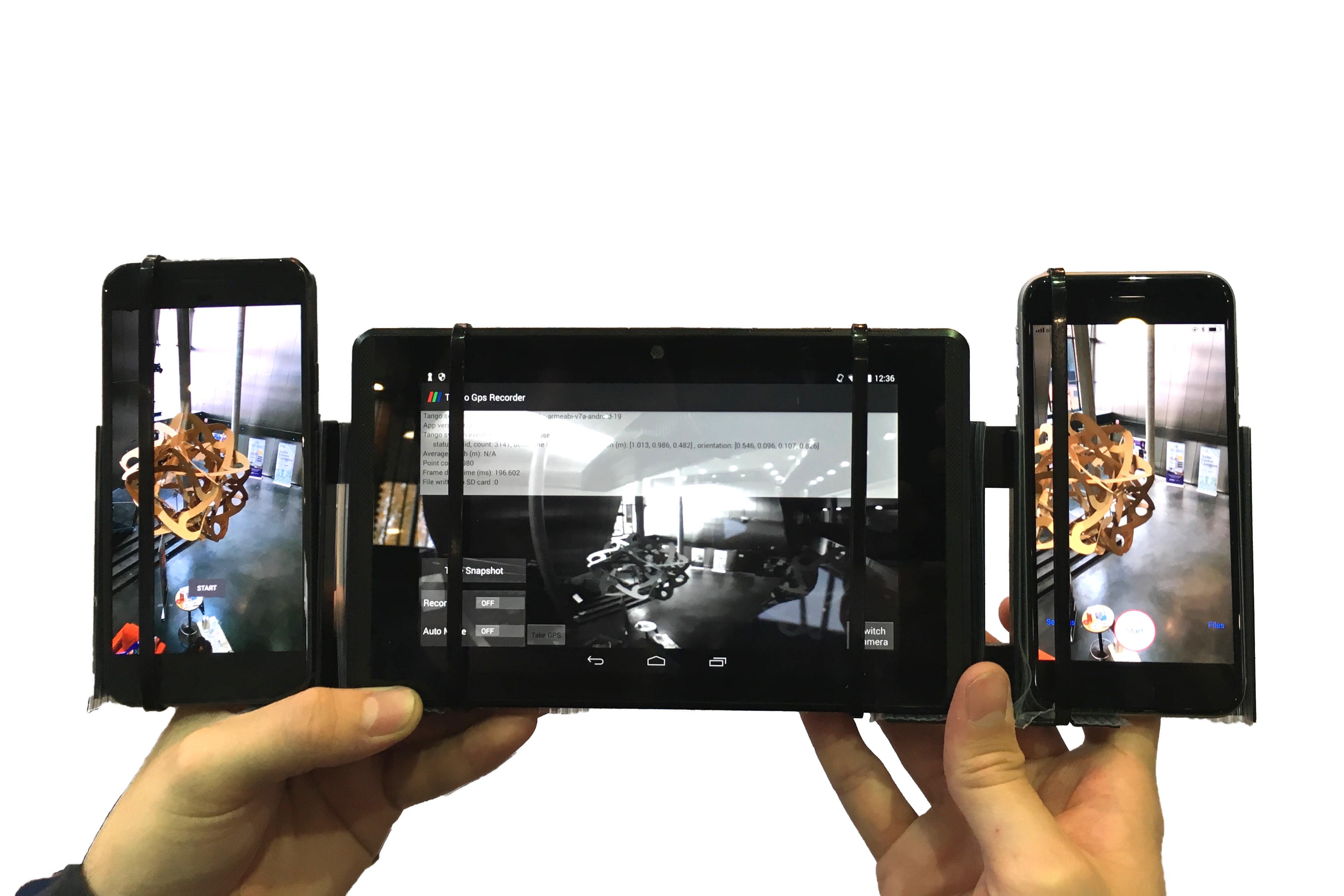}};
    \begin{scope}[x={(image.south east)},y={(image.north west)}]

      \tikzstyle{label} = [text width=2cm, align=center]
      \tikzstyle{line}  = [draw, very thick, black!50]
      \tikzstyle{point} = [circle,draw,very thick,black!50,fill=black!50,
                           minimum size=1mm,inner sep=0]

      \node [label] (lab_1) at (-0.10,0.90) {{\bf Google Pixel} \\ (ARCore pose)};
      \node [label, text width=5cm] (lab_2) at ( 0.50,0.95) {{\bf Google Tango} \\ (raw pose, area learning pose, fisheye video, point cloud)};
      \node [label] (lab_3) at ( 1.10,0.90) {{\bf Apple iPhone 6s} \\ (ARKit pose)};
      \node [label] (lab_4) at (-0.10,0.10) {{\bf Free hand-held motion}};      
      \node [label,text width=5cm] (lab_5) at ( 0.50,0.05) {{\bf Ground-truth} \\ (6-DoF pose)};
      \node [label, align=left] (lab_6) at ( 1.10,0.45) {{\bf Raw sensor data:} \\ $\bullet$~Video \\ $\bullet$~Accelerometer \\ $\bullet$~Gyroscope \\ $\bullet$~Magnetometer \\ $\bullet$~Barometer \\ $\bullet$~GNSS};

      \node [point] (p1) at (0.15,0.70) {};
      \node [point] (p2) at (0.50,0.65) {};
      \node [point] (p3) at (0.85,0.70) {};
      \node [point] (p4) at (0.15,0.10) {};

      \path [line] (lab_1) |- (p1);
      \path [line] (lab_2) |- (p2);
      \path [line] (lab_3) |- (p3);
      \path [line] (lab_4) |- (p4);

    \end{scope}
  \end{tikzpicture}\\[-1em]
  \caption{The custom-built capture rig with a Google Pixel smartphone on the left, a Google Tango device in the middle, and an Apple iPhone 6s on the right.}
  \label{fig:rig}
\end{figure}

Further, the availability of realistic sensor data, captured with smartphone sensors, together with sufficiently accurate ground-truth would be beneficial in order to speed up progress in academic research and also lower the threshold for new researchers entering the field. The importance of public datasets and benchmarks as a driving force for rapid progress has been clearly demonstrated in many computer vision problems, like image classification \cite{Everingham+Eslami+Gool+Williams+Winn+Zisserman:2015,Russakovsky+Deng+Su+Krause+Satheesh+Ma+Huang+Karpathy+Khosla+Bernstein+Berg+Fei-Fei:2015}, object detection \cite{Lin+Maire+Belongie+Hays+Perona+Ramanan+Dollar+Zitnick:2014}, stereo reconstruction \cite{Geiger+Lenz+Urtasun:2012} and semantic segmentation \cite{Lin+Maire+Belongie+Hays+Perona+Ramanan+Dollar+Zitnick:2014,Cordts+Omran+Ramos+Rehfeld+Enzweiler+Benenson+Franke+Roth+Schiele:2016}, to name a few. However, regarding visual-inertial odometry, there are no publicly available datasets or benchmarks that would allow evaluating recent methods in a typical smartphone context. Moreover, since the open-source software culture is not as common in this research area as, for example, it is in image classification and object detection, the research environment is not optimal for facilitating rapid progress. Further, due to the aforementioned reasons, there is a danger that the field could become accessible only for big research groups funded by large corporations, and that would slow down progress and decay open academic research.

\begin{figure*}[!t]
  \centering
  \hspace*{\fill}
  \begin{subfigure}[b]{0.25\textwidth}
    \centering
    \includegraphics[width=.95\textwidth,trim=0 0 0 10mm,clip]{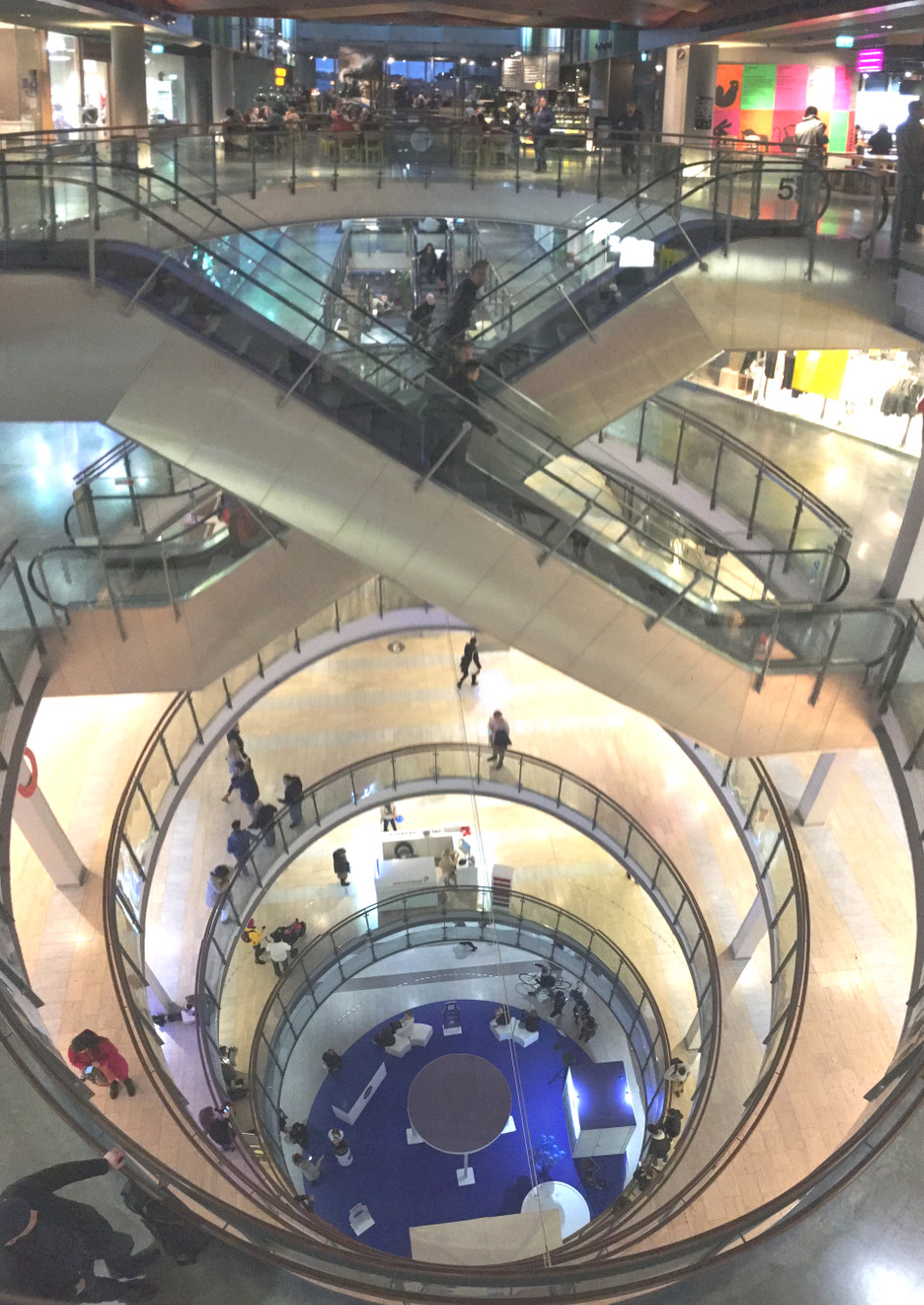}
    \caption{View inside mall}
    \label{fig:escalator}
  \end{subfigure}
  \hspace*{\fill}
  \begin{subfigure}[b]{0.36
  \textwidth}
    \centering
    \includegraphics[width=\textwidth]{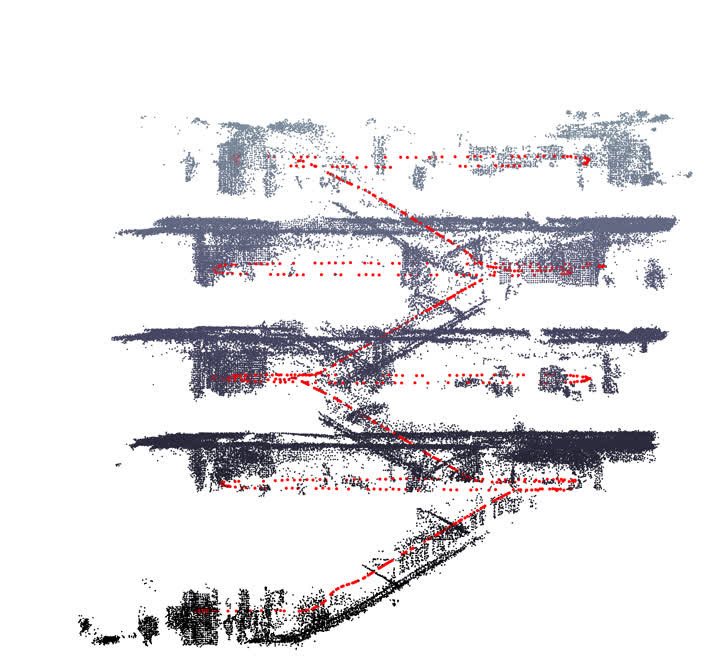}
    \caption{Tango point cloud}
    \label{fig:points}
  \end{subfigure}
  \hspace*{\fill}
  \begin{subfigure}[b]{0.26\textwidth}
    \centering
    \setlength{\figurewidth}{.95\textwidth}
    \pgfplotsset{clip=false}
    \input{./fig/floors.tex}
    \caption{Escalator data sets}
    \label{fig:floors}
  \end{subfigure}
  \hspace*{\fill}
  \\[-1em]
  \caption{Multi-floor environments such as (a) were considered. The point cloud (b) and escalator/elevator paths captured in the mall. The Tango track (red) in (b) has similar shape as the ground-truth in (c). Periodic locomotion can be seen in (c) if zoomed in.}
  \label{fig:mall}
\end{figure*}

In this work, we present a dataset that aims to facilitate the development of visual-inertial odometry and SLAM methods for smartphones and other mobile devices with low-cost sensors (\ie\ rolling-shutter cameras and MEMS based inertial sensors). Our sensor data is collected using a standard iPhone~6s device and contains the ground-truth pose trajectory and the raw synchronized data streams from the following sensors: RGB video camera, accelerometer, gyroscope, magnetometer, platform-provided geographic coordinates, and barometer. In total, the collected sequences contain about 4.5~kilometres of unconstrained hand-held movement in various environments both indoors and outdoors. One example sequence is illustrated in Figure~\ref{fig:mall}. The data sets are collected in public spaces, conforming the local legislation regarding filming and publishing.
The ground-truth is computed by combining a recent pure inertial navigation system (INS) \cite{Solin+Cortes+Rahtu+Kannala} with frequent manually determined position fixes based on a precise floor plan. The quality of our ground-truth is verified and its accuracy estimated.

Besides the benchmark dataset, we present a comparison of visual-inertial odometry methods, including three recent proprietary platforms: ARCore on a Google Pixel device, Apple ARKit on the iPhone, and Tango odometry on a Google Tango tablet device, and two recently published methods, namely ROVIO~\cite{Blosch+Omari+Hutter+Siegwart:2015,Bloesch+Burri+Omari+Hutter+Siegwart:2017} and PIVO~\cite{Solin+Cortes+Rahtu+Kannala:2017}. The data for the comparison was collected with a capture rig with the three devices and is illustrated in Figure~\ref{fig:rig}.
Custom applications for data capture were implemented for each device.

The main contributions of our work are summarized in the following:
\begin{itemize}
  \item A public dataset of iPhone sensor data with 6 degree-of-freedom pose ground-truth for benchmarking monocular visual-inertial odometry in real-life use cases involving motion in varying environments, and also including stairs, elevators and escalators. 
  \item Comparing state-of-the-art visual-inertial odometry platforms and methods.
  \item A method for collecting ground-truth for smartphone odometry in realistic use cases by combining pure inertial navigation with manual position fixes. 
\end{itemize}

\begin{table*}[!t]
  \caption{An overview of related datasets.}
  \label{tbl:related}
  \centering\fontsize{6}{7}\selectfont
  \setlength{\aboverulesep}{0pt}
  \setlength{\belowrulesep}{0pt}
  \setlength{\extrarowheight}{.5ex}
  \resizebox{\textwidth}{1.6cm}{%
  \begin{tabularx}{1.4\textwidth}{ l @{\extracolsep{\fill}} c c c c c >{\columncolor{Gray}}c } 
  \toprule
    & 
    {\bf Rawseeds} \cite{Ceriani+Fontana+Giusti+Marzorati+Matteucci+Migliore+Rizzi+Sorrenti+Taddei:2009} & 
    {\bf KITTI} \cite{Geiger+Lenz+Urtasun:2012} & 
    {\bf NCLT} \cite{Carlevaris-Bianco+Ushani+:2015} & 
    {\bf EuRoC} \cite{Burri+Nikolic+Gohl+Schneider+Rehder+Omari+Achtelik+Siegwart:2016} & 
    {\bf PennCOSYVIO} \cite{Pfrommer+Sanket+Daniilidis+Cleveland:2017} & 
    {\bf Proposed} \\
  \midrule
  Year & 2006 & 2012 & 2015 & 2016 & 2017 & (this paper) \\
  Carrier   & Wheeled robot & Car  & Segway & MAV & Hand-held device & Hand-held device \\
  Environment & Indoors/Outdoors & Outdoors & Indoors/Outdoors & Indoors & Indoors/Outdoors & Indoors/Outdoors \\
  Scene setup & Campus-scale & City-scale & Campus-scale & 2 Rooms & 150 m path on campus & Multiple levels in 3 buildings \\
        & & & & &  (walked 4 times) & + outdoor scenes \\

  Hardware setup    & Custom & Custom & Custom & Custom & Custom & Standard smartphone \\
  Distance (total)   & $\sim$10 km & $\sim$39 km & $\sim$147 km & $\sim$800 m & $\sim$600 m & $\sim$4.5 km \\

  Long-range use-case  &  \checkmark & \checkmark & \checkmark & --- &  \checkmark & \checkmark \\
  3D point cloud & --- & \checkmark & \checkmark & \checkmark & --- & \checkmark \\

  \midrule
  Ground-truth & GPS/Visual tags & GPS/IMU & GPS/IMU/Laser & MoCap/Laser & Visual tags & IMU + Position fixes \\
  Accuracy & $\sim$m & $\sim$dm & $\sim$dm & $\sim$mm & $\sim$dm & dm--m \\
  \bottomrule
  \end{tabularx}}
\end{table*}

\section{Related Work}
Despite visual-inertial odometry (VIO) being one of the most promising approaches for real-time tracking of hand-held and wearable devices, there is a lack of good public datasets for benchmarking different methods. A relevant benchmark should include both video and inertial sensor recordings with synchronized time stamps preferably captured with consumer-grade smartphone sensors. In addition, the dataset should be authentic and illustrate realistic use cases. That is, it should contain challenging environments with scarce visual features, both indoors and outdoors, and varying motions, also including rapid rotations without translation as they are problematic for monocular visual-only odometry. Our work is the first one addressing this need.

Regarding pure visual odometry or SLAM, there are several datasets and benchmarks available \cite{Smith+Baldwin+Churchill+Paul+Newman:2009,Cordts+Omran+Ramos+Rehfeld+Enzweiler+Benenson+Franke+Roth+Schiele:2016,Engel+Usenko+Cremers:2016,Sturm+Engelhard+Endres+Burgard+Cremers:2012} but they lack the inertial sensor data. Further, many of these datasets are limited because they \emph{(a)} are recorded using ground vehicles and hence do not have rapid rotations \cite{Smith+Baldwin+Churchill+Paul+Newman:2009,Cordts+Omran+Ramos+Rehfeld+Enzweiler+Benenson+Franke+Roth+Schiele:2016}, \emph{(b)} do not contain low-textured indoor scenes \cite{Smith+Baldwin+Churchill+Paul+Newman:2009,Cordts+Omran+Ramos+Rehfeld+Enzweiler+Benenson+Franke+Roth+Schiele:2016}, \emph{(c)} are captured with custom hardware (\eg\ fisheye lens or global shutter camera) \cite{Engel+Usenko+Cremers:2016}, \emph{(d)} lack full 6-degree of freedom ground-truth \cite{Engel+Usenko+Cremers:2016}, or \emph{(e)} are constrained to small environments and hence are ideal for SLAM systems but not suitable for benchmarking odometry for medium and long-range navigation \cite{Sturm+Engelhard+Endres+Burgard+Cremers:2012}. 

Nevertheless, besides pure vision datasets, there are some public datasets with inertial sensor data included, for example, \cite{Geiger+Lenz+Urtasun:2012,Ceriani+Fontana+Giusti+Marzorati+Matteucci+Migliore+Rizzi+Sorrenti+Taddei:2009,Carlevaris-Bianco+Ushani+:2015,Burri+Nikolic+Gohl+Schneider+Rehder+Omari+Achtelik+Siegwart:2016,Pfrommer+Sanket+Daniilidis+Cleveland:2017}. Most of these datasets are recorded with sensors rigidly attached to a wheeled ground vehicle. For example, the widely used KITTI dataset \cite{Geiger+Lenz+Urtasun:2012} contains LIDAR scans and videos from multiple cameras recorded from a moving car. The ground-truth is obtained using a very accurate GPS/IMU localization unit with RTK correction signals. However, the IMU data is captured only with a frequency of 10~Hz, which would not be sufficient for tracking rapidly moving hand-held devices. Further, even if high-frequency IMU data would be available, also KITTI has the constraints \emph{(a)}, \emph{(b)}, and \emph{(c)} mentioned above and this limits its usefulness for smartphone odometry.

Another analogue to KITTI is that we also use pure inertial navigation with external location fixes for determining the ground-truth. In our case, the GPS fixes are replaced with manual location fixes since GPS is not available or accurate indoors. Further, in contrast to KITTI, by utilizing recent advances in inertial navigation \cite{Solin+Cortes+Rahtu+Kannala} we are able to use the inertial sensors of the iPhone for the ground-truth calculation and are therefore not dependent on a high-grade IMU, which would be difficult to attach to the hand-held rig. In our case the manual location fixes are determined from a reference video (Fig.~\ref{fig:ref-video}), which views the recorder, by visually identifying landmarks that can be accurately localized from precise building floor plans or aerial images. The benefit of not using optical methods for establishing the ground-truth is that we can easily record long sequences and the camera of the recording device can be temporarily occluded. This makes our benchmark suitable also for evaluating occlusion robustness of VIO methods \cite{Solin+Cortes+Rahtu+Kannala:2017}.
Like KITTI, the Rawseeds \cite{Ceriani+Fontana+Giusti+Marzorati+Matteucci+Migliore+Rizzi+Sorrenti+Taddei:2009} and NCLT \cite{Carlevaris-Bianco+Ushani+:2015} datasets are recorded with a wheeled ground vehicle. Both of them use custom sensors (\eg\ omnidirectional camera or industrial-grade IMU). These datasets are for evaluating odometry and self-localization of slowly moving vehicles and not suitable for benchmarking VIO methods for hand-held devices and augmented reality.

The datasets that are most related to ours are EuRoC \cite{Burri+Nikolic+Gohl+Schneider+Rehder+Omari+Achtelik+Siegwart:2016} and PennCOSYVIO \cite{Pfrommer+Sanket+Daniilidis+Cleveland:2017}. EuRoC provides visual and inertial data captured with a global shutter stereo camera and a tactical-grade IMU onboard a micro aerial vehicle (MAV) \cite{Nikolic+Rehder+Burri+Gohl+Leutenegger+Furgale+Siegwart:2014}. The sequences are recorded in two different rooms that are equipped with motion capture system or laser tracker for obtaining accurate ground-truth motion.
In PennCOSYVIO, the data acquisition is performed using a hand-held rig containing two Google Tango tablets, three GoPro Hero~4 cameras, and a similar visual-inertial sensor unit as used in EuRoC. The data is collected by walking a 150 meter path several times at UPenn campus, and the ground-truth is obtained via optical markers.
Due to the need of optic localization for determining ground-truth, both EuRoC and PennCOSYVIO contain data only from a few environments that are all relatively small-scale. Moreover, both datasets use the same high-quality custom sensor with wide field-of-view stereo cameras \cite{Nikolic+Rehder+Burri+Gohl+Leutenegger+Furgale+Siegwart:2014}. In contrast, our dataset contains around 4.5 kilometers of sequences recorded with regular smartphone sensors in multiple floors in several different buildings and different outdoor environments. In addition, our dataset contains motion in stairs, elevators and escalators, as illustrated in Figure~\ref{fig:mall}, and also temporary occlusions and lack of visual features. We are not aware of any similar public dataset. The properties of different datasets are summarized in Table~\ref{tbl:related}.
The enabling factor for our flexible data collection procedure is to utilize recent advances in pure inertial navigation together with manual location fixes \cite{Solin+Cortes+Rahtu+Kannala}. In fact, the methodology for determining the ground-truth is one of the contributions of our work. In addition, as a third contribution, we present a comparison of recent VIO methods and proprietary state-of-the-art platforms based on our challenging dataset.

\section{Materials}

\begin{figure}[!t]
  \begin{subfigure}{.21\textwidth}
    \centering
    \includegraphics[height=4.6cm]{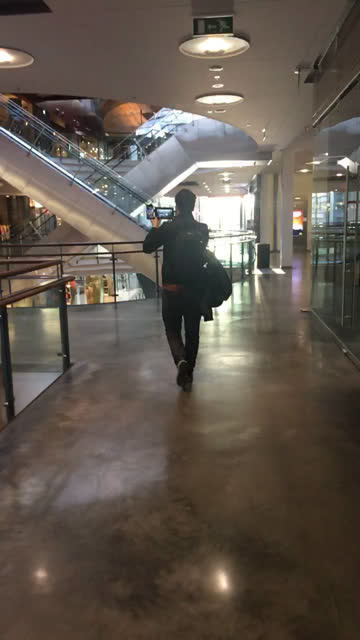}
    \caption{Reference}
    \label{fig:ref-video}
  \end{subfigure}
  \hspace*{\fill}
  \begin{subfigure}{.53\textwidth}
    \centering  
    \includegraphics[height=4.6cm]{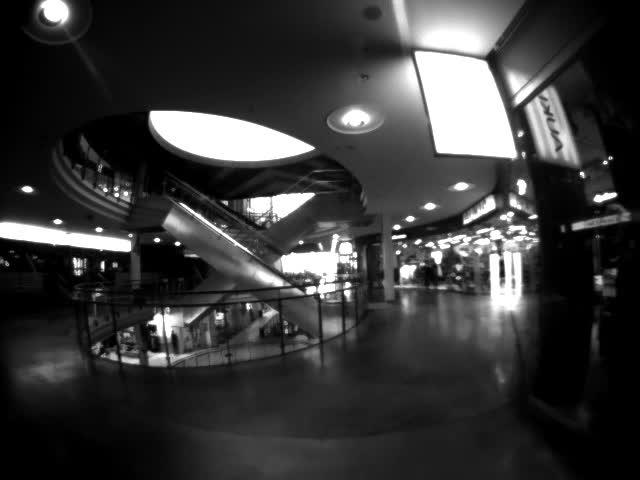}
    \caption{Tango (fisheye lens)}
    \label{fig:tango-video}
  \end{subfigure}
  \hspace*{\fill}
  \begin{subfigure}{.21\textwidth}
    \centering  
    \includegraphics[height=4.6cm]{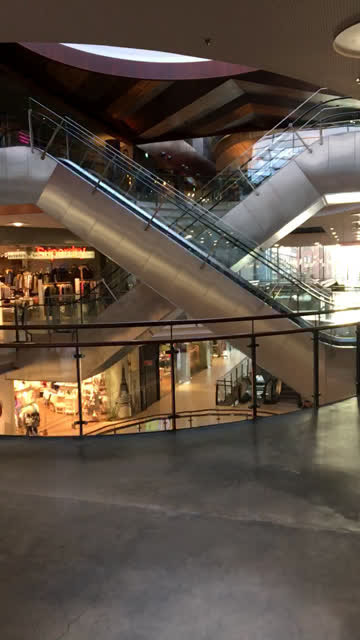}
    \caption{iPhone}
    \label{fig:iphone-video}
  \end{subfigure}\\[-1em]
  \caption{Example of simultaneously captured frames from three synchronized cameras. The external reference camera (a) is used for manual position fixes for determining the ground-truth trajectory in a separate post-processing stage.}
  \label{fig:frames}

\end{figure}

The data was recorded with the three devices (iPhone~6s, Pixel, Tango) rigidly attached to an aluminium rig (Fig.~\ref{fig:rig}). In addition, we captured the collection process with an external video camera that was viewing the recorder (Fig.~\ref{fig:frames}). The manual position fixes with respect to a 2D map (\ie\ a structural floor plan image or an aerial image/map) were determined afterwards from the view of the external camera. Since the device was hand-held, in most fix locations the height was given as a constant distance above the floor level (with a reasonable uncertainty estimate), so that the optimization could fit a trajectory that optimally balances the information from fix positions and IMU signals (details in Sec.~\ref{sec:methods}).

The data streams from all the four devices are synchronized using network provided time. That is, the device clock is synchronized over a network time protocol (NTP) request at the beginning of a capture session. All devices were connected to 4G network during recording. Further, in order to enable analysis of the data in the same coordinate frame, we calibrated the internal and external parameters of all cameras by capturing multiple views of a checkerboard. This was performed before each session to account for small movements during transport and storage.
The recorded data streams are listed in Table~\ref{tbl:stream}.

\subsection{Raw iPhone Sensor Capture}
An iOS data collection app was developed in Swift~4. It saves inertial and visual data synchronized to the Apple ARKit pose estimation. All individual data points are time stamped internally and then synchronized to global time. The global time is fetched using the Kronos Swift NTP client\footnote{https://github.com/lyft/Kronos}. The data was captured using an iPhone~6s running iOS~11.0.3. The same software and an identical iPhone was used for collecting the reference video. This model was chosen, because the iPhone~6s (published 2015) is hardware-wise closer to an average smartphone than most recent flagship iPhones and also matches well with the Google Pixel hardware.

During the capture the camera is controlled by the ARKit service. It is performing the usual auto exposure and white balance but the focal length is kept fixed (the camera matrix returned by ARKit is stored during capture). The resolution is also controlled by ARKit and it is 1280$\times$720. The frames are packed into an H.264/MPEG-4 video file.
The GNSS/network location data is collected through the CoreLocation API. Locations are requested with the desired accuracy of `kCLLocationAccuracyBest'. The location service provides latitude and longitude, horizontal accuracy, altitude, vertical accuracy, and speed.
The accelerometer, gyroscope, magnetometer, and barometer data are collected through the CoreMotion API and recorded at the maximum rate. The approximate capture rates of the multiple data streams are shown in Table~\ref{tbl:stream}. The magnetometer values are uncalibrated. The barometer samples contain both the barometric pressure and associated relative altitude readings.\sloppy

\begin{table}[tb!]
  \caption{Data captured by the devices.}
  \label{tbl:stream}
  \centering\scriptsize
  \begin{tabularx}{\columnwidth}{ l @{\extracolsep{\fill}} l l l r } 
  \toprule
  {\bf Device} & {\bf Data} & {\bf Format} & {\bf Units} & {\bf Capture rate}\\
  \midrule
    Ground-truth				& Pose & Position/orientation & Metric position & 100 Hz \\
  \midrule
	iPhone	& ARKit pose		& Position/orientation & Metric position & 60 Hz \\
			& Video			& RGB video		      & Resolution 1280$\times720$&	60 Hz \\
			& GNSS			& Latitude/Longitude	& World coordinates (incl.\ meta) & $\sim$1 Hz \\
   			& Barometer		& Pressure		& kPa & $\sim$10 Hz \\
			& Gyroscope		& Angular rate 	& rad/s & 100 Hz \\
			& Accelerometer	& Specific force & g &100 Hz\\
			& Magnetometer	& Magnetic field  & $\mu$T & 100 Hz\\
  \hline
	Pixel		& ARCore pose				& Position/orientation & Metric position & 30 Hz \\
\hline
 Tango 	& Raw pose 			& Position/orientation & Metric position & 60 Hz \\
  & Area learning 			& Position/orientation & Metric position & 60 Hz \\
				& Fisheye video			& Grayscale video & Resolution: 640$\times$480 & 60 Hz \\ 
				& Point cloud			& Array of 3D points	& Point coloud & $\sim$5 Hz \\						
  \bottomrule
  \end{tabularx}
\end{table}

\subsection{Apple ARKit Data}
The same application that captures the raw data is running the ARKit framework. It provides a pose estimate associated with every video frame. The pose is saved as a translation vector and a rotation expressed in Euler angles. Each pose is relative to a global coordinate frame created by the phone.

\subsection{Google ARCore Data}
We wrote an app based on Google's ARCore example\footnote{https://github.com/google-ar/arcore-android-sdk} for capturing the ARCore tracking result. Like ARKit, the pose data contains a translation to the first frame of the capture and a rotation to a global coordinate frame. Unlike ARKit, the orientation is stored as a unit quaternion. Note that the capture rate is slower than with ARKit. We do not save the video frames nor the sensor data on the Pixel. The capture was done on a Google Pixel device running Android~8.0.0 Oreo and using the Tango Core AR developer preview (Tango core version 1.57:2017.08.28-release-ar-sdk-preview-release-0-g0ce07954:250018377:stable).

\subsection{Google Tango Data}
A data collection app developed and published by \cite{Laskar+Huttunen+Herrera+Rahtu+Kannala}, based on the Paraview project\footnote{https://github.com/Kitware/ParaViewTangoRecorder}, was modified in order to collect the relevant data. The capture includes the position of the device relative to the first frame, the orientation in global coordinates, the fisheye grayscale image, and the point cloud created by the depth sensor.
The Tango service was run on a Project Tango tablet running Android~4.4.2 and using Tango Core Argentine (Tango Core version 1.47:2016.11-22-argentine\_tango-release-0-gce1d28c8:190012533:stable).
The Tango service produces two sets of poses, referred to as {\em raw odometry} and {\em area learning}\footnote{https://developers.google.com/tango/overview/area-learning}. The raw odometry is built frame to frame without long term memory whereas the area learning uses ongoing map building to close loops and reduce drift. Both tracks are captured and saved.

\subsection{Reference Video and Locations}
\label{sec:ref-video}
One important contribution of this paper is the flexible data collection framework that enables us to capture realistic use cases in large environments. In such conditions, it is not feasible to use visual markers, motion capture, or laser scanners for ground-truth. Instead, our work takes advantage of pure inertial navigation together with manual location fixes as described in Section~\ref{sec:gt}. 

In order to obtain the location fixes, we record an additional reference video, which is captured by an assisting person who walks within a short distance from the actual collector. Figure~\ref{fig:ref-video} illustrates an example frame of such video. The reference video allows us to determine the location of the data collection device with respect to the environment and to obtain the manual location fixes (subject to measurement noise) for the pure inertial navigation approach \cite{Solin+Cortes+Rahtu+Kannala}. 

In practice, the location fixes are produced as a post-processing step using a location marking tool developed for this paper. In this tool, one can browse the videos, and mark manual location fixes on the corresponding floor plan image. The location fixes are inserted on occasions where it is easy to determine the device position with respect to the floor plan image (\eg\ in the beginning and the end of escalators, entering and exiting elevator, passing through a door, or walking past a building corner). In all our recordings it was relatively easy to find enough such instances needed to build an accurate ground-truth. Note that it is enough to determine the device location manually, not orientation.

The initial location fixes have to be further transformed from pixel coordinates of floor plan images into metric world coordinates. This is done by first converting pixels to meters by using manually measured reference distances (\eg~distance between pillars). Then the floor plan images are registered with respect to each other using manually determined landmark points (\eg~pillars or stairs) and floor height measurements.

\section{Methods} \label{sec:methods}

\subsection{Ground-Truth}
\label{sec:gt}
The ground-truth is an implementation of the purely inertial odometry algorithm presented in \cite{Solin+Cortes+Rahtu+Kannala}, with the addition of manual fixation points recorded using the external reference video (see Sec.~\ref{sec:ref-video}). The IMU data used in the inertial navigation system for the ground-truth originated from the iPhone, and is the same data that is shared as part of the dataset. Furthermore, additional calibration data was acquired for the iPhone IMUs accounting for additive gyroscope bias, additive accelerometer bias, and multiplicative accelerometer scale bias.

The inference of the iPhone pose track (position and orientation) was implemented as described in \cite{Solin+Cortes+Rahtu+Kannala} with the addition of fusing the state estimation with both the additional calibration data and the manual fix points. The pose track corresponds to the INS estimates conditional to the fix points and external calibrations, 
\begin{equation}
  p\left(\vect{p}(t_k),\vect{q}(t_k) \mid \text{IMU}, \text{calibrations}, \{(t_i,\vect{p}_i)\}_{i=1}^N\right),
\end{equation}
where $\vect{p}(t_k) \in \R^3$ is the phone position and $\vect{q}(t_k)$ is the  orientation unit quaternion at time instant $t_k$. The set of fixpoints consists of time--position pairs $(t_i,\vect{p}_i)$, where the manual fixpoint $\vect{p}_i \in \R^3$ assigned to a time instant $t_i$. The `IMU' refers to all accelerometer and gyroscope data over the entire track.

Accounting for uncertainty and inaccuracy in the fixation point locations is taken into account by not enforcing the phone track to match the points, but including a Gaussian measurement noise term with a standard deviation of 25~cm in the position fixes (in all directions). This allows the estimate track to disagree with the fix. Position fixes are given either as 3D locations or 2D points with unknown altitude while moving between floors.

The inference problem was finally solved with an extended Kalman filter (forward pass) and extended Rauch--Tung--Striebel smoother (backward pass, see \cite{Solin+Cortes+Rahtu+Kannala} for technical details). As real-time computation is not required here, we could have also used batch optimization but that would not have caused noticeable change in the results. Calculated tracks were inspected manually frame by frame and the pose track was refined by additional fixation points until the track matched the movement seen in all three cameras and the floor plan images. Figure~\ref{fig:floors} shows examples of the estimated ground-truth track. The vertical line is an elevator ride (stopping in each floor). Walking-induced periodic movement can be seen if zoomed in. The obtained accuracy can be checked also from the example video in the supplementary material.

\subsection{Evaluation Metrics}
\label{sec:eval}
For odometry results captured on the fly while collecting the data, we propose the following evaluation metrics. All data was first temporally aligned to the same global clock (acquired by NTP requests while capturing the data), which seemed to give temporal alignments accurate to about 1--2 seconds. The temporal alignment was further improved by determining a constant time offset by minimizing the median error between the device yaw and roll tracks. This alignment accounts for both temporal registration errors between devices and internal delays in the odometry methods.

After the temporal alignment the tracks provided by the three devices are chopped to the same lengths covering the same time-span as there may be few seconds differences in the starting and stopping times of the recordings with different devices. The vertical direction is already aligned to gravity. To account for the relative poses between the devices, method estimates, and ground-truth, we estimate a planar rigid transform (2D rotation and translation) between estimate tracks and ground-truth based on the first 60~s of estimates in each method (using the entire path would not have had a clear effect on the results, though). The reason for not using the calibrated relative poses is that especially ARCore (and occasionally ARKit) showed wild jumps at the beginning of the tracks, which would have had considerable effects and ruined those datasets for the method.

The aligned tracks all start from origin, and we measure the absolute error to the ground-truth for every output given by each method. The empirical cumulative distribution function for the absolute position error is defined as
\begin{align}
  {\hat{F}}_{n}(d) 
    = \frac{\text{number of position errors} \leq d}{n}
    = \frac{1}{n} \sum_{i=1}^{n} \mathbf{1}_{e_{i} \leq d},
\end{align}
where $ \mathbf{1}_{E}$ is an indicator function for the event $E$, $\vect{e} \in \R^n$ is a vector of absolute position errors compared to ground-truth, and $n$ is the number of positions. The function tells the proportion of position estimates being less than $d$ meters from ground-truth.

\begin{figure}[t!]
  \centering
  \setlength{\figurewidth}{.09\textwidth}
  \begin{subfigure}{.49\columnwidth}
    \centering
    \includegraphics[width=\figurewidth]{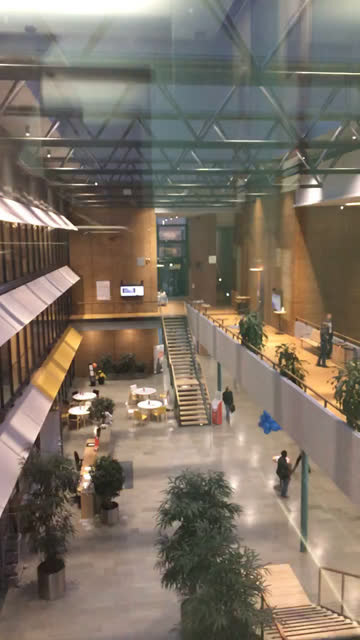}
    \includegraphics[width=\figurewidth]{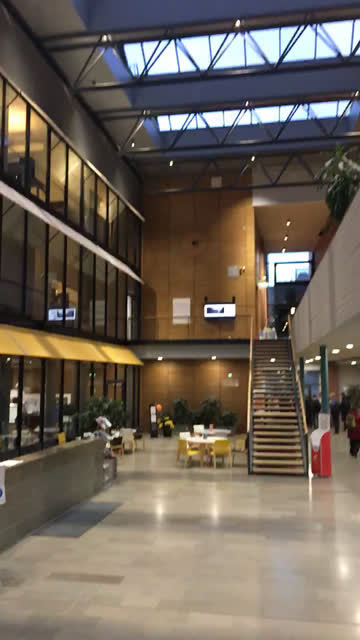}
    \includegraphics[width=\figurewidth]{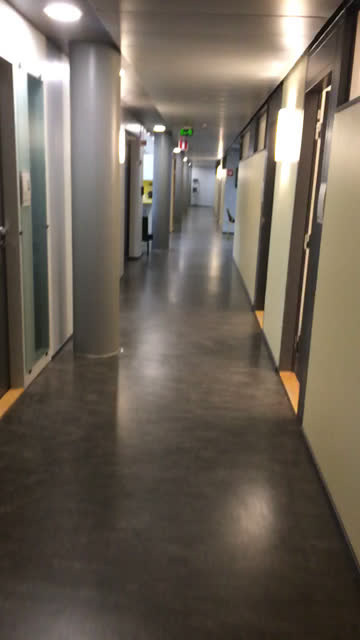}
    \includegraphics[width=\figurewidth]{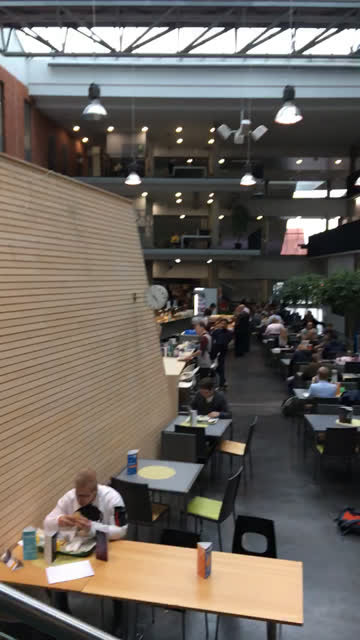}
    \includegraphics[width=\figurewidth]{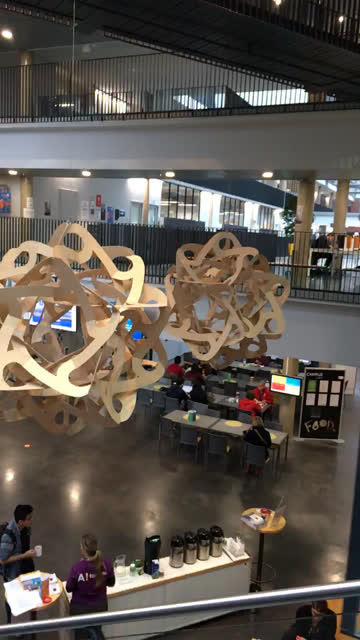} 
    \caption{Office indoor}
    \label{fig:frames-in-office}
  \end{subfigure}
  \hspace{\fill}
  \begin{subfigure}{.49\columnwidth}
    \centering  
    \includegraphics[width=\figurewidth]{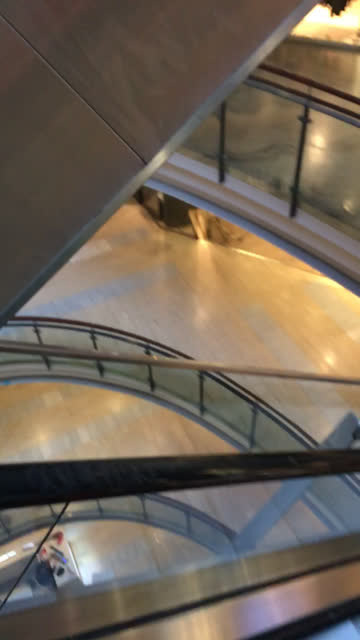}
    \includegraphics[width=\figurewidth]{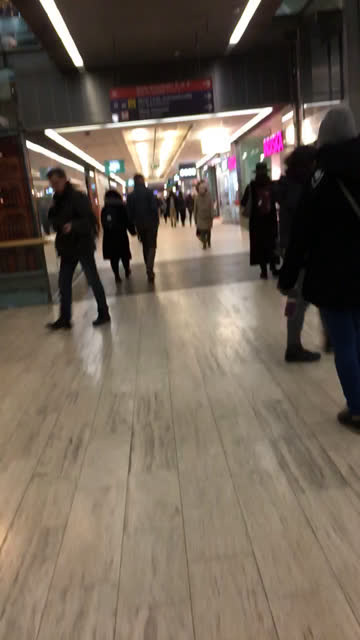}
    \includegraphics[width=\figurewidth]{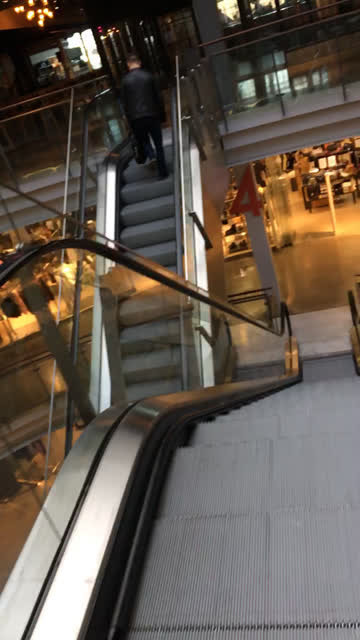}
    \includegraphics[width=\figurewidth]{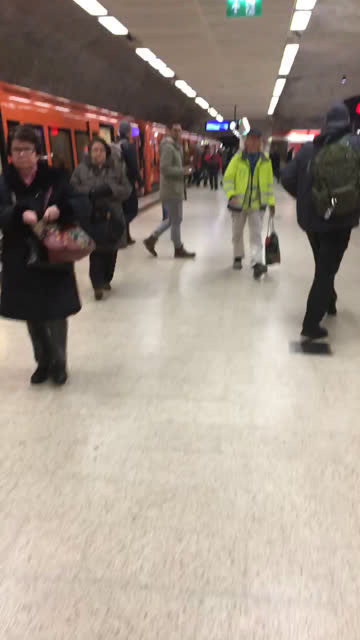}
    \includegraphics[width=\figurewidth]{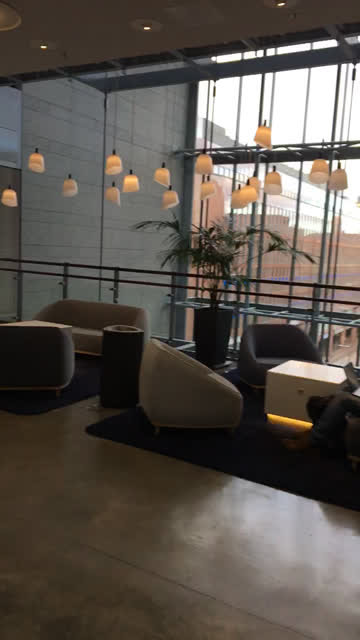} 
    \caption{Urban indoor (mall/metro)}
    \label{fig:frames-in-urban}
  \end{subfigure}
  \\[3pt]
  \begin{subfigure}{.49\columnwidth}
    \centering
    \includegraphics[width=\figurewidth]{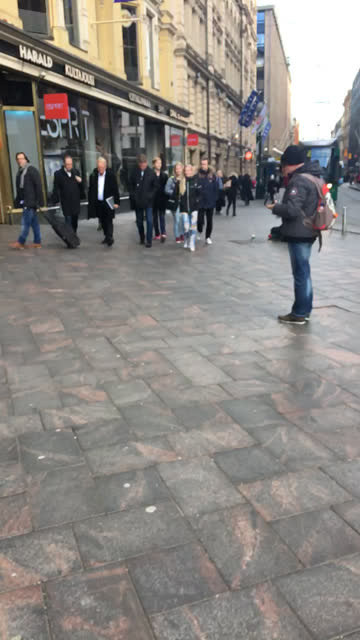}
    \includegraphics[width=\figurewidth]{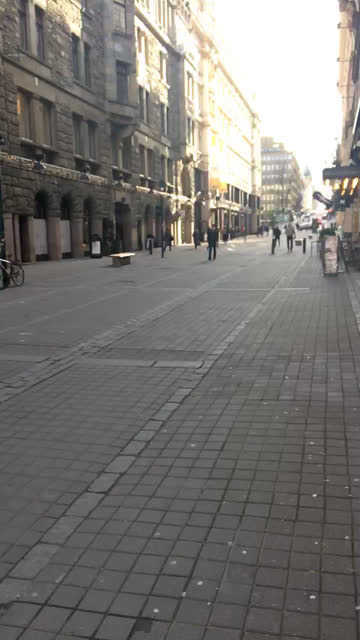}
    \includegraphics[width=\figurewidth]{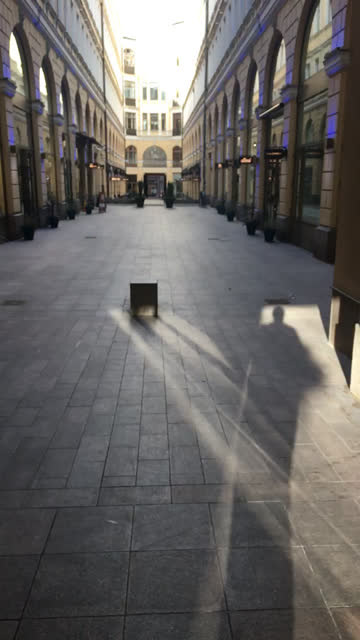}
    \includegraphics[width=\figurewidth]{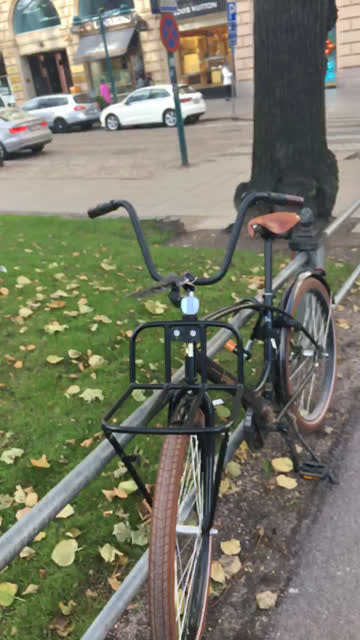}
    \includegraphics[width=\figurewidth]{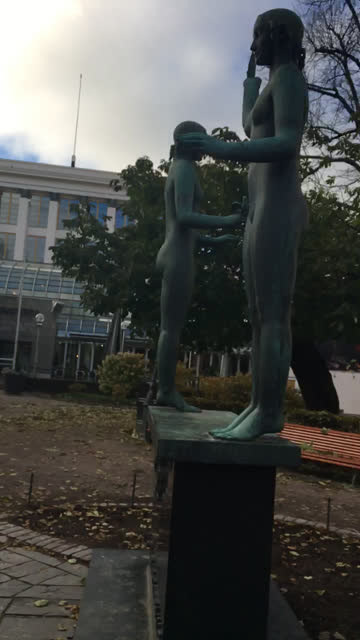} 
    \caption{Urban outdoor (city)}
    \label{fig:frames-out-urban}
  \end{subfigure}
  \hspace{\fill}
  \begin{subfigure}{.49\columnwidth}
    \centering
    \includegraphics[width=\figurewidth]{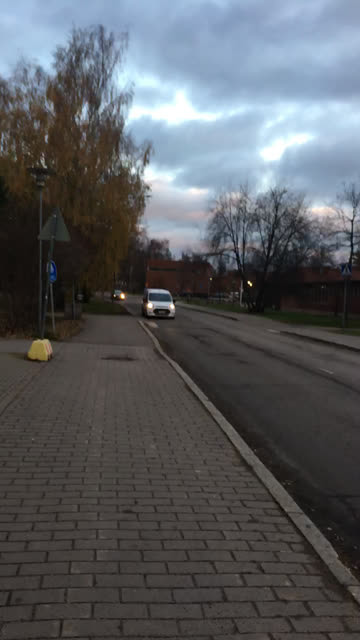}
    \includegraphics[width=\figurewidth]{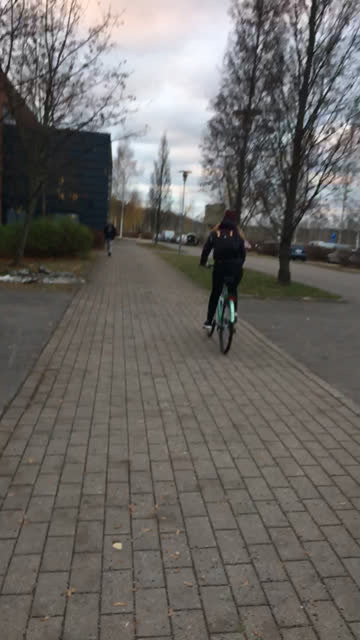}
    \includegraphics[width=\figurewidth]{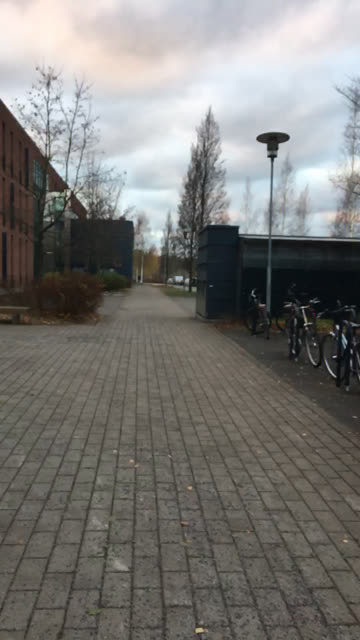}
    \includegraphics[width=\figurewidth]{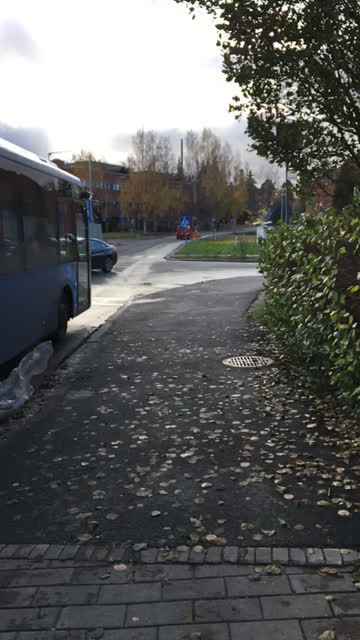}
    \includegraphics[width=\figurewidth]{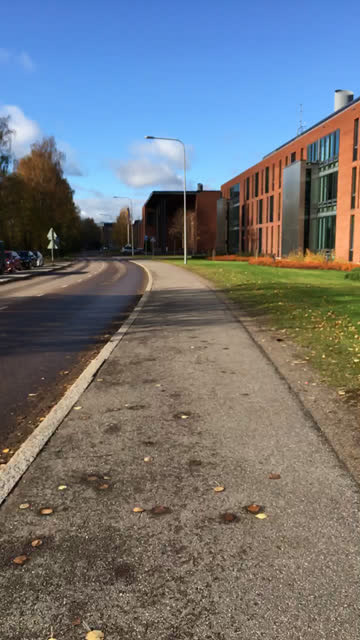} 
    \caption{Suburban outdoor (campus)}
    \label{fig:frames-out-suburban}
  \end{subfigure}\\[-1em]
  \caption{Example frames from datasets. There are 7~sequences from two separate office buildings, 12~sequences from urban indoor scences (malls and metro station), two from urban outdoor scenes, and two from suburban (campus) outdoor scenes.}
  \label{fig:example-frames}
\end{figure}

\section{Data and Results}
The dataset contains 23 separate recordings captured in six different locations. The total length of all sequences is $4.47$~kilometers and the total duration is $1$ hour $8$ minutes. There are 19 indoor and 4 outdoor sequences. In the indoor sequences there is a manual fix point on average every 3.7~meters (or 3.8~s), and outdoors every 14.7~m (or 10~s). The ground-truth 3D trajectories for all the sequences are illustrated in the supplementary material, where also additional details are given. In addition, one of the recordings and its ground-truth are illustrated in the supplementary video. The main characteristics of the dataset sequences and environments are briefly described below. 

Our dataset is primarily designed for benchmarking medium and long-range odometry. The most obvious use case is indoor navigation in large spaces, but we have also included outdoor paths for completeness. The indoor sequences were acquired in a 7-storey shopping mall ($\sim${}$135,\!000~\mathrm{m}^2$), in a metro station, and in two different office buildings.
The shopping mall and station are in the same building complex. The metro and bus station is located in the bottom floors, and there are plenty of moving people and occasional large vehicles visible in the collected videos, which makes pure visual odometry challenging. Also the lower floors of the mall contain a large number of moving persons. Figure~\ref{fig:mall} illustrates an overall view of the mall along with ground-truth path examples and a Tango point cloud (Fig.~\ref{fig:points}). Figure~\ref{fig:frames-in-urban} shows example frames from the mall and station. The use cases were as realistic as possible including motion in stairs, elevators and escalators, and also temporary occlusions and areas lacking visual features. There are ten sequences from the mall and two from the station.

Office building recordings were performed in the lobby and corridors in two office buildings. They contain some people in a static position and a few people moving. The sequences contain stair climbs and elevator rides. There are closed and open (glass) elevator sequences. Example frames are shown in Figure~\ref{fig:frames-in-office}.

The outdoor sequences were recorded in the city center (urban, two sequences) and university campus (suburban, two sequences). Figures~\ref{fig:frames-out-urban} and \ref{fig:frames-out-suburban} illustrate example frames from both locations. Urban outdoor captures were performed through city blocks; they contain open spaces, people, and vehicles. Suburban outdoor captures were performed through sparsely populated areas. They contain a few people walking and some vehicle encounters. Most of the spaces are open. The average length of the outdoor sequences is $334.6$~meters, ranging from $133$ to $514$ meters. The outdoor sequences were acquired in different times of the day illustrating several daylight conditions.  

\begin{figure}[!t]
  \raggedright\scriptsize
  \setlength{\figurewidth}{.24\columnwidth}
  \setlength{\figureheight}{0.6\figurewidth}
  \pgfplotsset{
      compat=newest,    
      hide y axis,
      tick align=outside,
      axis on top,
      minor tick num=1,
  }
  \begin{subfigure}[b]{0.27\columnwidth}
    \centering
%
%
\begin{tikzpicture}

\begin{axis}[%
xmin=0,
xmax=3,
ymin=0,
ymax=14000,
axis background/.style={fill=white},
axis x line*=bottom,
axis y line*=left,
width=\figurewidth,
height=\figureheight,
scale only axis
]

\addplot[area legend,solid,draw=white!60!black,fill=white!60!black,forget plot]
table[row sep=crcr] {%
x	y\\
0	0\\
0	1913\\
0.03	1913\\
0.03	5777\\
0.06	5777\\
0.06	4465\\
0.09	4465\\
0.09	3685\\
0.12	3685\\
0.12	3700\\
0.15	3700\\
0.15	3144\\
0.18	3144\\
0.18	2765\\
0.21	2765\\
0.21	2798\\
0.24	2798\\
0.24	2533\\
0.27	2533\\
0.27	2469\\
0.3	2469\\
0.3	2162\\
0.33	2162\\
0.33	2107\\
0.36	2107\\
0.36	2330\\
0.39	2330\\
0.39	2621\\
0.42	2621\\
0.42	2499\\
0.45	2499\\
0.45	2433\\
0.48	2433\\
0.48	2926\\
0.51	2926\\
0.51	3389\\
0.54	3389\\
0.54	3971\\
0.57	3971\\
0.57	5292\\
0.6	5292\\
0.6	8079\\
0.63	8079\\
0.63	8650\\
0.66	8650\\
0.66	7513\\
0.69	7513\\
0.69	6855\\
0.72	6855\\
0.72	6574\\
0.75	6574\\
0.75	6118\\
0.78	6118\\
0.78	5827\\
0.81	5827\\
0.81	6079\\
0.84	6079\\
0.84	6001\\
0.87	6001\\
0.87	6282\\
0.9	6282\\
0.9	6612\\
0.93	6612\\
0.93	7532\\
0.96	7532\\
0.96	8299\\
0.99	8299\\
0.99	9367\\
1.02	9367\\
1.02	9440\\
1.05	9440\\
1.05	9967\\
1.08	9967\\
1.08	10767\\
1.11	10767\\
1.11	12068\\
1.14	12068\\
1.14	12424\\
1.17	12424\\
1.17	12232\\
1.2	12232\\
1.2	11294\\
1.23	11294\\
1.23	9961\\
1.26	9961\\
1.26	8771\\
1.29	8771\\
1.29	8217\\
1.32	8217\\
1.32	7887\\
1.35	7887\\
1.35	7940\\
1.38	7940\\
1.38	8536\\
1.41	8536\\
1.41	8476\\
1.44	8476\\
1.44	8925\\
1.47	8925\\
1.47	9358\\
1.5	9358\\
1.5	9567\\
1.53	9567\\
1.53	9127\\
1.56	9127\\
1.56	8701\\
1.59	8701\\
1.59	7515\\
1.62	7515\\
1.62	6526\\
1.65	6526\\
1.65	6026\\
1.68	6026\\
1.68	5350\\
1.71	5350\\
1.71	4548\\
1.74	4548\\
1.74	4315\\
1.77	4315\\
1.77	3589\\
1.8	3589\\
1.8	2917\\
1.83	2917\\
1.83	2090\\
1.86	2090\\
1.86	1819\\
1.89	1819\\
1.89	1405\\
1.92	1405\\
1.92	1074\\
1.95	1074\\
1.95	895\\
1.98	895\\
1.98	761\\
2.01	761\\
2.01	656\\
2.04	656\\
2.04	739\\
2.07	739\\
2.07	993\\
2.1	993\\
2.1	934\\
2.13	934\\
2.13	921\\
2.16	921\\
2.16	686\\
2.19	686\\
2.19	622\\
2.22	622\\
2.22	509\\
2.25	509\\
2.25	365\\
2.28	365\\
2.28	400\\
2.31	400\\
2.31	375\\
2.34	375\\
2.34	276\\
2.37	276\\
2.37	209\\
2.4	209\\
2.4	241\\
2.43	241\\
2.43	285\\
2.46	285\\
2.46	312\\
2.49	312\\
2.49	300\\
2.52	300\\
2.52	250\\
2.55	250\\
2.55	230\\
2.58	230\\
2.58	150\\
2.61	150\\
2.61	109\\
2.64	109\\
2.64	142\\
2.67	142\\
2.67	99\\
2.7	99\\
2.7	124\\
2.73	124\\
2.73	115\\
2.76	115\\
2.76	74\\
2.79	74\\
2.79	100\\
2.82	100\\
2.82	95\\
2.85	95\\
2.85	82\\
2.88	82\\
2.88	98\\
2.91	98\\
2.91	81\\
2.94	81\\
2.94	39\\
2.97	39\\
2.97	28\\
3	28\\
3	0\\
}--cycle;
\end{axis}
\end{tikzpicture}%
    \caption{Speed, \\ all data}
    \label{fig:histogram-all}
  \end{subfigure}%
  \hspace*{\fill}
  \begin{subfigure}[b]{0.27\columnwidth}
    \centering
%
%
\begin{tikzpicture}

\begin{axis}[%
xmin=0,
xmax=3,
ymin=0,
ymax=3000,
axis background/.style={fill=white},
axis x line*=bottom,
axis y line*=left,
width=\figurewidth,
height=\figureheight,
scale only axis
]

\addplot[area legend,solid,draw=white!60!black,fill=white!60!black,forget plot]
table[row sep=crcr] {%
x	y\\
0	0\\
0	43\\
0.03	43\\
0.03	303\\
0.06	303\\
0.06	142\\
0.09	142\\
0.09	100\\
0.12	100\\
0.12	249\\
0.15	249\\
0.15	239\\
0.18	239\\
0.18	192\\
0.21	192\\
0.21	267\\
0.24	267\\
0.24	171\\
0.27	171\\
0.27	163\\
0.3	163\\
0.3	122\\
0.33	122\\
0.33	178\\
0.36	178\\
0.36	173\\
0.39	173\\
0.39	142\\
0.42	142\\
0.42	178\\
0.45	178\\
0.45	193\\
0.48	193\\
0.48	209\\
0.51	209\\
0.51	262\\
0.54	262\\
0.54	474\\
0.57	474\\
0.57	1265\\
0.6	1265\\
0.6	2828\\
0.63	2828\\
0.63	1723\\
0.66	1723\\
0.66	1167\\
0.69	1167\\
0.69	1241\\
0.72	1241\\
0.72	910\\
0.75	910\\
0.75	374\\
0.78	374\\
0.78	295\\
0.81	295\\
0.81	225\\
0.84	225\\
0.84	185\\
0.87	185\\
0.87	201\\
0.9	201\\
0.9	251\\
0.93	251\\
0.93	267\\
0.96	267\\
0.96	271\\
0.99	271\\
0.99	326\\
1.02	326\\
1.02	300\\
1.05	300\\
1.05	307\\
1.08	307\\
1.08	260\\
1.11	260\\
1.11	255\\
1.14	255\\
1.14	177\\
1.17	177\\
1.17	204\\
1.2	204\\
1.2	188\\
1.23	188\\
1.23	293\\
1.26	293\\
1.26	409\\
1.29	409\\
1.29	488\\
1.32	488\\
1.32	354\\
1.35	354\\
1.35	451\\
1.38	451\\
1.38	634\\
1.41	634\\
1.41	608\\
1.44	608\\
1.44	867\\
1.47	867\\
1.47	656\\
1.5	656\\
1.5	665\\
1.53	665\\
1.53	634\\
1.56	634\\
1.56	521\\
1.59	521\\
1.59	356\\
1.62	356\\
1.62	309\\
1.65	309\\
1.65	254\\
1.68	254\\
1.68	189\\
1.71	189\\
1.71	121\\
1.74	121\\
1.74	110\\
1.77	110\\
1.77	126\\
1.8	126\\
1.8	119\\
1.83	119\\
1.83	63\\
1.86	63\\
1.86	77\\
1.89	77\\
1.89	56\\
1.92	56\\
1.92	55\\
1.95	55\\
1.95	54\\
1.98	54\\
1.98	49\\
2.01	49\\
2.01	27\\
2.04	27\\
2.04	38\\
2.07	38\\
2.07	35\\
2.1	35\\
2.1	38\\
2.13	38\\
2.13	46\\
2.16	46\\
2.16	21\\
2.19	21\\
2.19	23\\
2.22	23\\
2.22	23\\
2.25	23\\
2.25	17\\
2.28	17\\
2.28	11\\
2.31	11\\
2.31	12\\
2.34	12\\
2.34	11\\
2.37	11\\
2.37	10\\
2.4	10\\
2.4	11\\
2.43	11\\
2.43	9\\
2.46	9\\
2.46	14\\
2.49	14\\
2.49	25\\
2.52	25\\
2.52	17\\
2.55	17\\
2.55	14\\
2.58	14\\
2.58	14\\
2.61	14\\
2.61	13\\
2.64	13\\
2.64	15\\
2.67	15\\
2.67	13\\
2.7	13\\
2.7	4\\
2.73	4\\
2.73	0\\
2.76	0\\
2.76	0\\
2.79	0\\
2.79	0\\
2.82	0\\
2.82	0\\
2.85	0\\
2.85	0\\
2.88	0\\
2.88	0\\
2.91	0\\
2.91	0\\
2.94	0\\
2.94	0\\
2.97	0\\
2.97	0\\
3	0\\
3	0\\
}--cycle;
\end{axis}
\end{tikzpicture}%
    \caption{Speed, \\ escalators/walking}
    \label{fig:histogram-escalator}
  \end{subfigure}
  \begin{minipage}[c]{0.4\columnwidth}
    ~
  \end{minipage}\\[6pt]
  \begin{subfigure}[t]{0.27\columnwidth}
    \centering
%
%
\begin{tikzpicture}

\begin{axis}[%
xmin=-3.14159265358979,
xmax=3.14159265358979,
xtick={-3.14159265358979,-1.5707963267949,0,1.5707963267949,3.14159265358979},
xticklabels={{$-\pi$},{$-\pi/2$},{0},{$\pi/2$},{$-\pi$}},
ymin=0,
ymax=120000,
axis background/.style={fill=white},
axis x line*=bottom,
axis y line*=left,
width=\figurewidth,
height=\figureheight,
scale only axis
]

\addplot[area legend,solid,draw=white!60!black,fill=white!60!black,forget plot]
table[row sep=crcr] {%
x	y\\
-3.14159265358979	0\\
-3.14159265358979	0\\
-3.078760800518	0\\
-3.078760800518	0\\
-3.0159289474462	0\\
-3.0159289474462	0\\
-2.95309709437441	0\\
-2.95309709437441	0\\
-2.89026524130261	0\\
-2.89026524130261	0\\
-2.82743338823081	0\\
-2.82743338823081	0\\
-2.76460153515902	0\\
-2.76460153515902	0\\
-2.70176968208722	0\\
-2.70176968208722	0\\
-2.63893782901543	0\\
-2.63893782901543	0\\
-2.57610597594363	0\\
-2.57610597594363	0\\
-2.51327412287183	0\\
-2.51327412287183	0\\
-2.45044226980004	0\\
-2.45044226980004	0\\
-2.38761041672824	0\\
-2.38761041672824	0\\
-2.32477856365645	0\\
-2.32477856365645	0\\
-2.26194671058465	0\\
-2.26194671058465	0\\
-2.19911485751286	0\\
-2.19911485751286	2\\
-2.13628300444106	2\\
-2.13628300444106	3\\
-2.07345115136926	3\\
-2.07345115136926	0\\
-2.01061929829747	0\\
-2.01061929829747	0\\
-1.94778744522567	0\\
-1.94778744522567	1\\
-1.88495559215388	1\\
-1.88495559215388	0\\
-1.82212373908208	0\\
-1.82212373908208	0\\
-1.75929188601028	0\\
-1.75929188601028	0\\
-1.69646003293849	0\\
-1.69646003293849	0\\
-1.63362817986669	0\\
-1.63362817986669	0\\
-1.5707963267949	0\\
-1.5707963267949	0\\
-1.5079644737231	0\\
-1.5079644737231	0\\
-1.4451326206513	0\\
-1.4451326206513	0\\
-1.38230076757951	0\\
-1.38230076757951	6\\
-1.31946891450771	6\\
-1.31946891450771	165\\
-1.25663706143592	165\\
-1.25663706143592	398\\
-1.19380520836412	398\\
-1.19380520836412	403\\
-1.13097335529233	403\\
-1.13097335529233	138\\
-1.06814150222053	138\\
-1.06814150222053	83\\
-1.00530964914873	83\\
-1.00530964914873	719\\
-0.942477796076938	719\\
-0.942477796076938	616\\
-0.879645943005142	616\\
-0.879645943005142	1016\\
-0.816814089933346	1016\\
-0.816814089933346	1295\\
-0.753982236861551	1295\\
-0.753982236861551	1514\\
-0.691150383789755	1514\\
-0.691150383789755	1438\\
-0.628318530717959	1438\\
-0.628318530717959	3145\\
-0.565486677646163	3145\\
-0.565486677646163	4511\\
-0.502654824574367	4511\\
-0.502654824574367	7671\\
-0.439822971502571	7671\\
-0.439822971502571	9408\\
-0.376991118430775	9408\\
-0.376991118430775	17488\\
-0.314159265358979	17488\\
-0.314159265358979	48309\\
-0.251327412287184	48309\\
-0.251327412287184	81477\\
-0.188495559215387	81477\\
-0.188495559215387	101023\\
-0.125663706143592	101023\\
-0.125663706143592	64035\\
-0.062831853071796	64035\\
-0.062831853071796	34914\\
0	34914\\
0	12315\\
0.062831853071796	12315\\
0.062831853071796	5672\\
0.125663706143592	5672\\
0.125663706143592	3161\\
0.188495559215388	3161\\
0.188495559215388	2873\\
0.251327412287183	2873\\
0.251327412287183	2575\\
0.314159265358979	2575\\
0.314159265358979	524\\
0.376991118430775	524\\
0.376991118430775	112\\
0.439822971502571	112\\
0.439822971502571	113\\
0.502654824574367	113\\
0.502654824574367	103\\
0.565486677646163	103\\
0.565486677646163	67\\
0.628318530717958	67\\
0.628318530717958	116\\
0.691150383789755	116\\
0.691150383789755	100\\
0.75398223686155	100\\
0.75398223686155	79\\
0.816814089933346	79\\
0.816814089933346	137\\
0.879645943005142	137\\
0.879645943005142	92\\
0.942477796076938	92\\
0.942477796076938	6\\
1.00530964914873	6\\
1.00530964914873	11\\
1.06814150222053	11\\
1.06814150222053	3\\
1.13097335529233	3\\
1.13097335529233	3\\
1.19380520836412	3\\
1.19380520836412	3\\
1.25663706143592	3\\
1.25663706143592	2\\
1.31946891450771	2\\
1.31946891450771	0\\
1.38230076757951	0\\
1.38230076757951	0\\
1.4451326206513	0\\
1.4451326206513	0\\
1.5079644737231	0\\
1.5079644737231	0\\
1.5707963267949	0\\
1.5707963267949	0\\
1.63362817986669	0\\
1.63362817986669	0\\
1.69646003293849	0\\
1.69646003293849	0\\
1.75929188601028	0\\
1.75929188601028	0\\
1.82212373908208	0\\
1.82212373908208	0\\
1.88495559215388	0\\
1.88495559215388	0\\
1.94778744522567	0\\
1.94778744522567	0\\
2.01061929829747	0\\
2.01061929829747	0\\
2.07345115136926	0\\
2.07345115136926	0\\
2.13628300444106	0\\
2.13628300444106	0\\
2.19911485751286	0\\
2.19911485751286	0\\
2.26194671058465	0\\
2.26194671058465	0\\
2.32477856365645	0\\
2.32477856365645	0\\
2.38761041672824	0\\
2.38761041672824	0\\
2.45044226980004	0\\
2.45044226980004	0\\
2.51327412287183	0\\
2.51327412287183	0\\
2.57610597594363	0\\
2.57610597594363	0\\
2.63893782901543	0\\
2.63893782901543	0\\
2.70176968208722	0\\
2.70176968208722	0\\
2.76460153515902	0\\
2.76460153515902	0\\
2.82743338823081	0\\
2.82743338823081	0\\
2.89026524130261	0\\
2.89026524130261	0\\
2.95309709437441	0\\
2.95309709437441	0\\
3.0159289474462	0\\
3.0159289474462	0\\
3.078760800518	0\\
3.078760800518	0\\
3.14159265358979	0\\
3.14159265358979	0\\
}--cycle;
\end{axis}
\end{tikzpicture}%
    \caption{Orientation, roll}
    \label{fig:histogram-roll}
  \end{subfigure}%
  \hspace*{\fill}
  \begin{subfigure}[t]{0.27\columnwidth}
    \centering
%
%
\begin{tikzpicture}

\begin{axis}[%
xmin=-3.14159265358979,
xmax=3.14159265358979,
xtick={-3.14159265358979,-1.5707963267949,0,1.5707963267949,3.14159265358979},
xticklabels={{$-\pi$},{$-\pi/2$},{0},{$\pi/2$},{$-\pi$}},
ymin=0,
ymax=250000,
axis background/.style={fill=white},
axis x line*=bottom,
axis y line*=left,
width=\figurewidth,
height=\figureheight,
scale only axis
]

\addplot[area legend,solid,draw=white!60!black,fill=white!60!black,forget plot]
table[row sep=crcr] {%
x	y\\
-3.14159265358979	0\\
-3.14159265358979	0\\
-3.078760800518	0\\
-3.078760800518	0\\
-3.0159289474462	0\\
-3.0159289474462	0\\
-2.95309709437441	0\\
-2.95309709437441	0\\
-2.89026524130261	0\\
-2.89026524130261	0\\
-2.82743338823081	0\\
-2.82743338823081	0\\
-2.76460153515902	0\\
-2.76460153515902	0\\
-2.70176968208722	0\\
-2.70176968208722	0\\
-2.63893782901543	0\\
-2.63893782901543	0\\
-2.57610597594363	0\\
-2.57610597594363	0\\
-2.51327412287183	0\\
-2.51327412287183	2\\
-2.45044226980004	2\\
-2.45044226980004	1\\
-2.38761041672824	1\\
-2.38761041672824	1\\
-2.32477856365645	1\\
-2.32477856365645	1\\
-2.26194671058465	1\\
-2.26194671058465	8\\
-2.19911485751286	8\\
-2.19911485751286	10\\
-2.13628300444106	10\\
-2.13628300444106	22\\
-2.07345115136926	22\\
-2.07345115136926	18\\
-2.01061929829747	18\\
-2.01061929829747	25\\
-1.94778744522567	25\\
-1.94778744522567	248\\
-1.88495559215388	248\\
-1.88495559215388	481\\
-1.82212373908208	481\\
-1.82212373908208	827\\
-1.75929188601028	827\\
-1.75929188601028	934\\
-1.69646003293849	934\\
-1.69646003293849	7329\\
-1.63362817986669	7329\\
-1.63362817986669	73331\\
-1.5707963267949	73331\\
-1.5707963267949	226764\\
-1.5079644737231	226764\\
-1.5079644737231	85714\\
-1.4451326206513	85714\\
-1.4451326206513	9955\\
-1.38230076757951	9955\\
-1.38230076757951	1526\\
-1.31946891450771	1526\\
-1.31946891450771	452\\
-1.25663706143592	452\\
-1.25663706143592	70\\
-1.19380520836412	70\\
-1.19380520836412	17\\
-1.13097335529233	17\\
-1.13097335529233	16\\
-1.06814150222053	16\\
-1.06814150222053	23\\
-1.00530964914873	23\\
-1.00530964914873	12\\
-0.942477796076938	12\\
-0.942477796076938	5\\
-0.879645943005142	5\\
-0.879645943005142	12\\
-0.816814089933346	12\\
-0.816814089933346	12\\
-0.753982236861551	12\\
-0.753982236861551	4\\
-0.691150383789755	4\\
-0.691150383789755	6\\
-0.628318530717959	6\\
-0.628318530717959	6\\
-0.565486677646163	6\\
-0.565486677646163	3\\
-0.502654824574367	3\\
-0.502654824574367	5\\
-0.439822971502571	5\\
-0.439822971502571	4\\
-0.376991118430775	4\\
-0.376991118430775	1\\
-0.314159265358979	1\\
-0.314159265358979	0\\
-0.251327412287184	0\\
-0.251327412287184	0\\
-0.188495559215387	0\\
-0.188495559215387	0\\
-0.125663706143592	0\\
-0.125663706143592	0\\
-0.062831853071796	0\\
-0.062831853071796	0\\
0	0\\
0	0\\
0.062831853071796	0\\
0.062831853071796	0\\
0.125663706143592	0\\
0.125663706143592	0\\
0.188495559215388	0\\
0.188495559215388	0\\
0.251327412287183	0\\
0.251327412287183	0\\
0.314159265358979	0\\
0.314159265358979	0\\
0.376991118430775	0\\
0.376991118430775	0\\
0.439822971502571	0\\
0.439822971502571	0\\
0.502654824574367	0\\
0.502654824574367	0\\
0.565486677646163	0\\
0.565486677646163	0\\
0.628318530717958	0\\
0.628318530717958	0\\
0.691150383789755	0\\
0.691150383789755	0\\
0.75398223686155	0\\
0.75398223686155	0\\
0.816814089933346	0\\
0.816814089933346	0\\
0.879645943005142	0\\
0.879645943005142	0\\
0.942477796076938	0\\
0.942477796076938	0\\
1.00530964914873	0\\
1.00530964914873	0\\
1.06814150222053	0\\
1.06814150222053	0\\
1.13097335529233	0\\
1.13097335529233	0\\
1.19380520836412	0\\
1.19380520836412	0\\
1.25663706143592	0\\
1.25663706143592	0\\
1.31946891450771	0\\
1.31946891450771	0\\
1.38230076757951	0\\
1.38230076757951	0\\
1.4451326206513	0\\
1.4451326206513	0\\
1.5079644737231	0\\
1.5079644737231	0\\
1.5707963267949	0\\
1.5707963267949	0\\
1.63362817986669	0\\
1.63362817986669	0\\
1.69646003293849	0\\
1.69646003293849	0\\
1.75929188601028	0\\
1.75929188601028	0\\
1.82212373908208	0\\
1.82212373908208	0\\
1.88495559215388	0\\
1.88495559215388	0\\
1.94778744522567	0\\
1.94778744522567	0\\
2.01061929829747	0\\
2.01061929829747	0\\
2.07345115136926	0\\
2.07345115136926	0\\
2.13628300444106	0\\
2.13628300444106	0\\
2.19911485751286	0\\
2.19911485751286	0\\
2.26194671058465	0\\
2.26194671058465	0\\
2.32477856365645	0\\
2.32477856365645	0\\
2.38761041672824	0\\
2.38761041672824	0\\
2.45044226980004	0\\
2.45044226980004	0\\
2.51327412287183	0\\
2.51327412287183	0\\
2.57610597594363	0\\
2.57610597594363	0\\
2.63893782901543	0\\
2.63893782901543	0\\
2.70176968208722	0\\
2.70176968208722	0\\
2.76460153515902	0\\
2.76460153515902	0\\
2.82743338823081	0\\
2.82743338823081	0\\
2.89026524130261	0\\
2.89026524130261	0\\
2.95309709437441	0\\
2.95309709437441	0\\
3.0159289474462	0\\
3.0159289474462	0\\
3.078760800518	0\\
3.078760800518	0\\
3.14159265358979	0\\
3.14159265358979	0\\
}--cycle;
\end{axis}
\end{tikzpicture}%
    \caption{Orientation, yaw}
    \label{fig:histogram-yaw}
  \end{subfigure}
  \begin{subfigure}[t]{0.4\columnwidth}
    \centering
    \setlength{\figurewidth}{1.2\columnwidth}

    \vspace*{-\figurewidth}\rotatebox{-90}{\input{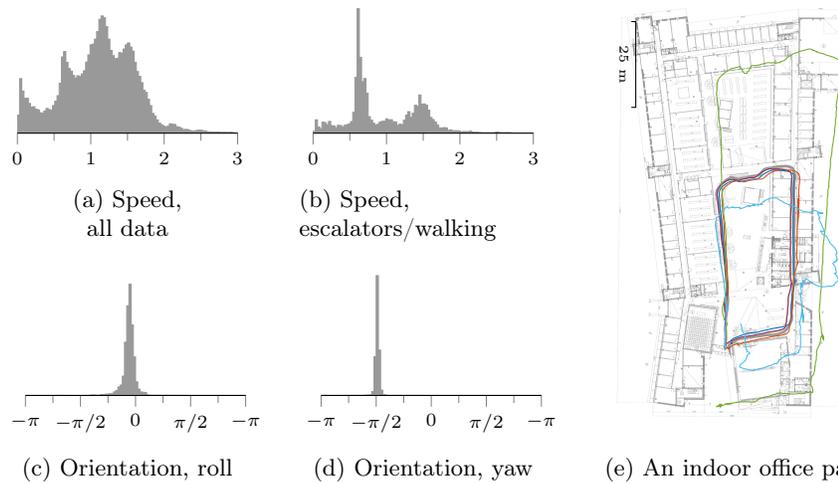}}

    \caption{An indoor office path}
    \label{fig:comparison-all}
  \end{subfigure}\\[-1em]
  \caption{(a)~Speed histograms;
 peaks correspond to escalators, stairs, and walking. (b)~the histogram for one data set with escalator rides/walking. (c--d) the histogram for roll and yaw. (e)~the paths for \textcolor{white!60!black}{ground-truth} (\ref{addplot:cs01-comparison0}), \textcolor{mycolor1}{ARKit} (\ref{addplot:cs01-comparison1}), \textcolor{mycolor2}{ARCore} (\ref{addplot:cs01-comparison2}), \textcolor{mycolor3}{Tango/Raw} (\ref{addplot:cs01-comparison3}), \textcolor{mycolor4}{Tango/Area learning} (\ref{addplot:cs01-comparison4}), \textcolor{mycolor5}{ROVIO} (\ref{addplot:cs01-comparison5}), and \textcolor{mycolor6}{PIVO}~(\ref{addplot:cs01-comparison6}).}
  \label{fig:histograms}
\end{figure}

Figure~\ref{fig:histogram-all} shows the histograms of different motion metrics extracted from the ground-truth. Figure~\ref{fig:histogram-all} shows the speed histogram which has three peaks that reflect the three main motion modes. From slower to faster they are escalator, stairs, and walking. Figure~\ref{fig:histogram-escalator} shows the speed histogram for just one sequence that contained both escalator rides and normal walking. The orientation histograms show that the phone was kept generally in the same position relative to the carrier (portrait orientation, slightly pointing downward). The pitch angle which reflects the heading direction has a close to uniform distribution.

\subsection{Benchmark Results}
We evaluated two research level VIO systems using the raw iPhone data and the three proprietary solutions run on the respective devices (ARCore on Pixel, ARKit on iPhone, and Tango on the tablet). The research systems used were ROVIO~\cite{Blosch+Omari+Hutter+Siegwart:2015,Bloesch+Burri+Omari+Hutter+Siegwart:2017,Schneider+Dymczyk+Fehr+Egger+Lynen+Gilitschenski+Siegwart:2018} and PIVO \cite{Solin+Cortes+Rahtu+Kannala:2017}. ROVIO is a fairly recent method, which has been shown to work well on high-quality IMU and large field-of-view camera data. PIVO is a recent method which has shown promising results in comparison with Google Tango \cite{Solin+Cortes+Rahtu+Kannala:2017} using smartphone data. For both methods, implementations (ROVIO as part of maplab\footnote{\url{https://github.com/ethz-asl/maplab}}) from the original authors were used (in odometry-only mode without map building or loop-closures). We used pre-calibrated camera parameters and rigid transformation from camera to IMU, and pre-estimated the process and measurement noise scale parameters.

For testing purposes, we also ran two visual-only odometry methods on the raw data (DSO \cite{Engel+Koltun+Cremers:2016} and ORB-SLAM2 \cite{Mur-Artal+Tardos:2016}). Both were able to track subsets of the paths, but the small field-of-view, rapid motion with rotations, and challenging environments caused them not to succeed for any of the entire paths.

\begin{figure}[t!]
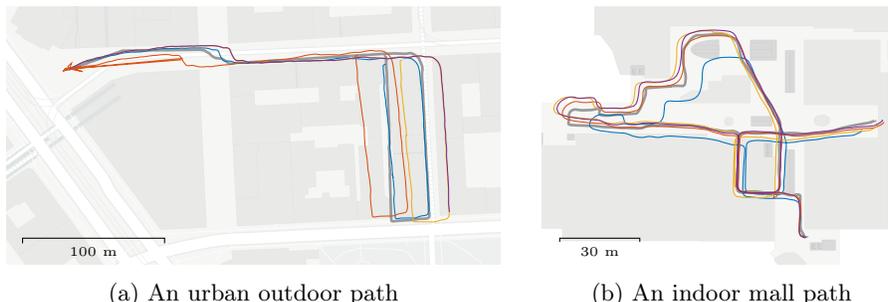

   \centering\footnotesize
   \setlength{\figurewidth}{.28\columnwidth}
   \begin{subfigure}[t]{.58\columnwidth}
     \centering   
     \setlength{\figurewidth}{1.9577\figurewidth}
     \input{./fig/outdoor.tex}
     \caption{An urban outdoor path}
     \label{fig:path-outdoor}
   \end{subfigure}
   \hspace*{\fill}
   \begin{subfigure}[t]{.40\columnwidth}
     \centering
     \setlength{\figurewidth}{1.4013\figurewidth}
     \input{./fig/indoor.tex}
     \caption{An indoor mall path}
     \label{fig:path-indoor}
   \end{subfigure}\\[-1em]
  \caption{Example paths showing \textcolor{white!60!black}{ground-truth} (\ref{addplot:outdoor0}), \textcolor{mycolor1}{ARKit} (\ref{addplot:outdoor1}), \textcolor{mycolor2}{ARCore} (\ref{addplot:outdoor2}), \textcolor{mycolor3}{Tango/Raw} (\ref{addplot:outdoor3}), and \textcolor{mycolor4}{Tango/Area learning} (\ref{addplot:outdoor4}) that stopped prematurely in (a). Map data {\copyright}\ OpenStreetMap. The ground-truth fix points were marked on an architectural drawing. ROVIO and PIVO diverge and are not shown.}
  \label{fig:path-examples}
\end{figure}

\begin{figure}[t!]
  \centering\scriptsize
  \setlength{\figurewidth}{.38\columnwidth}
  \setlength{\figureheight}{0.66\figurewidth}
  \pgfplotsset{grid style={dotted,draw=gray!50,line width=.1pt}}
  \begin{subfigure}{.48\columnwidth}
%
%
\definecolor{mycolor1}{rgb}{0.00000,0.44700,0.74100}%
\definecolor{mycolor2}{rgb}{0.85000,0.32500,0.09800}%
\definecolor{mycolor3}{rgb}{0.92900,0.69400,0.12500}%
\definecolor{mycolor4}{rgb}{0.49400,0.18400,0.55600}%
\definecolor{mycolor5}{rgb}{0.46600,0.67400,0.18800}%
\definecolor{mycolor6}{rgb}{0.30100,0.74500,0.93300}%
\begin{tikzpicture}

\begin{axis}[%
xmode=log,
xmin=0.1,
xmax=100,
xminorticks=true,
xmajorgrids,
xminorgrids,
ymin=0,
ymax=1,
ylabel={Proportion of positions},
ymajorgrids,
axis background/.style={fill=white},
legend style={legend cell align=left,align=left,draw=white!15!black},
width=\figurewidth,
height=\figureheight,
scale only axis
]
\addplot [color=mycolor1,solid,line width=1.0pt,forget plot]
  table[row sep=crcr]{%
0.1	0.00578182684292009\\
0.153992652605949	0.00964630225080386\\
0.237137370566166	0.0156722639037752\\
0.365174127254838	0.0269322377039419\\
0.562341325190349	0.0500178635226867\\
0.865964323360065	0.0845956889365249\\
1.33352143216332	0.176366559485531\\
2.05352502645715	0.311658925806836\\
3.16227766016838	0.531124211027748\\
4.86967525165863	0.76892342503275\\
7.49894209332456	0.925711563653686\\
11.5478198468946	0.993979992854591\\
17.7827941003892	0.998237465761582\\
27.3841963426436	0.999994045492438\\
42.1696503428582	1\\
64.9381631576211	1\\
100	1\\
};
\label{addplot:position-cdf-indoor0}
\addplot [color=mycolor2,solid,line width=1.0pt,forget plot]
  table[row sep=crcr]{%
0.1	0.00571421768788467\\
0.153992652605949	0.0136307901096416\\
0.237137370566166	0.0204997559552862\\
0.365174127254838	0.0408447518481923\\
0.562341325190349	0.0812014142688778\\
0.865964323360065	0.157795740526898\\
1.33352143216332	0.275091963190914\\
2.05352502645715	0.403614242687587\\
3.16227766016838	0.514993869120606\\
4.86967525165863	0.628694896489328\\
7.49894209332456	0.740800704753515\\
11.5478198468946	0.8513589123939\\
17.7827941003892	0.910382019261676\\
27.3841963426436	0.942119736669802\\
42.1696503428582	0.990095356007667\\
64.9381631576211	0.996547660146903\\
100	1\\
};
\label{addplot:position-cdf-indoor1}
\addplot [color=mycolor3,solid,line width=1.0pt,forget plot]
  table[row sep=crcr]{%
0.1	0.00610023183017638\\
0.153992652605949	0.00975396087732231\\
0.237137370566166	0.0179695095242674\\
0.365174127254838	0.0383000544854332\\
0.562341325190349	0.0759804707114088\\
0.865964323360065	0.17701355725778\\
1.33352143216332	0.34726451075286\\
2.05352502645715	0.593538668632416\\
3.16227766016838	0.806181425809002\\
4.86967525165863	0.902973195303569\\
7.49894209332456	0.99033150646881\\
11.5478198468946	1\\
17.7827941003892	1\\
27.3841963426436	1\\
42.1696503428582	1\\
64.9381631576211	1\\
100	1\\
};
\label{addplot:position-cdf-indoor2}
\addplot [color=mycolor4,solid,line width=1.0pt,forget plot]
  table[row sep=crcr]{%
0.1	0.00608130339577259\\
0.153992652605949	0.0117087781799204\\
0.237137370566166	0.0239621506937905\\
0.365174127254838	0.047266249900725\\
0.562341325190349	0.0840490588729166\\
0.865964323360065	0.18832752810901\\
1.33352143216332	0.390326643143217\\
2.05352502645715	0.627145758404339\\
3.16227766016838	0.84312279467659\\
4.86967525165863	0.949103121206276\\
7.49894209332456	0.997957771247688\\
11.5478198468946	1\\
17.7827941003892	1\\
27.3841963426436	1\\
42.1696503428582	1\\
64.9381631576211	1\\
100	1\\
};
\label{addplot:position-cdf-indoor3}
\addplot [color=mycolor5,dashed,line width=1.0pt,forget plot]
  table[row sep=crcr]{%
0.1	0.00261186898792352\\
0.153992652605949	0.00470500441378557\\
0.237137370566166	0.00880026937742872\\
0.365174127254838	0.0160352374798649\\
0.562341325190349	0.0303504636749998\\
0.865964323360065	0.0457031569942575\\
1.33352143216332	0.0685092325473458\\
2.05352502645715	0.110817869916184\\
3.16227766016838	0.168579307081168\\
4.86967525165863	0.284966737347906\\
7.49894209332456	0.402327930617111\\
11.5478198468946	0.475687776999172\\
17.7827941003892	0.588780794117379\\
27.3841963426436	0.626348024717199\\
42.1696503428582	0.657117115477371\\
64.9381631576211	0.699225539892431\\
100	0.729221080603915\\
};
\label{addplot:position-cdf-indoor4}
\addplot [color=mycolor6,dashed,line width=1.0pt,forget plot]
  table[row sep=crcr]{%
0.1	0.000997686368886489\\
0.153992652605949	0.00151262126895694\\
0.237137370566166	0.00273916759343029\\
0.365174127254838	0.00538178489309737\\
0.562341325190349	0.0116182186828394\\
0.865964323360065	0.0191384137859516\\
1.33352143216332	0.0277027824364288\\
2.05352502645715	0.0372898690134348\\
3.16227766016838	0.049208466387982\\
4.86967525165863	0.0677067874856516\\
7.49894209332456	0.112552610970259\\
11.5478198468946	0.192299577681863\\
17.7827941003892	0.304558961834026\\
27.3841963426436	0.452423948763977\\
42.1696503428582	0.569142526113279\\
64.9381631576211	0.710635193654858\\
100	0.882054876326226\\
};
\label{addplot:position-cdf-indoor5}
\end{axis}
\end{tikzpicture}%
    \caption{Absolute error (m), indoor data sets}
    \label{fig:position-cdf-indoor}
  \end{subfigure}
  \hspace{\fill}
  \begin{subfigure}{.48\columnwidth}
%
%
\definecolor{mycolor1}{rgb}{0.00000,0.44700,0.74100}%
\definecolor{mycolor2}{rgb}{0.85000,0.32500,0.09800}%
\definecolor{mycolor3}{rgb}{0.92900,0.69400,0.12500}%
\definecolor{mycolor4}{rgb}{0.49400,0.18400,0.55600}%
\definecolor{mycolor5}{rgb}{0.46600,0.67400,0.18800}%
\definecolor{mycolor6}{rgb}{0.30100,0.74500,0.93300}%
\begin{tikzpicture}

\begin{axis}[%
xmode=log,
xmin=0.1,
xmax=100,
xminorticks=true,
xmajorgrids,
xminorgrids,
ymin=0,
ymax=1,
ylabel={Proportion of positions},
ymajorgrids,
axis background/.style={fill=white},
legend style={legend cell align=left,align=left,draw=white!15!black},
width=\figurewidth,
height=\figureheight,
scale only axis
]
\addplot [color=mycolor1,solid,line width=1.0pt,forget plot]
  table[row sep=crcr]{%
0.1	0.00429134530137231\\
0.153992652605949	0.00713836779035259\\
0.237137370566166	0.0116336664571635\\
0.365174127254838	0.0200248906352107\\
0.562341325190349	0.0368614490678499\\
0.865964323360065	0.0623431328069394\\
1.33352143216332	0.129060857186859\\
2.05352502645715	0.232090605242018\\
3.16227766016838	0.416343740504722\\
4.86967525165863	0.621666507111313\\
7.49894209332456	0.764512946876392\\
11.5478198468946	0.837927833807143\\
17.7827941003892	0.861777890622724\\
27.3841963426436	0.889041044574216\\
42.1696503428582	0.93119279420273\\
64.9381631576211	0.951221847151521\\
100	0.951925278146605\\
};
\label{addplot:position-cdf0}
\addplot [color=mycolor2,solid,line width=1.0pt,forget plot]
  table[row sep=crcr]{%
0.1	0.00421154262314184\\
0.153992652605949	0.009829707189586\\
0.237137370566166	0.0147570455945267\\
0.365174127254838	0.0295557072228788\\
0.562341325190349	0.0583456794233682\\
0.865964323360065	0.112995855043031\\
1.33352143216332	0.19701862733674\\
2.05352502645715	0.293010170958667\\
3.16227766016838	0.378614352537746\\
4.86967525165863	0.478159905448371\\
7.49894209332456	0.595417242355135\\
11.5478198468946	0.713823181795482\\
17.7827941003892	0.778710901736221\\
27.3841963426436	0.923185124764869\\
42.1696503428582	0.992941920663193\\
64.9381631576211	0.997486391556939\\
100	0.999925091139114\\
};
\label{addplot:position-cdf1}
\addplot [color=mycolor3,solid,line width=1.0pt,forget plot]
  table[row sep=crcr]{%
0.1	0.00451408123849022\\
0.153992652605949	0.00715665284244883\\
0.237137370566166	0.0132128580197931\\
0.365174127254838	0.0283870581365753\\
0.562341325190349	0.0558084173017323\\
0.865964323360065	0.12877483493285\\
1.33352143216332	0.252548996122232\\
2.05352502645715	0.434534592984085\\
3.16227766016838	0.598389004506595\\
4.86967525165863	0.683812190265155\\
7.49894209332456	0.77397403841835\\
11.5478198468946	0.814915183183363\\
17.7827941003892	0.915183183363028\\
27.3841963426436	0.948406222395233\\
42.1696503428582	0.998607596831908\\
64.9381631576211	1\\
100	1\\
};
\label{addplot:position-cdf2}
\addplot [color=mycolor4,solid,line width=1.0pt,forget plot]
  table[row sep=crcr]{%
0.1	0.00500017857780635\\
0.153992652605949	0.00958069931068967\\
0.237137370566166	0.0195185542340798\\
0.365174127254838	0.0385013750491089\\
0.562341325190349	0.068118504232294\\
0.865964323360065	0.151523268688167\\
1.33352143216332	0.317743490838959\\
2.05352502645715	0.513911211114683\\
3.16227766016838	0.694248008857459\\
4.86967525165863	0.787688846030215\\
7.49894209332456	0.843315832708311\\
11.5478198468946	0.883067252401871\\
17.7827941003892	0.95504303725133\\
27.3841963426436	0.96817743490839\\
42.1696503428582	1\\
64.9381631576211	1\\
100	1\\
};
\label{addplot:position-cdf3}
\addplot [color=mycolor5,dashed,line width=1.0pt,forget plot]
  table[row sep=crcr]{%
0.1	0.00176568106463188\\
0.153992652605949	0.00315097424717264\\
0.237137370566166	0.0058647863014943\\
0.365174127254838	0.0105429895081074\\
0.562341325190349	0.0197915247308898\\
0.865964323360065	0.0304707725848208\\
1.33352143216332	0.0478437116773402\\
2.05352502645715	0.0790582277331153\\
3.16227766016838	0.12836671662806\\
4.86967525165863	0.228726665758278\\
7.49894209332456	0.323182086569469\\
11.5478198468946	0.38972725621111\\
17.7827941003892	0.524566244265795\\
27.3841963426436	0.609029386383249\\
42.1696503428582	0.70218467547804\\
64.9381631576211	0.776121860380615\\
100	0.812082708815915\\
};
\label{addplot:position-cdf4}
\addplot [color=mycolor6,dashed,line width=1.0pt,forget plot]
  table[row sep=crcr]{%
0.1	0.000820824929053699\\
0.153992652605949	0.00124875499877377\\
0.237137370566166	0.00217718807401438\\
0.365174127254838	0.00414916991576534\\
0.562341325190349	0.00869373720589192\\
0.865964323360065	0.0143594312283845\\
1.33352143216332	0.0209785834763938\\
2.05352502645715	0.0284761184990916\\
3.16227766016838	0.0377329215861941\\
4.86967525165863	0.0527430067217554\\
7.49894209332456	0.0873427794933909\\
11.5478198468946	0.148166407239275\\
17.7827941003892	0.237869058403696\\
27.3841963426436	0.349421168274116\\
42.1696503428582	0.451190946901636\\
64.9381631576211	0.568829173319186\\
100	0.722260872176537\\
};
\label{addplot:position-cdf5}
\end{axis}
\end{tikzpicture}%
    \caption{Absolute error (m), all data sets}
    \label{fig:position-cdf-all}
  \end{subfigure}\\[-1em]
  \caption{Cumulative distributions of position error: \textcolor{mycolor1}{ARKit}~(\ref{addplot:position-cdf0}), \textcolor{mycolor2}{ARCore}~(\ref{addplot:position-cdf1}), \textcolor{mycolor3}{Tango/Raw} (\ref{addplot:position-cdf2}), \textcolor{mycolor4}{Tango/Area learning} (\ref{addplot:position-cdf3}), \textcolor{mycolor5}{ROVIO} (\ref{addplot:position-cdf4}), and \textcolor{mycolor6}{PIVO}~(\ref{addplot:position-cdf5}).}
  \label{fig:position-cdf}
\end{figure}
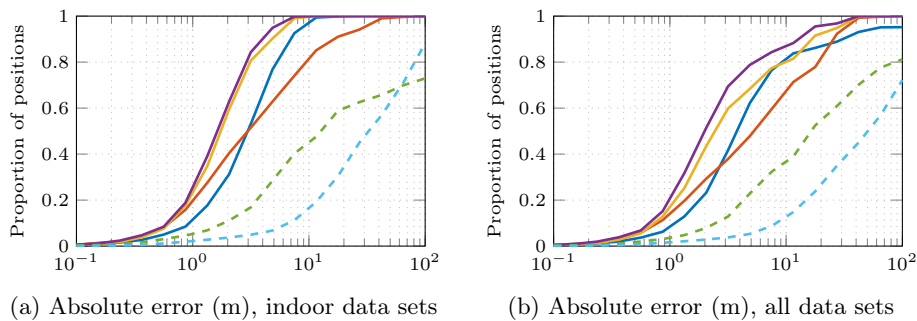

In general, the proprietary systems work better than the research methods, as shown in Figure~\ref{fig:position-cdf}. In indoor sequences, all proprietary systems work well in general (Fig.~\ref{fig:position-cdf-indoor}). Tango has the best performance, ARKit performs well and robustly with only a few clear failure cases (95th percentile $\sim$10~meters), and ARCore occasionally fails, apparently due to incorrect visual loop-closures. Including the outdoor sequences changes the metrics slightly (Fig.\ref{fig:position-cdf-all}). ARKit had severe problems with drifting in the outdoor sequences. In terms of the orientation error all systems were accurate with less than $<\!2^\circ$ error from the ground-truth on average. This is due to the orientation tracking by integrating the gyroscope performing well if the gyroscope is well calibrated.

As shown in Figure~\ref{fig:position-cdf}, the research methods have challenges with our iPhone data which has narrow field-of-view and a low-cost IMU. There are many sequences where both methods diverge completely (\eg~Fig.~\ref{fig:path-examples}). On the other hand, there are also sequences where they work reasonably well. This may be partially explained by the fact that both ROVIO and PIVO estimate the calibration parameters of the IMU (\eg~accelerometer and gyroscope biases) internally on the fly and neither software directly supports giving pre-calibrated IMU parameters as input. ROVIO only considers additive accelerometer bias, which shows in many sequences as exponential crawl in position. We provide the ground-truth IMU calibration parameters with our data, and it would hence be possible to evaluate their performance also with pre-calibrated values. Alternatively, part of the sequences could be used for self-calibration and others for testing. Proprietary systems may benefit from factory-calibrated parameters.
Figures~\ref{fig:comparison-all} and \ref{fig:path-examples} show examples of the results. In these cases all commercial solutions worked well. Still, ARCore had some issues at the beginning of the outdoor path. Moreover, in multi-floor cases drifting was typically more severe and there were sequences where also proprietary systems had clear failures.

In general, ROVIO had problems with long-term occlusions and disagreements between visual and inertial data. Also, in Figure~\ref{fig:comparison-all} it has clearly inaccurate scale---most likely due to the not modelled scale bias in the accelerations, which is clearly inadequate for consumer-grade sensors that also show multiplicative biases \cite{Shelley:2014}. On the other hand, PIVO uses a model with both additive and multiplicate accelerometer biases. However, with PIVO the main challenge seems to be that without suitable motion the online calibration of various IMU parameters from scratch for each sequence takes considerable time and hence slows convergence onto the right track.

\section{Discussion and Conclusion}
We have presented the first public benchmark dataset for long-range visual-inertial odometry for hand-held devices using standard smartphone sensors. The dataset contains 23 sequences recorded both outdoors and indoors on multiple floor levels in varying authentic environments. The total length of the sequences is 4.5~km.
In addition, we provide quantitative comparison of three proprietary visual-inertial odometry platforms and two recent academic VIO methods, where we use the raw sensor data. To the best of our knowledge, this is the first back-to-back comparison of ARKit, ARCore, and Tango.

Apple's ARKit performed well in most scenarios. Only in one hard outdoor sequence the ARKit had the classic inertial dead-reckoning failure where the estimated position grew out of control. Google's ARCore showed more aggressive visual loop-closure use than ARKit, which is seen in false positive `jumps' scattered troughout the tracks (between visually similar areas). The specialized hardware in the Tango gives it a upper hand, which can also be seen in Figure~\ref{fig:position-cdf}. The area learning was the most robust and accurate system tested. However, all systems performed relatively well in the open elevator where the glass walls let the camera see the open lobby as the elevator moves. In the case of the closed elevator none of the systems were capable of reconciling the inertial motion with the static visual scene.
The need for a dataset of this kind is clear from the ROVIO and PIVO results. The community needs challenging narrow field-of-view and low-grade IMU data for developing and testing new VIO methods that generalize to customer-grade hardware.

The collection procedure scales well to new environments. Hence, in future the dataset can be extended with a reasonably small effort. The purpose of the dataset is to enable fair comparison of visual-inertial odometry methods and to speed up development in this area of research. This is relevant because VIO is currently the most common approach for enabling real-time tracking of mobile devices for augmented reality.

Further details of the dataset and the download links can be found on the web page: \url{https://github.com/AaltoVision/ADVIO}.

{\small
\bibliographystyle{splncs04}
\bibliography{bibliography}
}

\clearpage
\appendix

\begin{center}
{\bf\Large {\large Supplement for} \\ {ADVIO: An Authentic Dataset \\[3pt] for Visual-Inertial Odometry}}
\end{center}

\section{Description of supplementary video}

The attached supplementary video shows the ground-truth track for data set \#16 (captured in one of the two office buildings). The visualized track is the ground-truth track calculated from the entire IMU data sequence. The fix points used for track calculation are visualized by dots. The track on the current floor shows in red. The video has been sped-up.

\begin{center}
  \begin{tikzpicture}
    \node[anchor=south west,inner sep=0] (image) at (0,0) %
      {\fbox{\includegraphics[width=.6\textwidth,keepaspectratio]{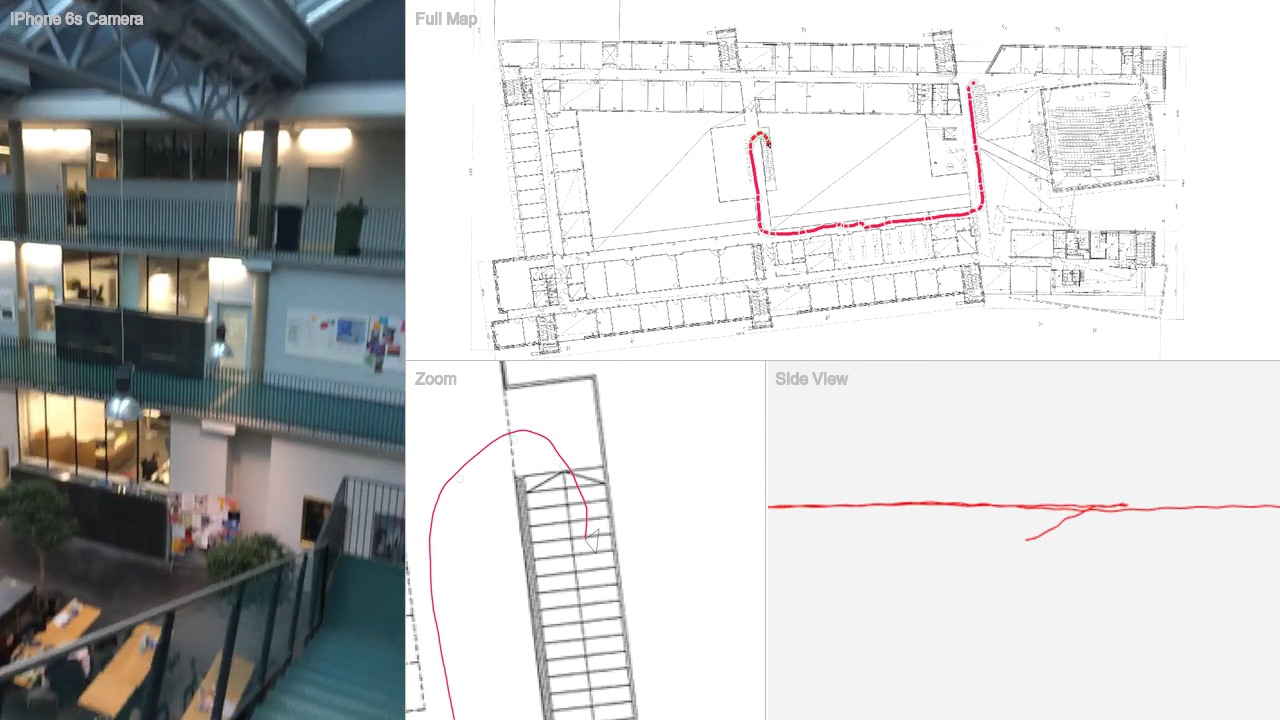}}};
    \begin{scope}[x={(image.south east)},y={(image.north west)}]

      \tikzstyle{label} = [text width=2cm, align=center]
      \tikzstyle{line}  = [draw, very thick, black!50]
      \tikzstyle{point} = [circle,draw,very thick,black!50,fill=black!50,
                           minimum size=1mm,inner sep=0]

      \node [label] (lab_1) at (-0.20,0.90) {Captured video};
      \node [label] (lab_2) at (-0.20,0.10) {Zoomed in view};
      \node [label] (lab_3) at ( 1.20,0.90) {Ground-truth track};
      \node [label] (lab_4) at ( 1.20,0.10) {Vertical view};

      \node [point] (p1) at (0.15,0.70) {};
      \node [point] (p2) at (0.45,0.30) {};
      \node [point] (p3) at (0.85,0.70) {};
      \node [point] (p4) at (0.85,0.30) {};

      \path [line] (lab_1) |- (p1);
      \path [line] (lab_2) |- (p2);
      \path [line] (lab_3) |- (p3);
      \path [line] (lab_4) |- (p4);

    \end{scope}
  \end{tikzpicture}
\end{center}

\section{Details on collected data}

\subsection{Ground-truth}

\begin{description}

\item[Ground-truth poses:] Camera pose (translation and orientation) calculated based on the raw IMU data and a set of known fixation points. The ground-truth track is sampled at 100~Hz.

\item[Fix points:] A set of ground-truth points marked with a visual editor. The points are based on the three videos stored by the system (primarily the iPhone and the second iPhone that filmed a reference track showing the capturer) and floor plan layouts.

\end{description}

\subsection{iPhone}

\begin{description}

\item[Camera frames:] Camera frames are captured at 60~fps (1280 $\times$ 720, portrait). The exact frame acquisition times reported by the platform are stored. The frames are packed into an H.264/MPEG-4 video file.

\item[Platform location:] Data collected through CoreLocation. The update rate depends on the device and its capabilities. Locations are requested with the desired accuracy of kCLLocationAccuracyBest. The timestamps are converted to follow the same clock as the other sensors (time interval since device boot). The stored values are
\begin{itemize}
    \item coordinate.latitude
    \item coordinate.longitude
    \item horizontalAccuracy
    \item altitude
    \item verticalAccuracy
    \item speed
\end{itemize}

\item[Accelerometer:] Data collected through CoreMotion/CMMotionManager. Acquired at 100 Hz, which is the maximum rate. CoreMotion reports the accelerations in `g's (at standstill you expect to have 1~g in the vertical direction).

\item[Gyroscope:] Data collected through CoreMotion/CMMotionManager. Acquired at 100 Hz, which is the maximum rate. Note that the readings are in the Apple device coordinate frame (not altered in any way here).

\item[Magnetometer:] Data collected through CoreMotion/CMMotionManager. Acquired at 100 Hz, which is the maximum rate. Values are the three-axis magnetometer readings in uT. All values are uncalibrated.

\item[Barometric altimeter:] Data collected through CoreMotion/CMAltimeter. Acquired at an uneven sampling rate ($\sim$1~Hz). Samples are stored as they arrive from the delegare callback. The actual barometric pressure is in val0 and the inferred relative altutude (calculated by Apple magic) is stored in val1.

\item[ARKit poses:] The Apple ARKit poses (translation and orientation) are captured at 60~Hz. The camera parameters reported by ARKit on the iPhone are stored as well.

\end{description}

\subsection{Tango}

\begin{description}

\item[Tango poses (raw):] The Google Tango raw poses (translation and orientation) are captured at 60~Hz.

\item[Tango poses (area learning):] The Google Tango area learning poses (translation and orientation) are captured at 60~Hz.

\item[Camera frames:] Video from the wide-angle (fisheye) camera on the Tango. Captured at $\sim$5~fps / 640$\times$480. The frames are packed into an MPEG-4 video file.

\item[Tango point clouds:] Tango point cloud data acquired by the Tango device and aligned to the current pose of the device. Sampling rate is not uniform. Timestamps are stored in \texttt{point-cloud.csv}. The actual point clouds are stored in the corresponding \texttt{point-cloud-\$index.csv}. \sloppy

\end{description}

\subsection{Pixel}

\begin{description}

\item[ARCore poses:] The Google ARCore poses (translation and orientation) are captured at 30~Hz.

\end{description}

\clearpage

\section{Dataset structure}
To maximize compatibility, all data is published in open and simple file formats. The comma separated value (CSV) files hold a timestamp in the first column and the respective data in the columns that follow. All time stamps are synchronized between sensor types and devices. Camera frames are stored as H.264/MPEG video and the associated frame time stamps are available in separate CSV files. 

The folder structure for one data set looks like the following:
\begin{verbatim}
data
  advio-01
    ground-truth
      poses.csv
      fixpoints.csv
    iphone
      frames.mov
      frames.csv
      platform-location.csv
      accelerometer.csv
      gyroscope.csv
      magnetometer.csv
      barometer.csv
      arkit.csv
    tango
      frames.mov
      frames.csv
      raw.csv
      area-learning.csv
      point-cloud.csv
      point-cloud-001.csv
      point-cloud-002.csv
      ...
    pixel
      arcore.csv
      
   advio-02
   ...
\end{verbatim}

\clearpage

\section{List of data set features}
{
\footnotesize
\resizebox{\textwidth}{!}{%
\begin{tabularx}{1.05\textwidth}{c c c c c c c c c}
\toprule
{\bf N:o}  & {\bf Venue}  & {\bf Data set}  & {\bf In/Out}  & {\bf Stairs}  & {\bf Escalator}  & {\bf Elevator}  & {\bf People}  & {\bf Vehicles} \\
\midrule
1  & Mall  & 01  & Indoor  &  & \checkmark &  & Moderate  &  \\
2  & Mall  & 02  & Indoor  &  &  &  & Moderate  &  \\
3  & Mall  & 03  & Indoor  &  &  &  & Moderate  &  \\
4  & Mall  & 04  & Indoor  &  & \checkmark &  & Moderate  &  \\
5  & Mall  & 05  & Indoor  & \checkmark &  &  & Moderate  &  \\ \midrule
6  & Mall  & 06  & Indoor  &  &  &  & High  &  \\
7  & Mall  & 07  & Indoor  &  &  & \checkmark & Low  &  \\
8  & Mall  & 08  & Indoor  &  & \checkmark &  & Low  &  \\
9  & Mall  & 09  & Indoor  &  & \checkmark &  & Low  &  \\
10  & Mall  & 10  & Indoor  &  &  &  & Low  &  \\ \midrule
11  & Metro  & 01  & Indoor  &  &  &  & High  & \checkmark \\
12  & Metro  & 02  & Indoor  &  &  &  & High  & \checkmark \\
13  & Office  & 01  & Indoor  & \checkmark &  &  & Low  &  \\
14  & Office  & 02  & Indoor  & \checkmark &  & \checkmark & Low  &  \\
15  & Office  & 03  & Indoor  &  &  &  & None  &  \\ \midrule
16  & Office  & 04  & Indoor  & \checkmark &  &  & None  &  \\
17  & Office  & 05  & Indoor  & \checkmark &  &  & None  &  \\
18  & Office  & 06  & Indoor  & \checkmark &  & \checkmark & None  &  \\
19  & Office  & 07  & Indoor  & \checkmark &  &  & None  &  \\
20  & Outdoor & 01  & Outdoor  &  &  &  & Low  & \checkmark \\ \midrule
21  & Outdoor & 02  & Outdoor  &  &  &  & Low  & \checkmark \\
22  & Outdoor (urban)  & 01  & Outdoor  &  &  &  & High  & \checkmark \\
23  & Outdoor (urban)  & 02  & Outdoor  &  &  &  & High  & \checkmark \\
\bottomrule
\end{tabularx}
}}

\clearpage

\section{Data set paths}

The following table lists each path shape (top/side views) and summary information. The fix points are visualized in the top view by dots.

\begin{longtable}{@{\extracolsep{\fill}}m{4cm} m{4cm} m{3.5cm} @{}}
{\bf Path (top)} & {\bf Path (side view)} & {\bf Information} \\
\midrule
\includegraphics[width=4cm]{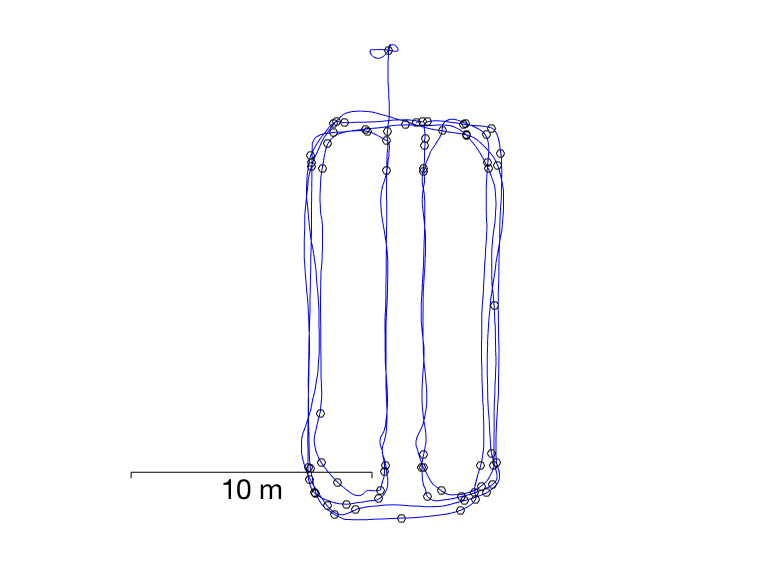} & \includegraphics[width=4cm]{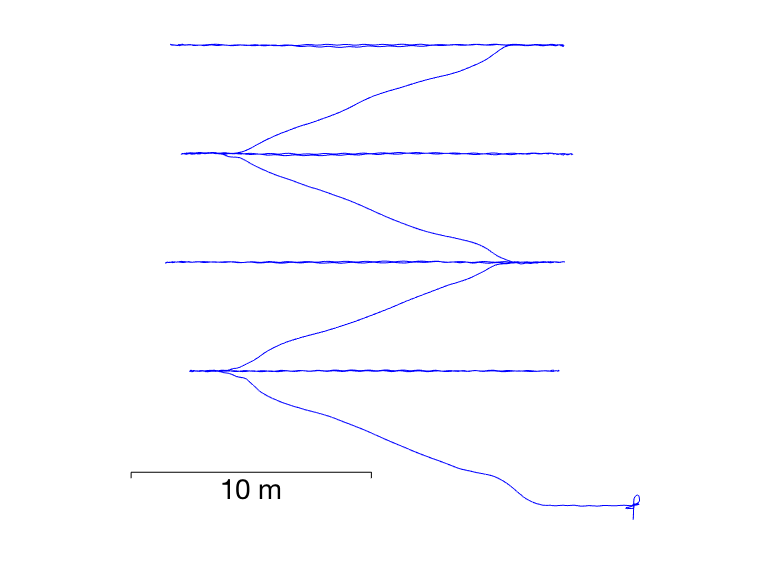} & Number: 1 \newline Length: 245.4 m \newline Duration: 4 min 20 s \newline Levels: 5 \newline Fix points: 70 \\ 
\includegraphics[width=4cm]{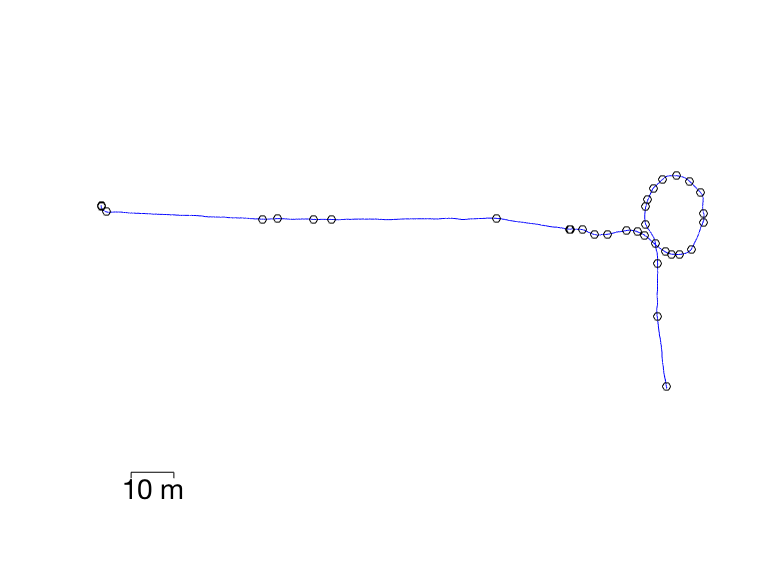} & \includegraphics[width=4cm]{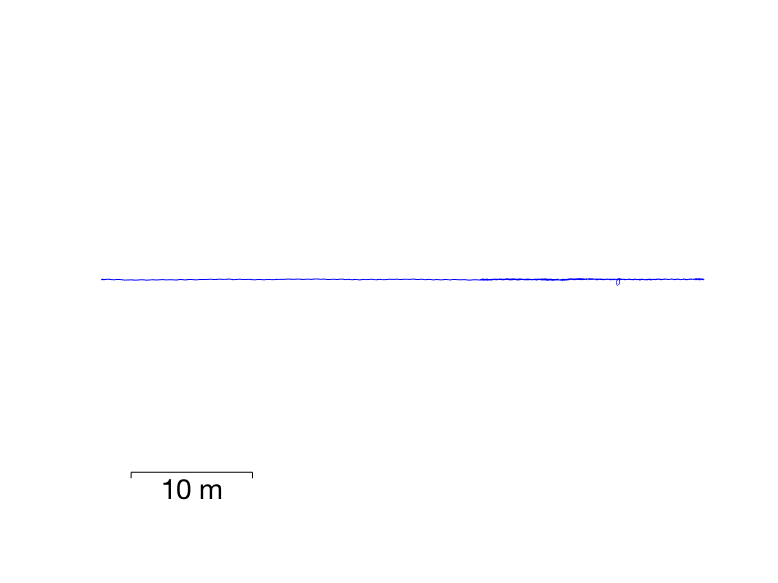} & Number: 2 \newline Length: 220.6 m \newline Duration: 3 min 19 s \newline Levels: 1 \newline Fix points: 35 \\ 
\includegraphics[width=4cm]{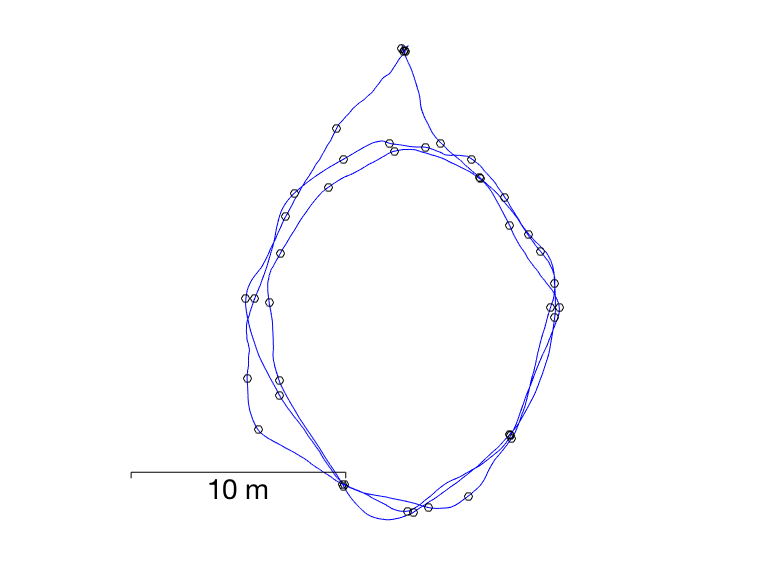} & \includegraphics[width=4cm]{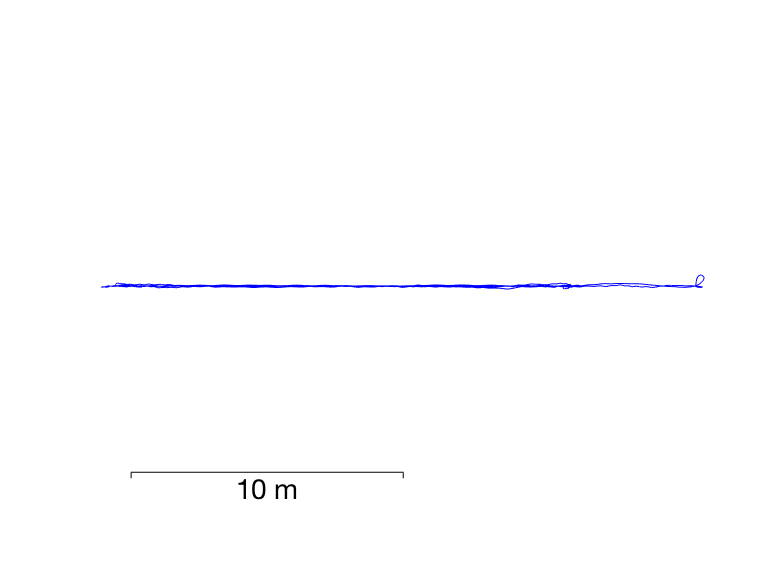} & Number: 3 \newline Length: 153.1 m \newline Duration: 2 min 28 s \newline Levels: 1 \newline Fix points: 42 \\ 
\includegraphics[width=4cm]{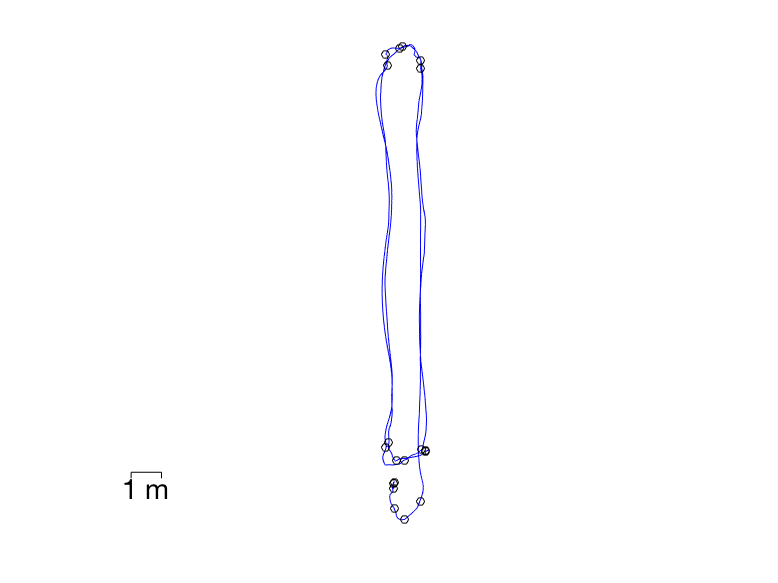} & \includegraphics[width=4cm]{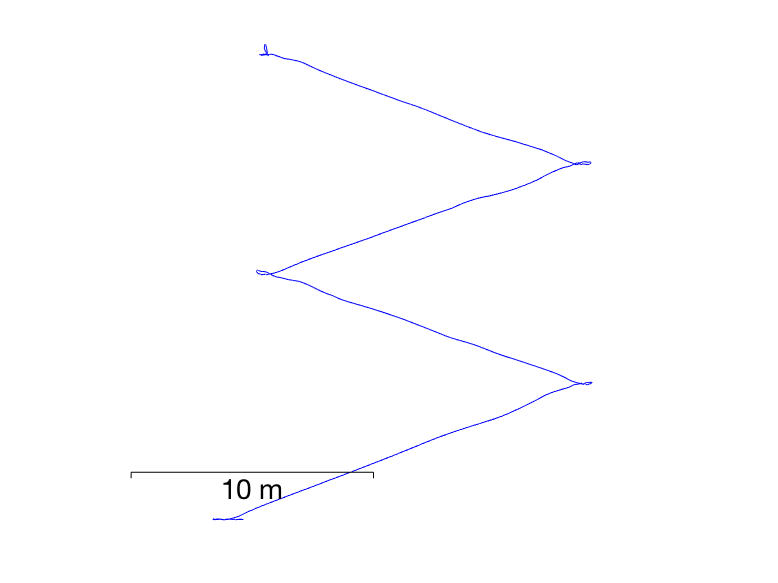} & Number: 4 \newline Length: 68.1 m \newline Duration: 1 min 57 s \newline Levels: 5 \newline Fix points: 20 \\ 
\includegraphics[width=4cm]{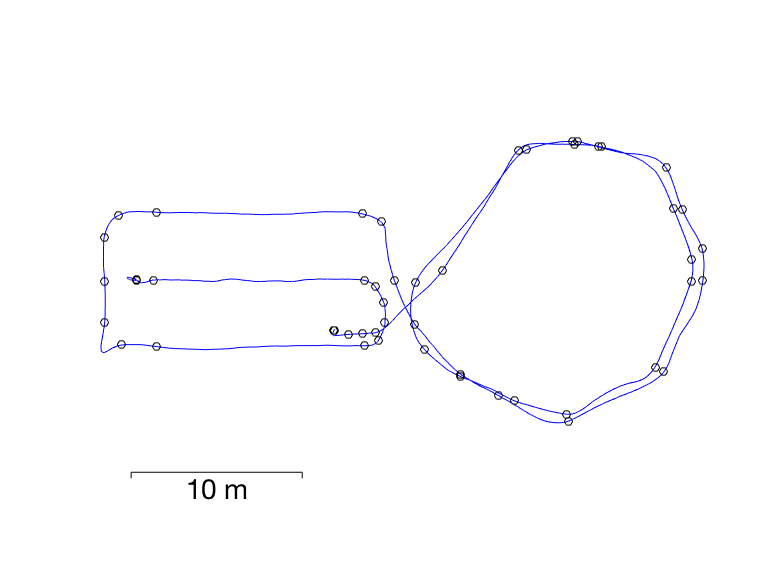} & \includegraphics[width=4cm]{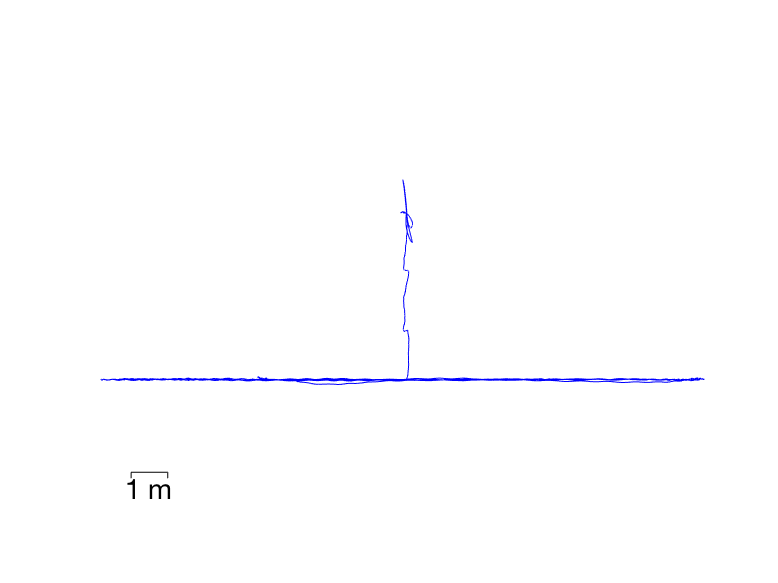} & Number: 5 \newline Length: 176.4 m \newline Duration: 2 min 41 s \newline Levels: 2 \newline Fix points: 50 \\ 
\includegraphics[width=4cm]{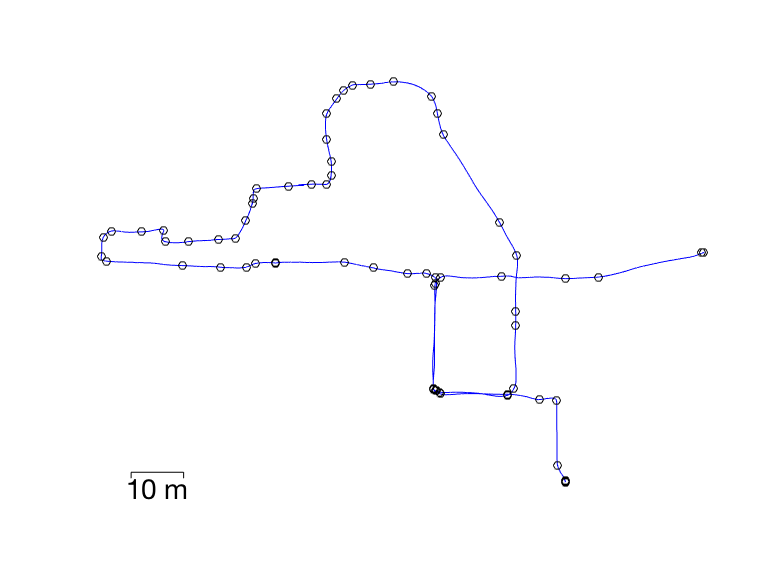} & \includegraphics[width=4cm]{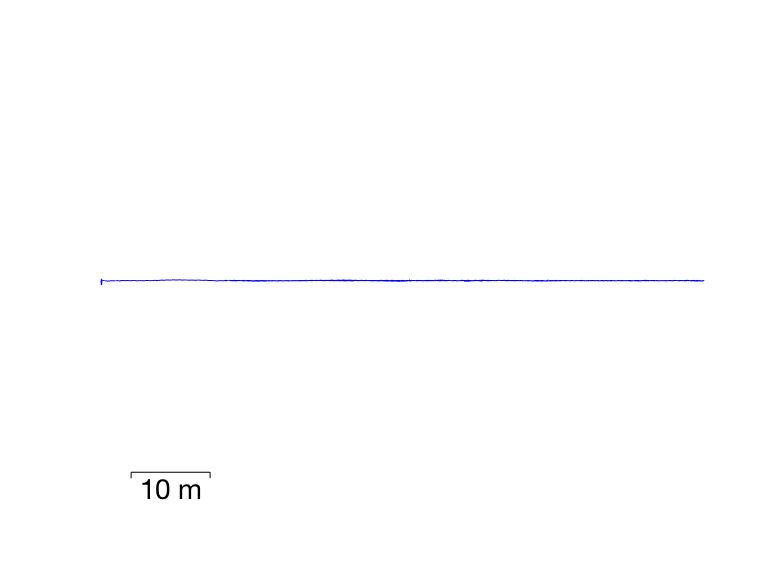} & Number: 6 \newline Length: 374.0 m \newline Duration: 4 min 36 s \newline Levels: 1 \newline Fix points: 68 \\ 
\includegraphics[width=4cm]{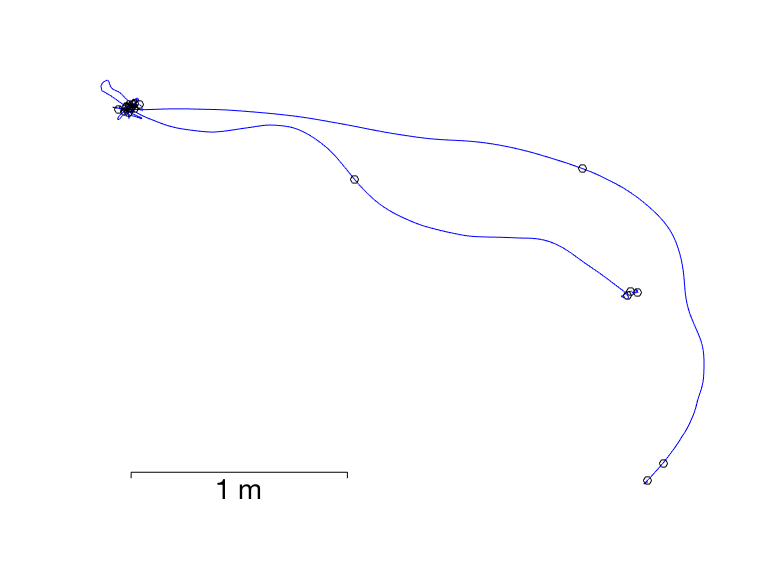} & \includegraphics[width=4cm]{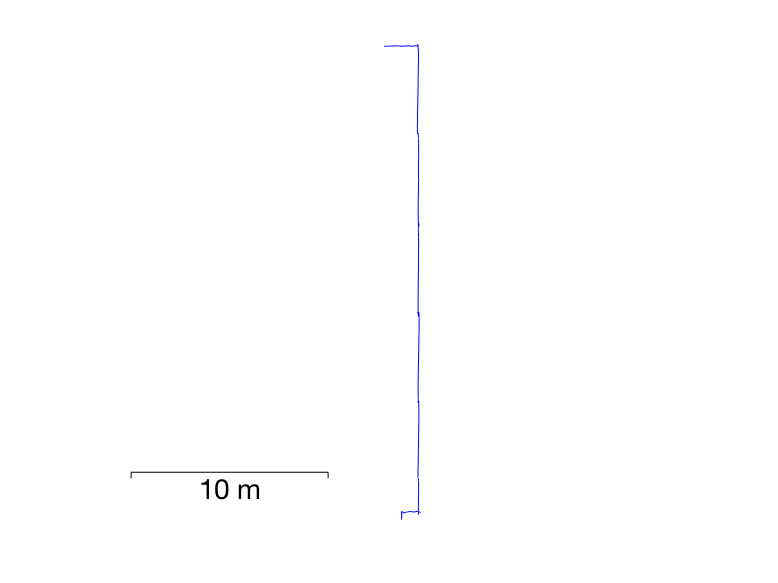} & Number: 7 \newline Length: 34.8 m \newline Duration: 1 min 36 s \newline Levels: 6 \newline Fix points: 23 \\ 
\includegraphics[width=4cm]{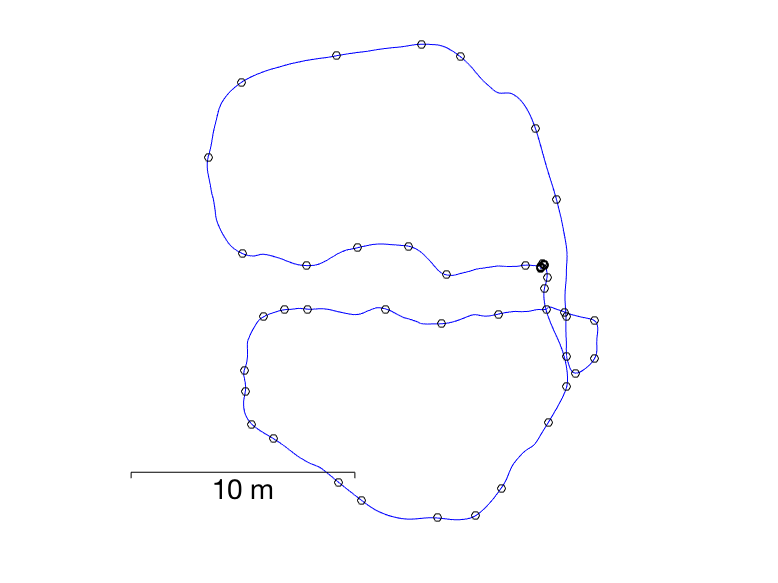} & \includegraphics[width=4cm]{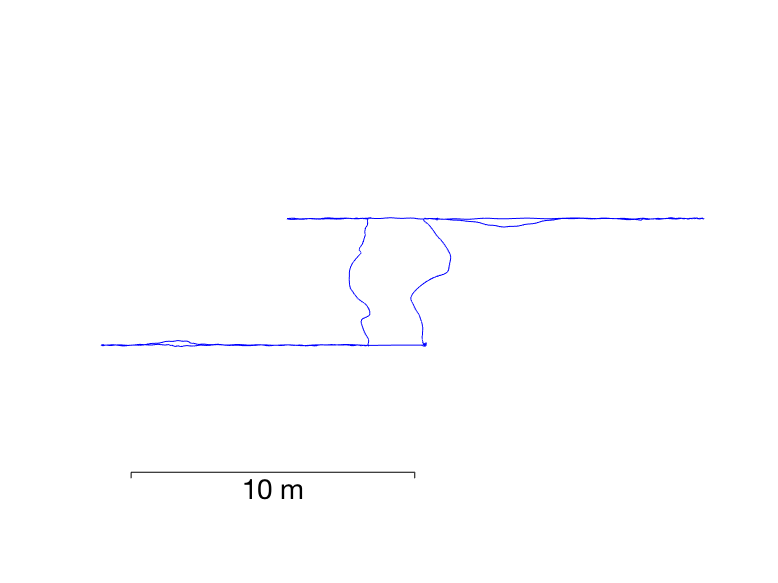} & Number: 8 \newline Length: 99.7 m \newline Duration: 1 min 47 s \newline Levels: 2 \newline Fix points: 49 \\ 
\includegraphics[width=4cm]{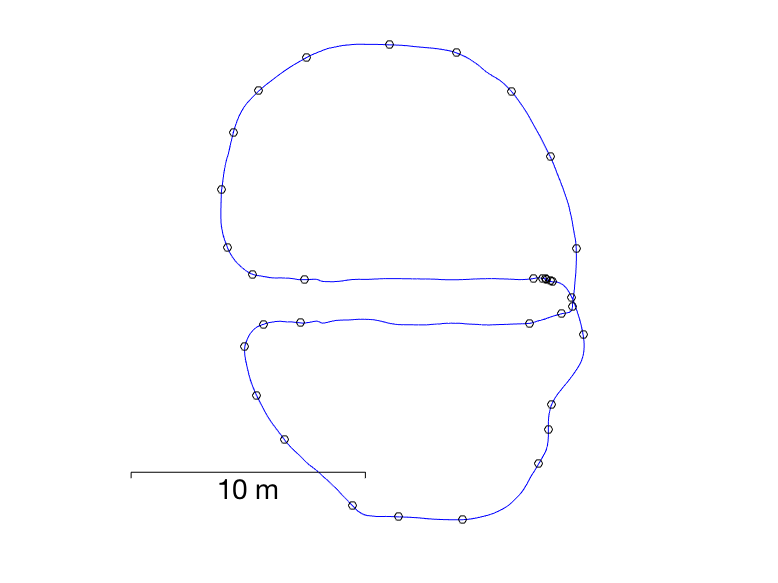} & \includegraphics[width=4cm]{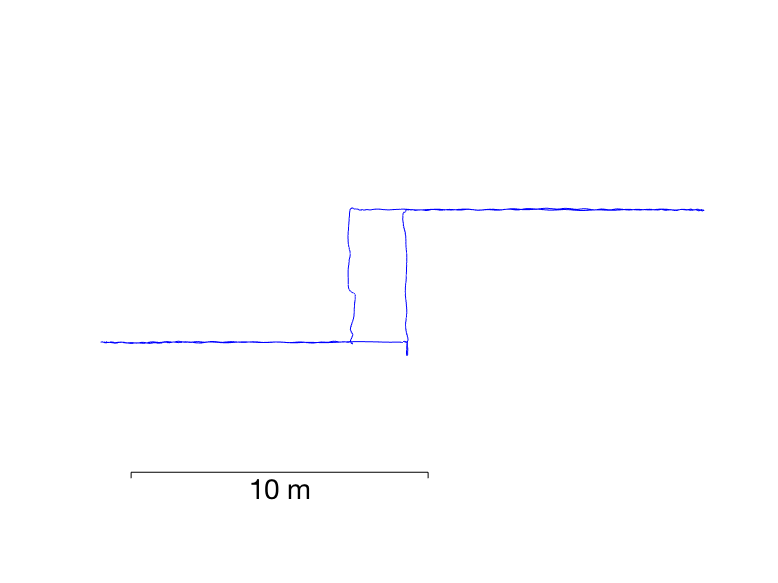} & Number: 9 \newline Length: 88.4 m \newline Duration: 1 min 37 s \newline Levels: 2 \newline Fix points: 36 \\ 
\includegraphics[width=4cm]{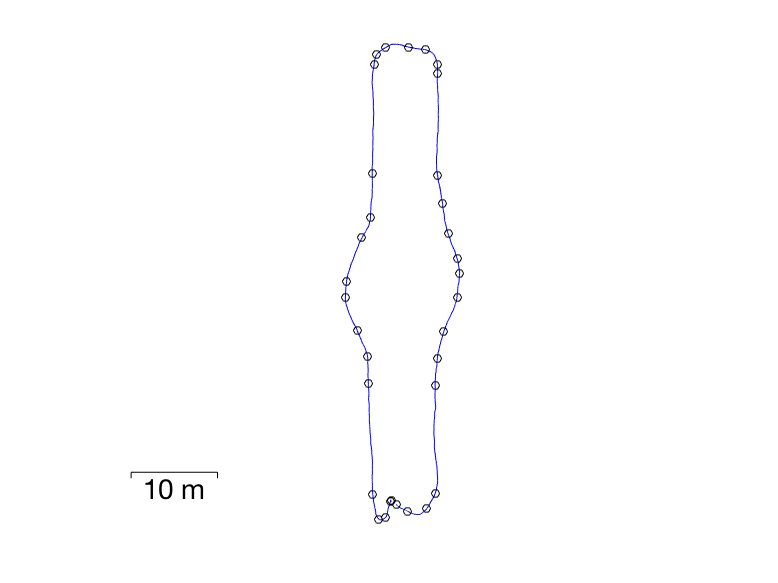} & \includegraphics[width=4cm]{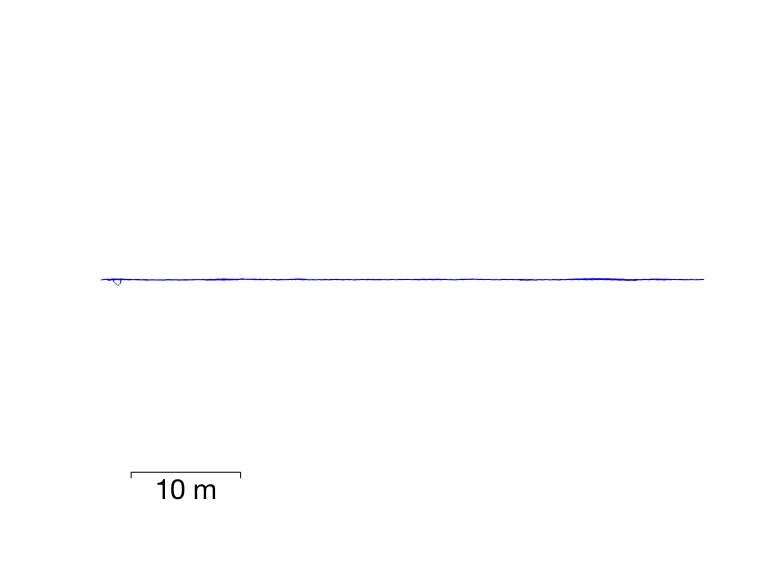} & Number: 10 \newline Length: 124.9 m \newline Duration: 2 min 3 s \newline Levels: 1 \newline Fix points: 33 \\ 
\includegraphics[width=4cm]{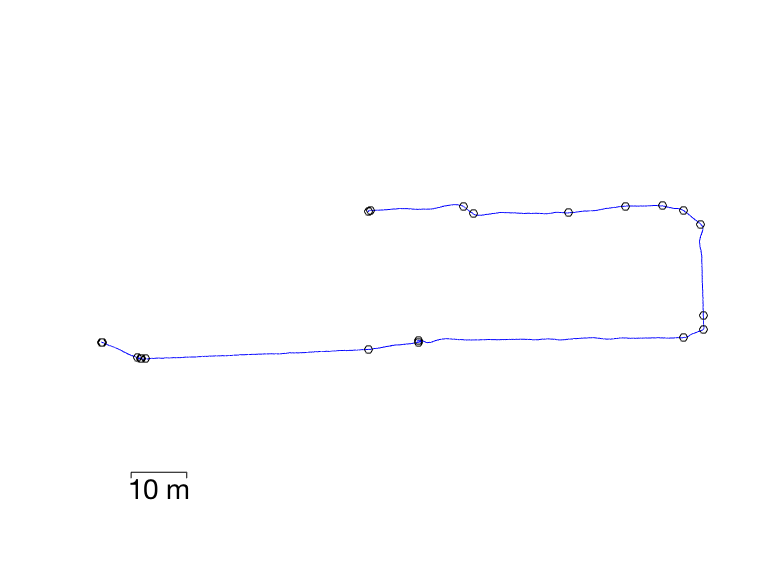} & \includegraphics[width=4cm]{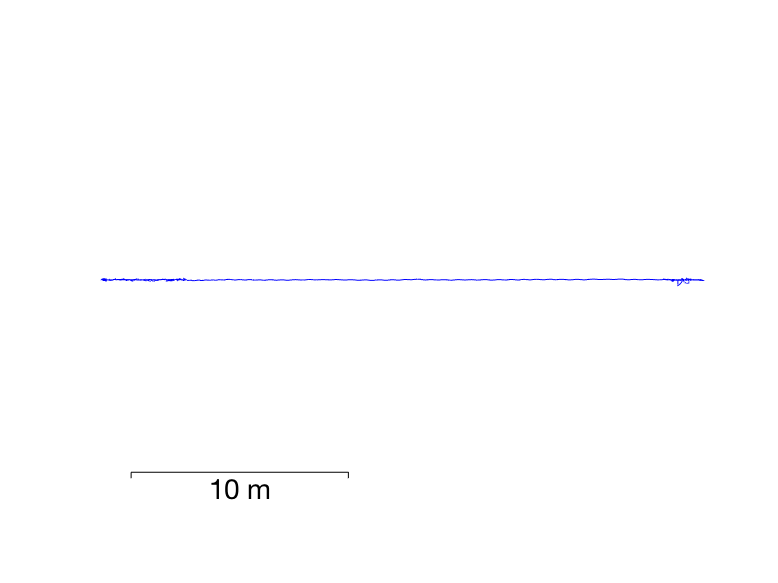} & Number: 11 \newline Length: 196.5 m \newline Duration: 3 min 27 s \newline Levels: 1 \newline Fix points: 21 \\ 
\includegraphics[width=4cm]{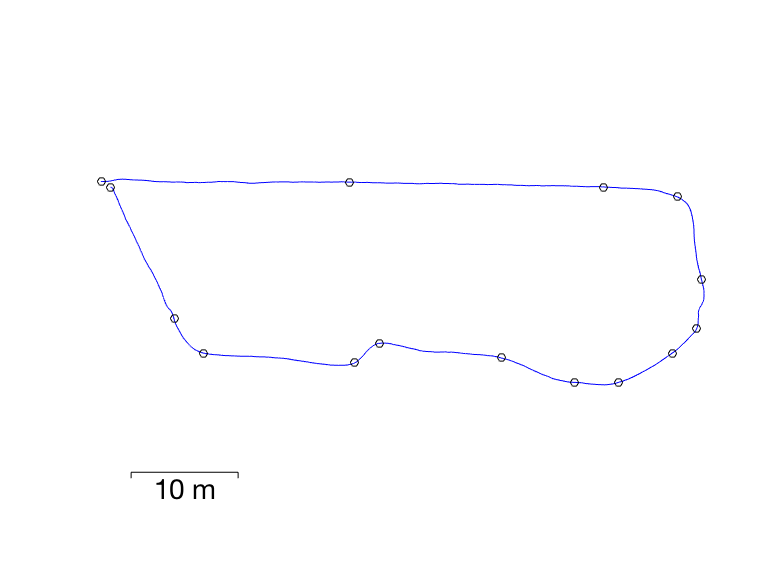} & \includegraphics[width=4cm]{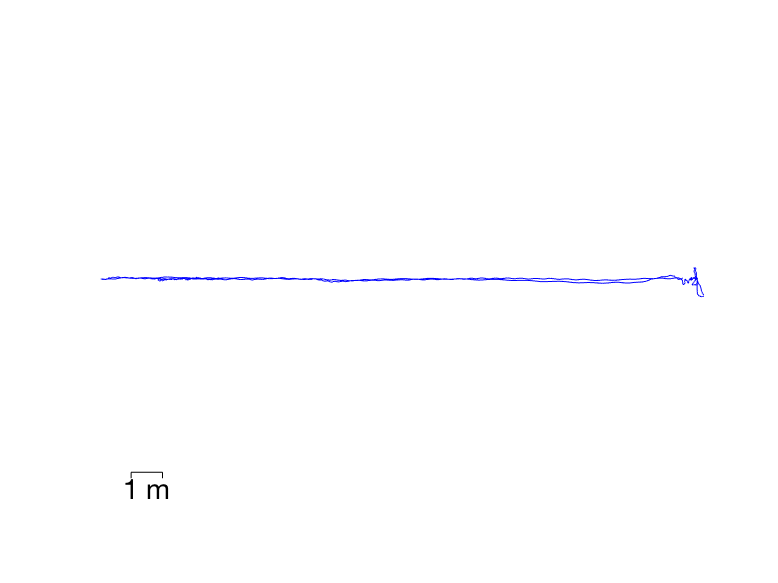} & Number: 12 \newline Length: 135.5 m \newline Duration: 1 min 55 s \newline Levels: 1 \newline Fix points: 15 \\ 
\includegraphics[width=4cm]{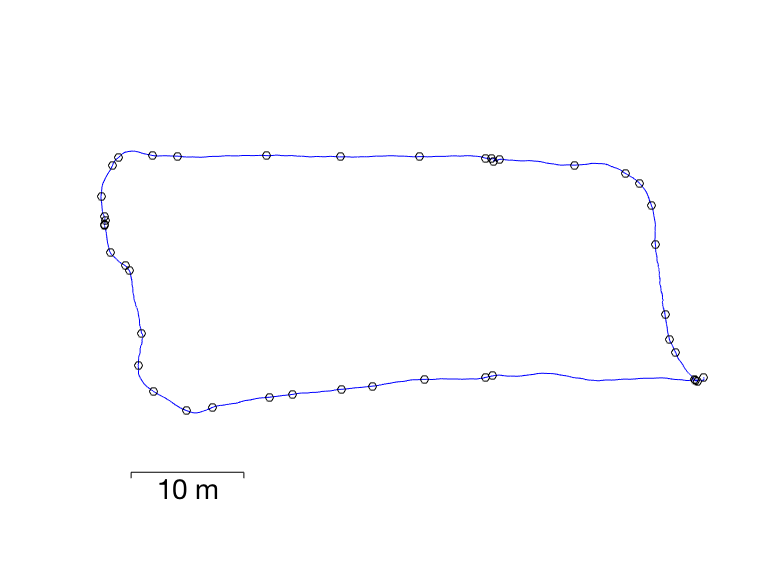} & \includegraphics[width=4cm]{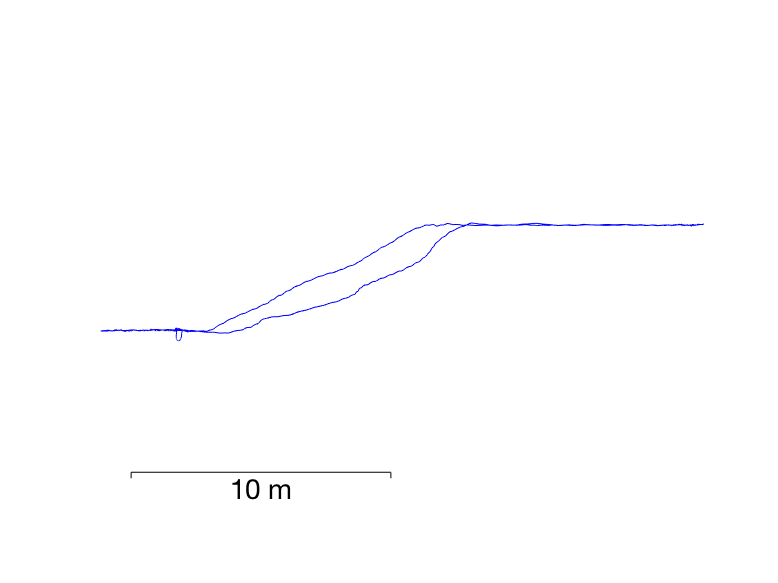} & Number: 13 \newline Length: 142.7 m \newline Duration: 2 min 32 s \newline Levels: 2 \newline Fix points: 44 \\ 
\includegraphics[width=4cm]{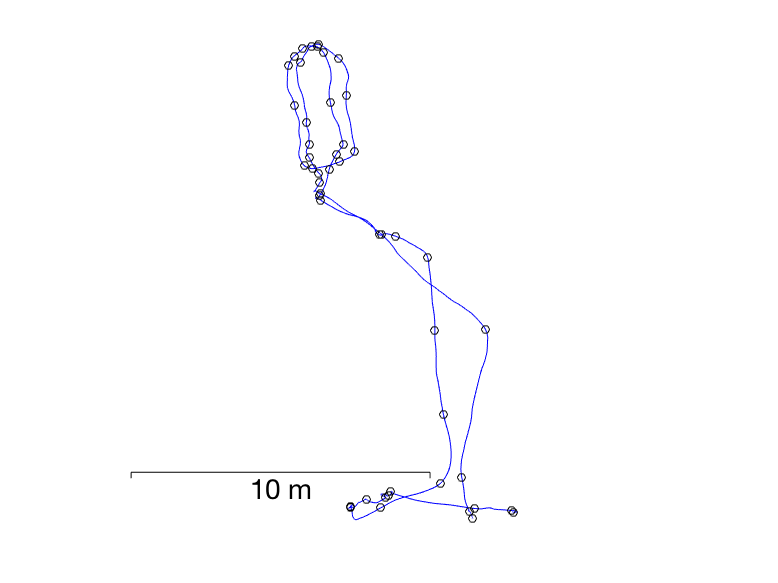} & \includegraphics[width=4cm]{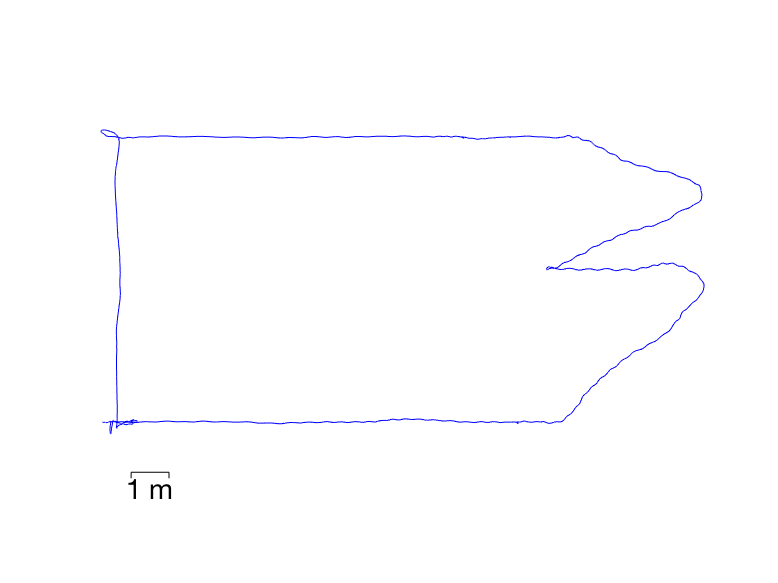} & Number: 14 \newline Length: 69.7 m \newline Duration: 1 min 53 s \newline Levels: 3 \newline Fix points: 48 \\ 
\includegraphics[width=4cm]{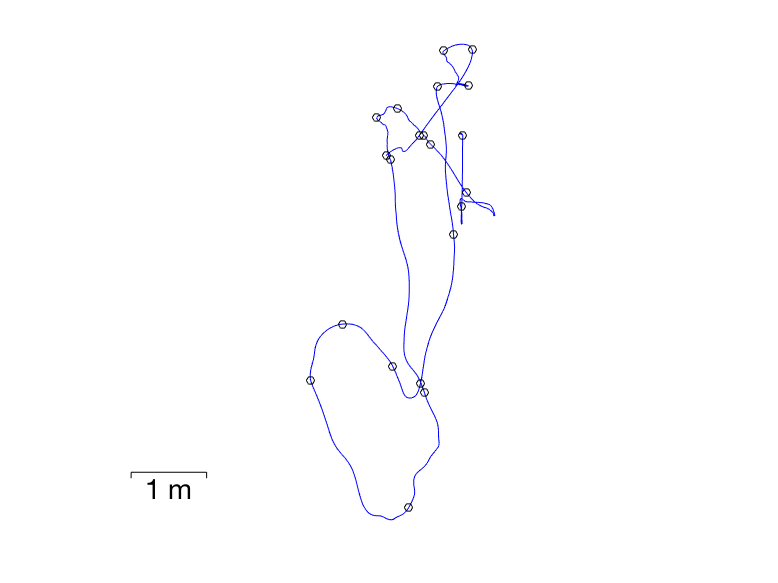} & \includegraphics[width=4cm]{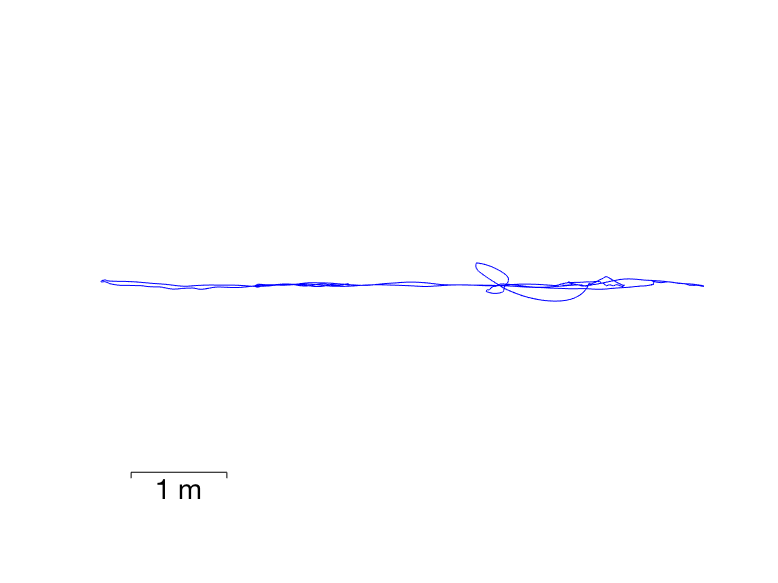} & Number: 15 \newline Length: 23.2 m \newline Duration: 0 min 52 s \newline Levels: 1 \newline Fix points: 21 \\ 
\includegraphics[width=4cm]{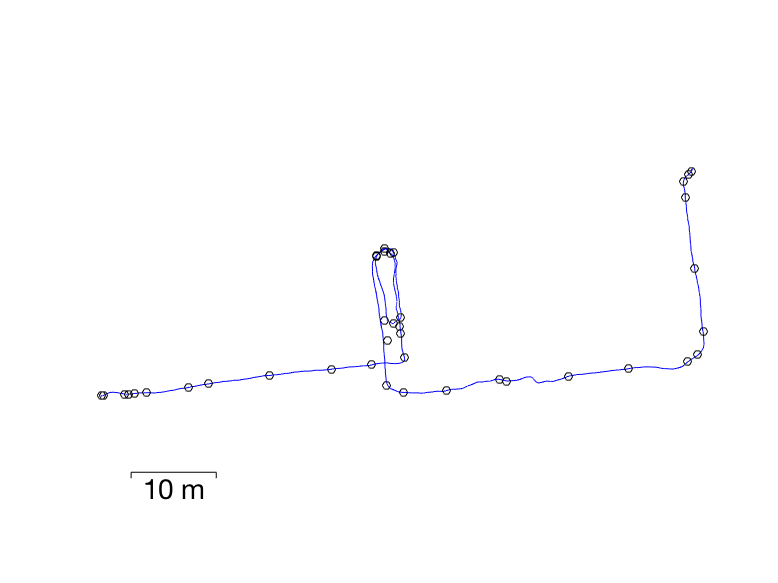} & \includegraphics[width=4cm]{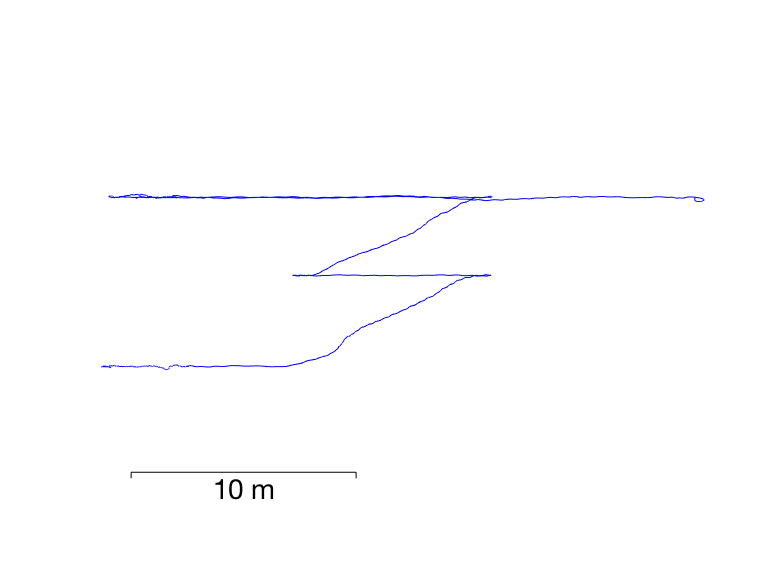} & Number: 16 \newline Length: 151.3 m \newline Duration: 2 min 27 s \newline Levels: 3 \newline Fix points: 40 \\ 
\includegraphics[width=4cm]{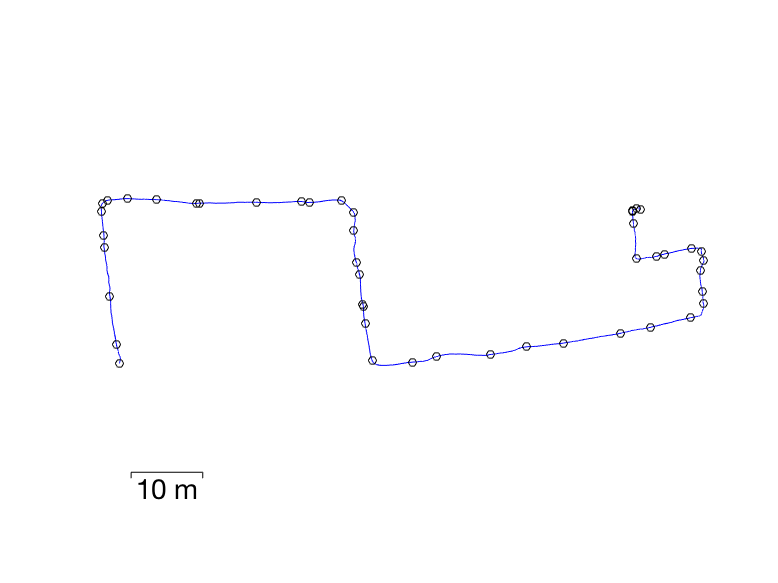} & \includegraphics[width=4cm]{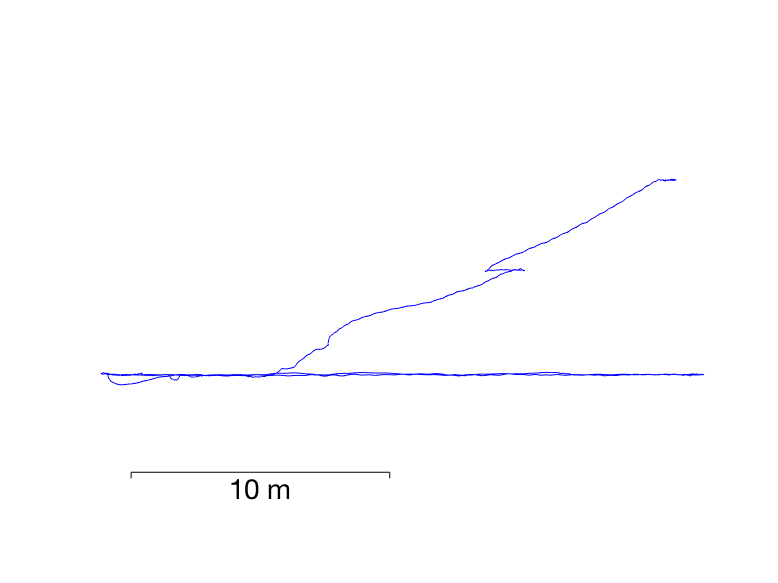} & Number: 17 \newline Length: 157.6 m \newline Duration: 2 min 38 s \newline Levels: 3 \newline Fix points: 46 \\ 
\includegraphics[width=4cm]{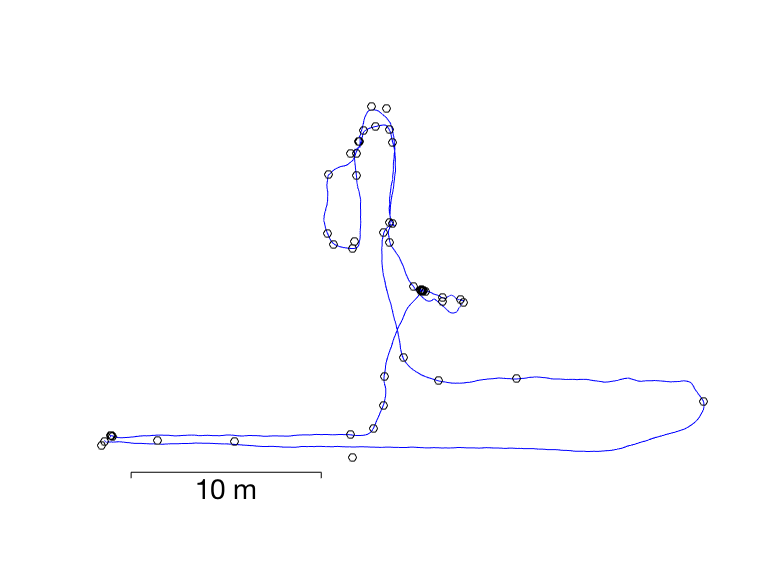} & \includegraphics[width=4cm]{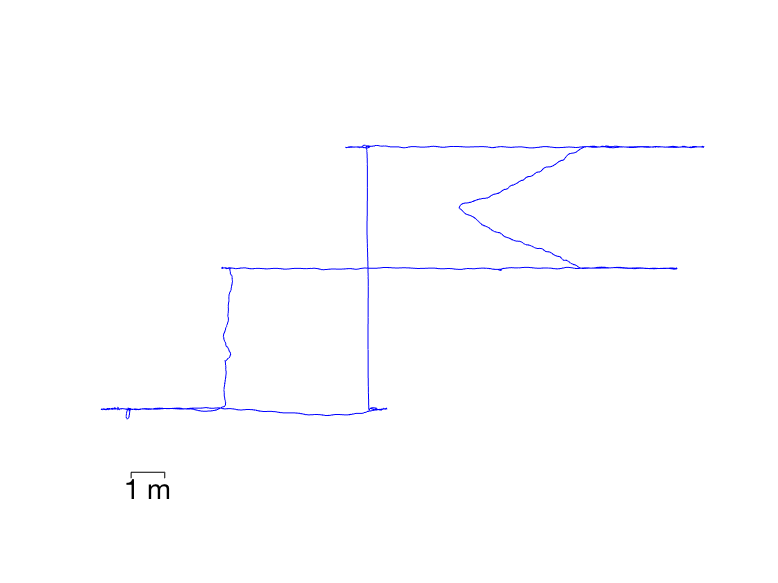} & Number: 18 \newline Length: 130.6 m \newline Duration: 3 min 17 s \newline Levels: 3 \newline Fix points: 52 \\ 
\includegraphics[width=4cm]{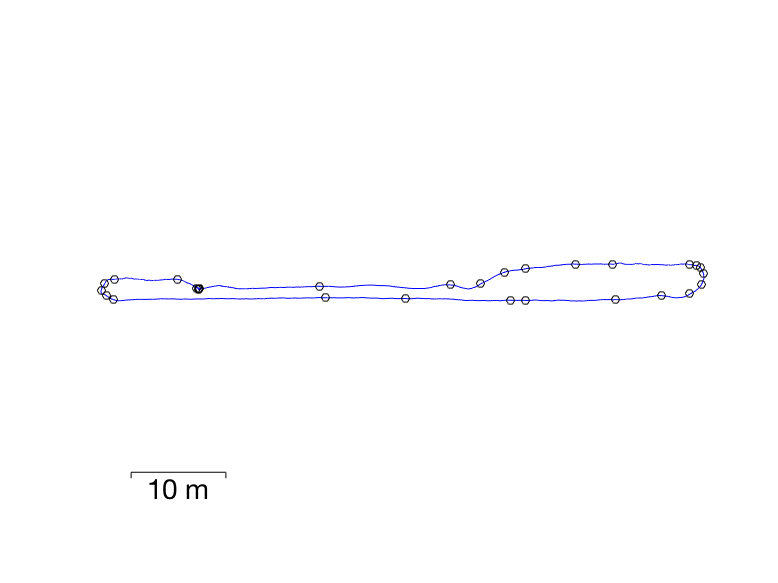} & \includegraphics[width=4cm]{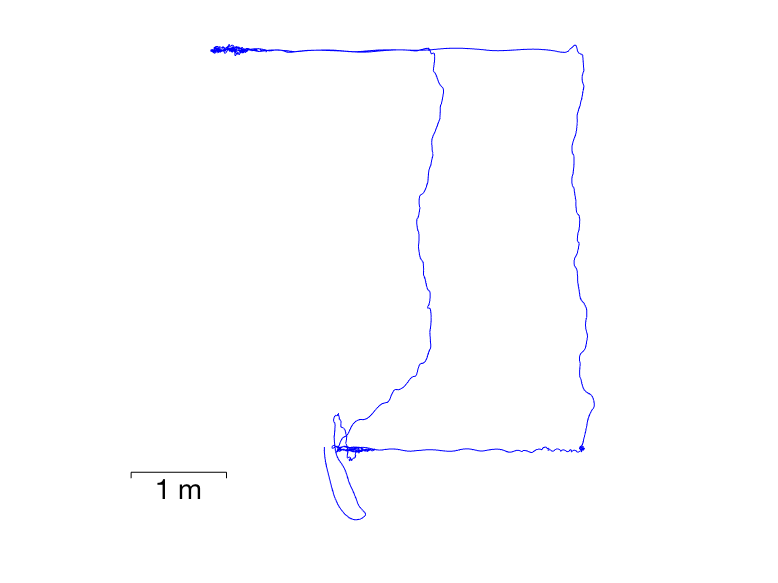} & Number: 19 \newline Length: 136.0 m \newline Duration: 2 min 22 s \newline Levels: 2 \newline Fix points: 29 \\ 
\includegraphics[width=4cm]{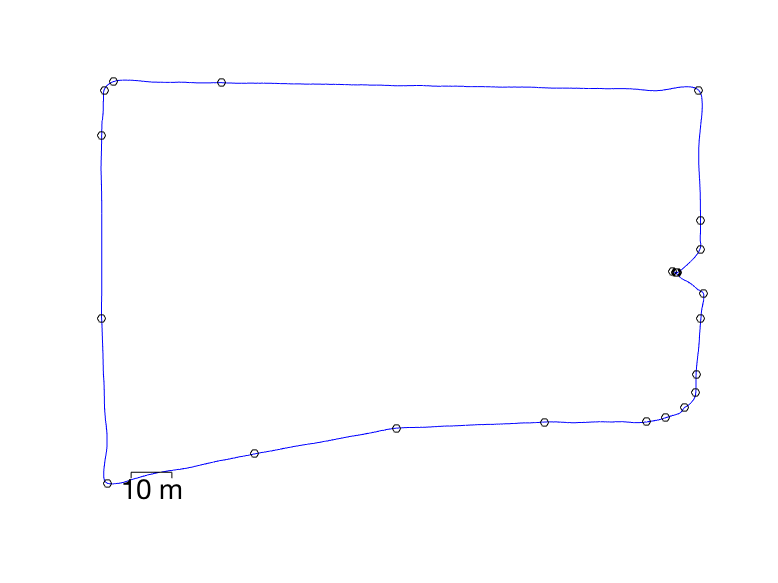} & \includegraphics[width=4cm]{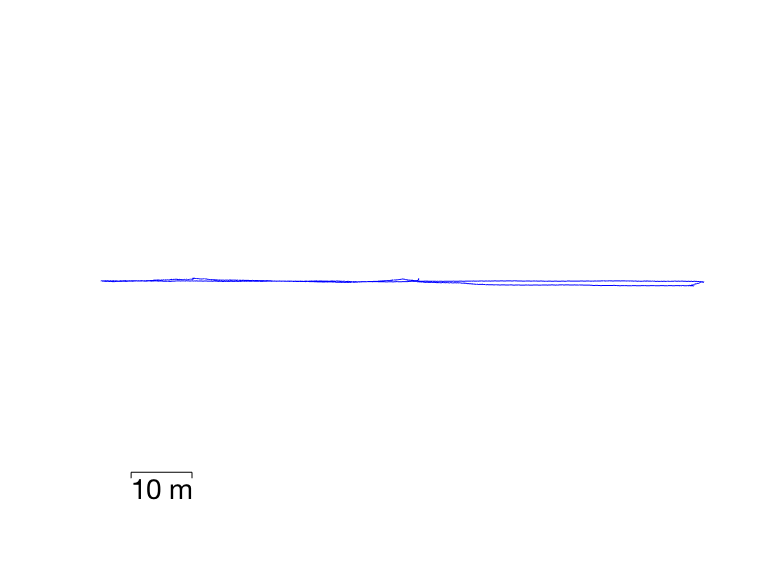} & Number: 20 \newline Length: 476.5 m \newline Duration: 5 min 3 s \newline Levels: 1 \newline Fix points: 23 \\ 
\includegraphics[width=4cm]{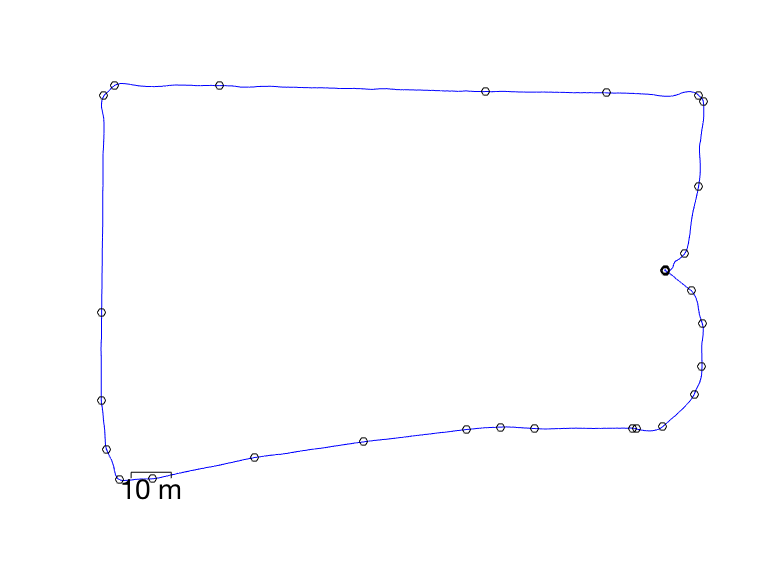} & \includegraphics[width=4cm]{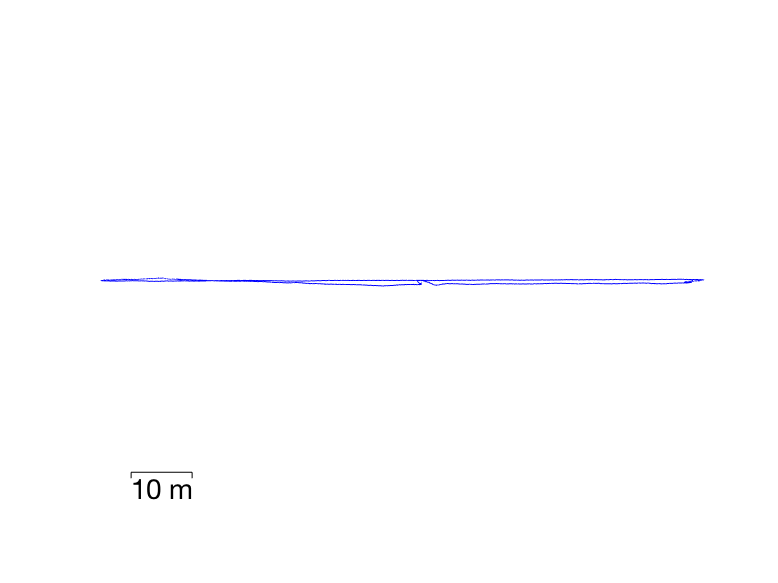} & Number: 21 \newline Length: 480.7 m \newline Duration: 5 min 17 s \newline Levels: 1 \newline Fix points: 32 \\ 
\includegraphics[width=4cm]{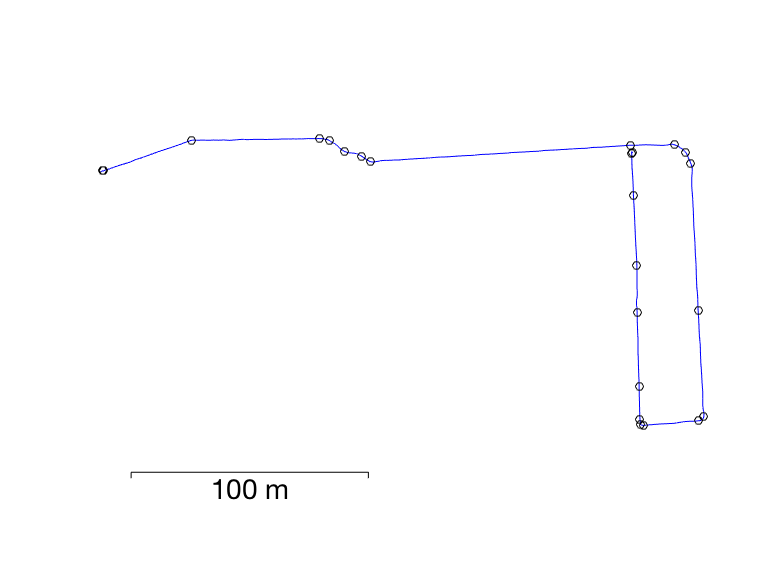} & \includegraphics[width=4cm]{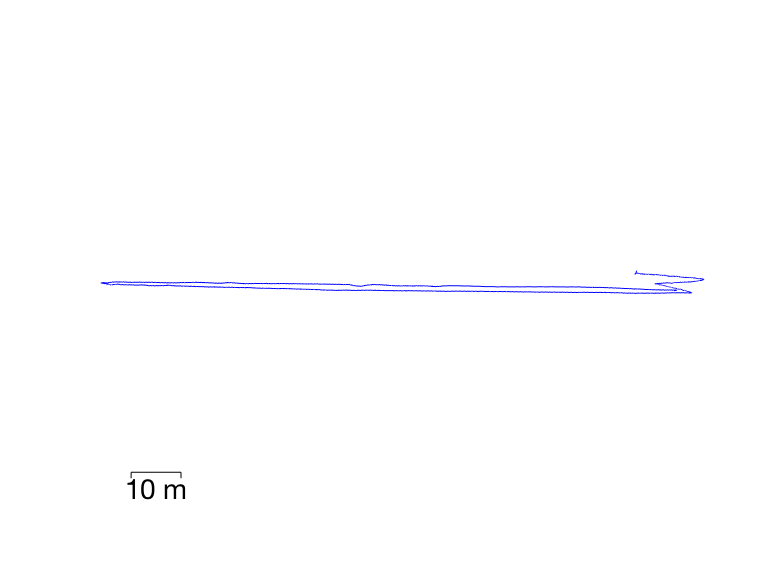} & Number: 22 \newline Length: 514.0 m \newline Duration: 6 min 26 s \newline Levels: 1 \newline Fix points: 24 \\ 
\includegraphics[width=4cm]{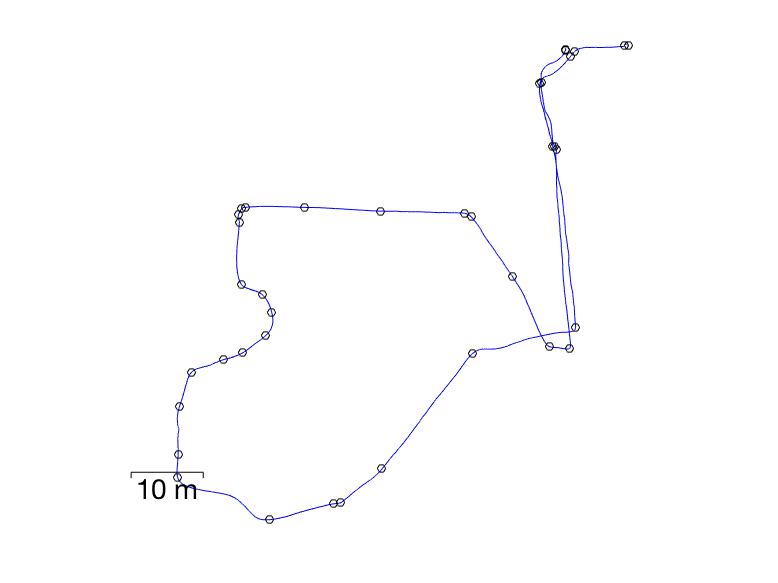} & \includegraphics[width=4cm]{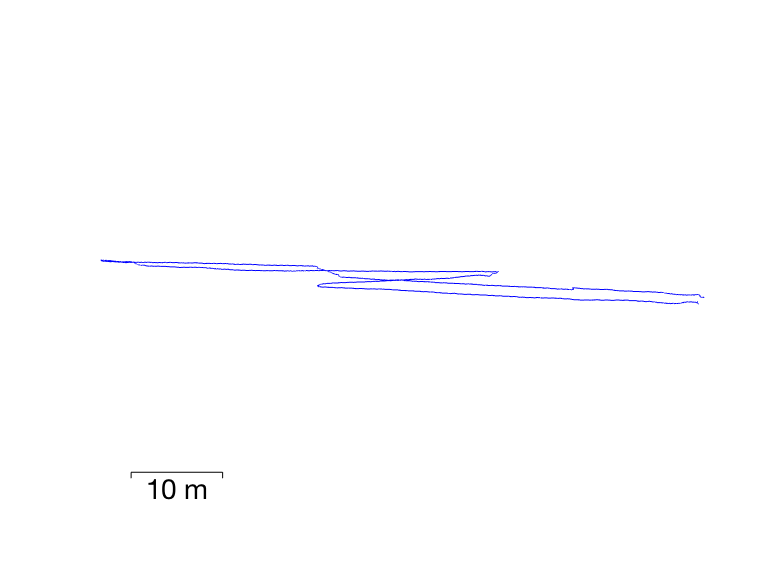} & Number: 23 \newline Length: 266.4 m \newline Duration: 3 min 26 s \newline Levels: 1 \newline Fix points: 38 \\ 
\end{longtable}

\clearpage

\section{Speed histograms per data set}

\noindent
\begin{minipage}{0.23\textwidth}
  \centering\footnotesize
  \includegraphics[width=\textwidth]{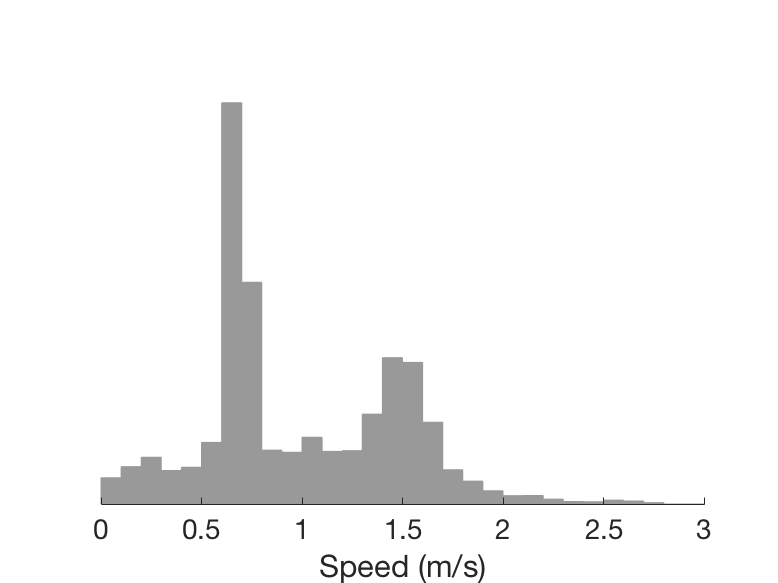}
  {1}
\end{minipage}
\hspace*{\fill}
\begin{minipage}{0.23\textwidth}
  \centering\footnotesize
  \includegraphics[width=\textwidth]{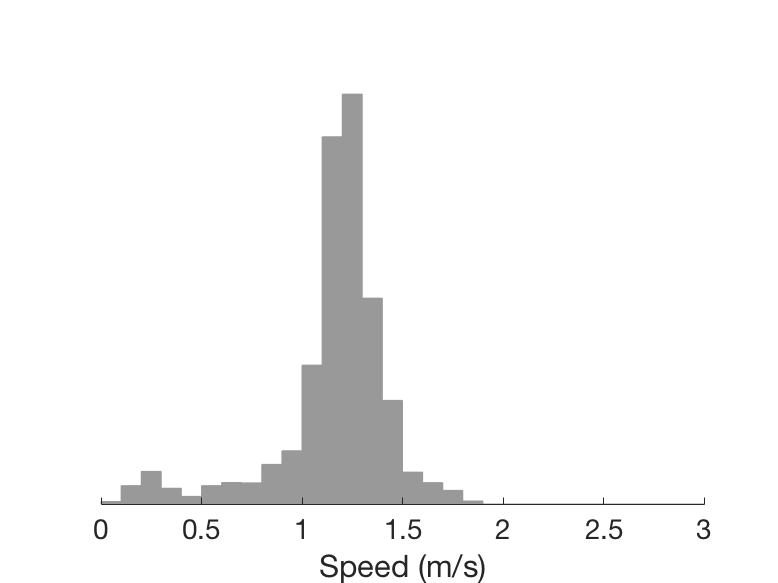}
  {2}
\end{minipage}
\hspace*{\fill}
\begin{minipage}{0.23\textwidth}
  \centering\footnotesize
  \includegraphics[width=\textwidth]{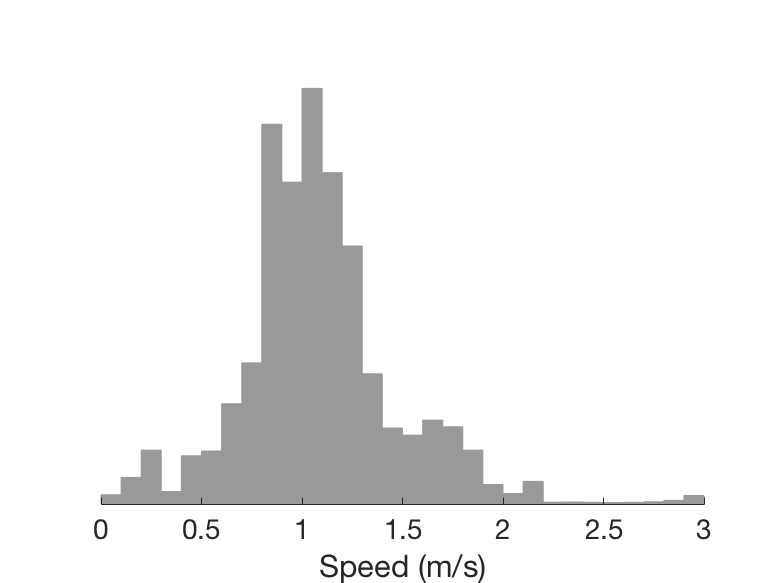}
  {3}
\end{minipage}
\hspace*{\fill}
\begin{minipage}{0.23\textwidth}
  \centering\footnotesize
  \includegraphics[width=\textwidth]{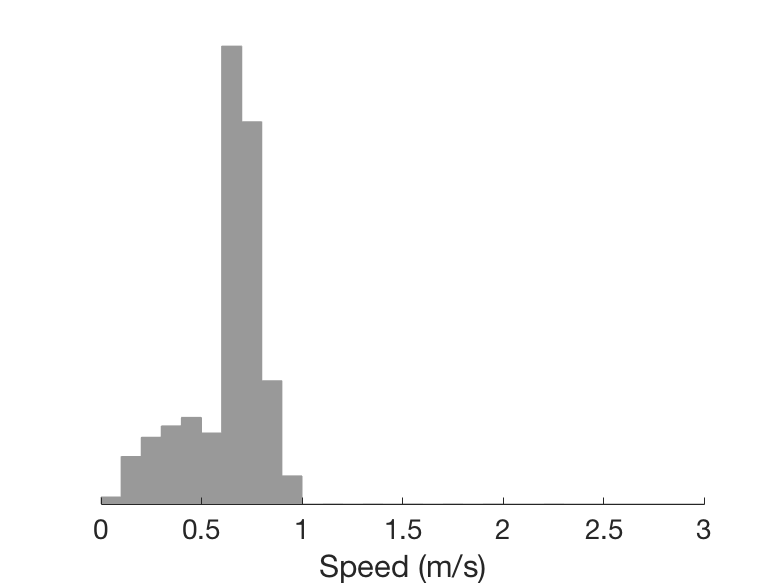}
  {4}
\end{minipage}

\noindent
\begin{minipage}{0.23\textwidth}
  \centering\footnotesize
  \includegraphics[width=\textwidth]{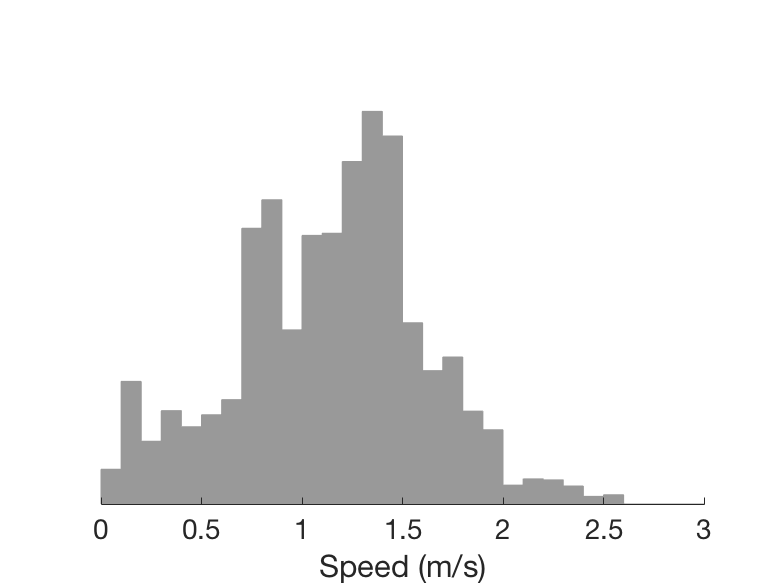}
  {5}
\end{minipage}
\hspace*{\fill}
\begin{minipage}{0.23\textwidth}
  \centering\footnotesize
  \includegraphics[width=\textwidth]{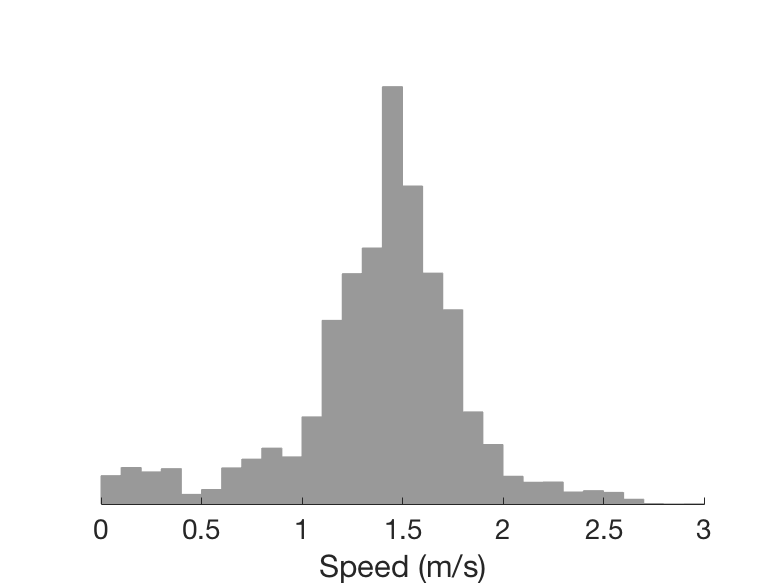}
  {6}
\end{minipage}
\hspace*{\fill}
\begin{minipage}{0.23\textwidth}
  \centering\footnotesize
  \includegraphics[width=\textwidth]{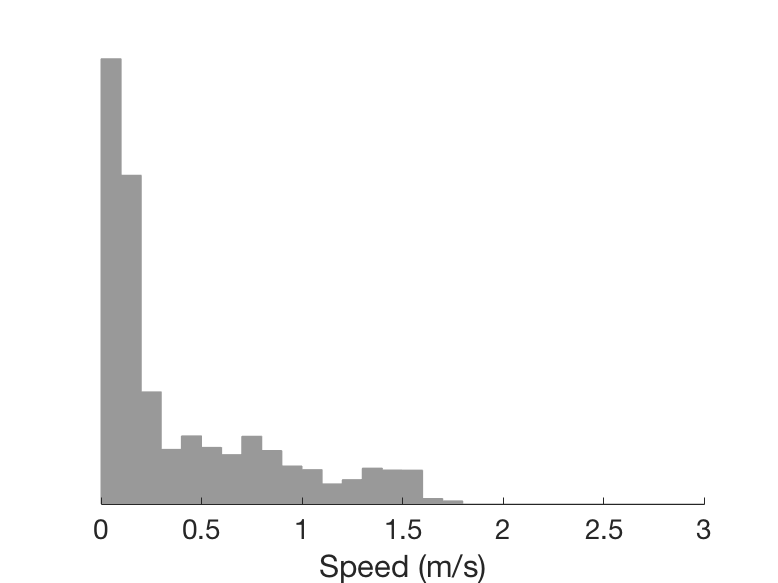}
  {7}
\end{minipage}
\hspace*{\fill}
\begin{minipage}{0.23\textwidth}
  \centering\footnotesize
  \includegraphics[width=\textwidth]{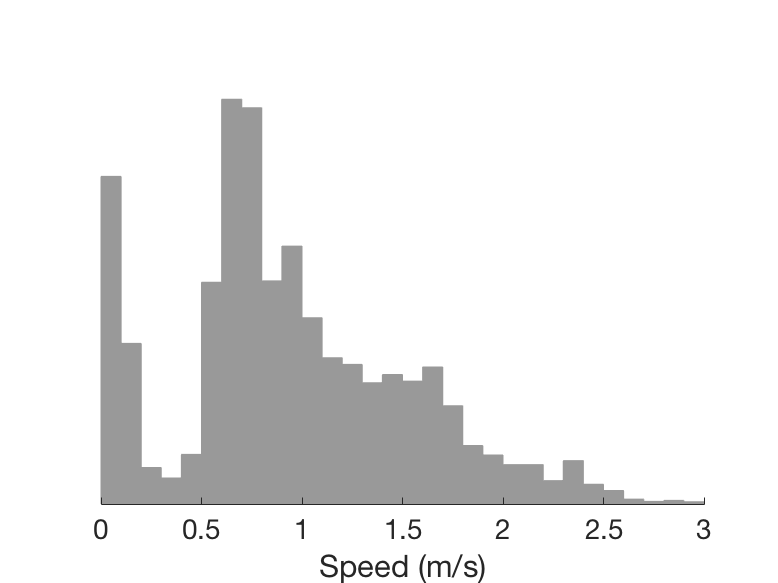}
  {8}
\end{minipage}

\noindent
\begin{minipage}{0.23\textwidth}
  \centering\footnotesize
  \includegraphics[width=\textwidth]{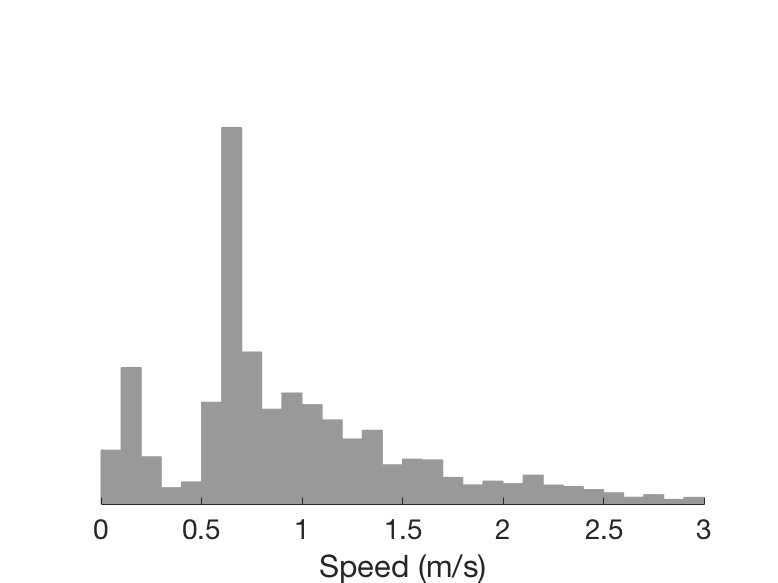}
  {9}
\end{minipage}
\hspace*{\fill}
\begin{minipage}{0.23\textwidth}
  \centering\footnotesize
  \includegraphics[width=\textwidth]{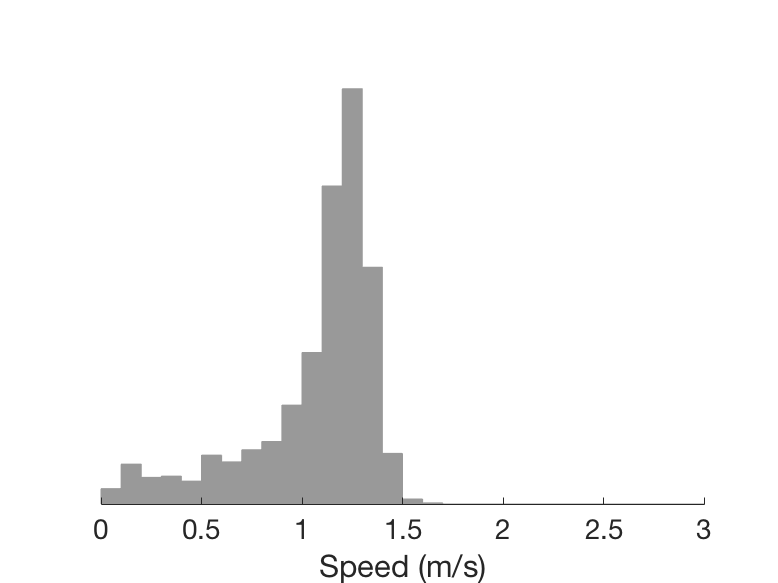}
  {10}
\end{minipage}
\hspace*{\fill}
\begin{minipage}{0.23\textwidth}
  \centering\footnotesize
  \includegraphics[width=\textwidth]{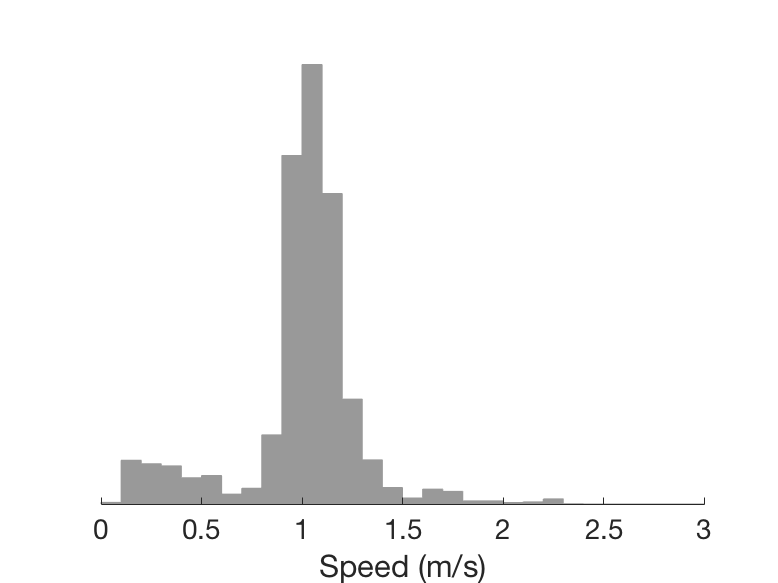}
  {11}
\end{minipage}
\hspace*{\fill}
\begin{minipage}{0.23\textwidth}
  \centering\footnotesize
  \includegraphics[width=\textwidth]{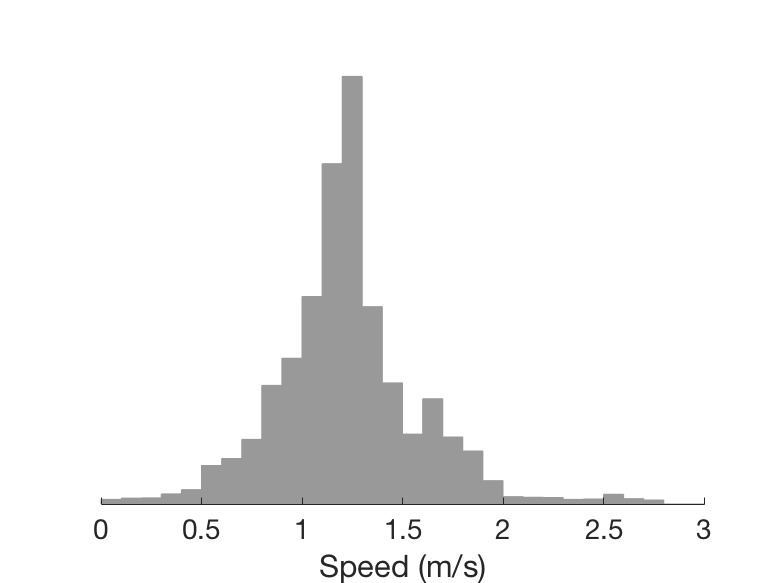}
  {12}
\end{minipage}

\noindent
\begin{minipage}{0.23\textwidth}
  \centering\footnotesize
  \includegraphics[width=\textwidth]{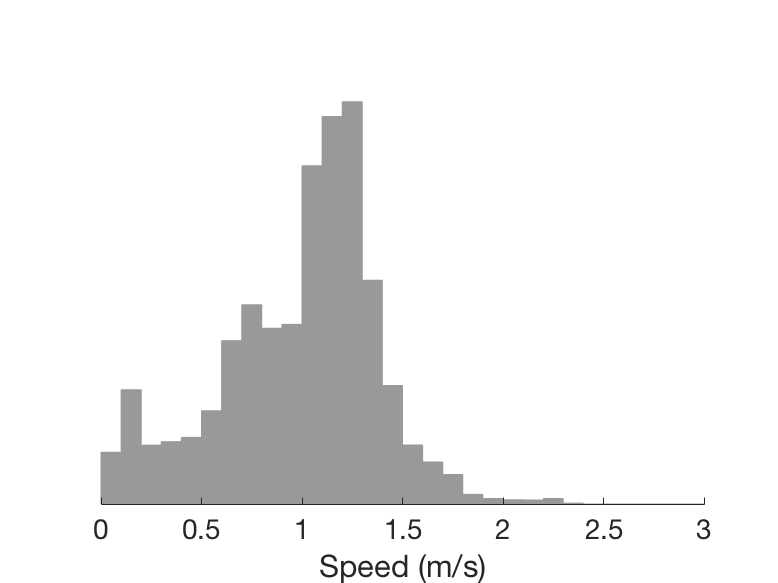}
  {13}
\end{minipage}
\hspace*{\fill}
\begin{minipage}{0.23\textwidth}
  \centering\footnotesize
  \includegraphics[width=\textwidth]{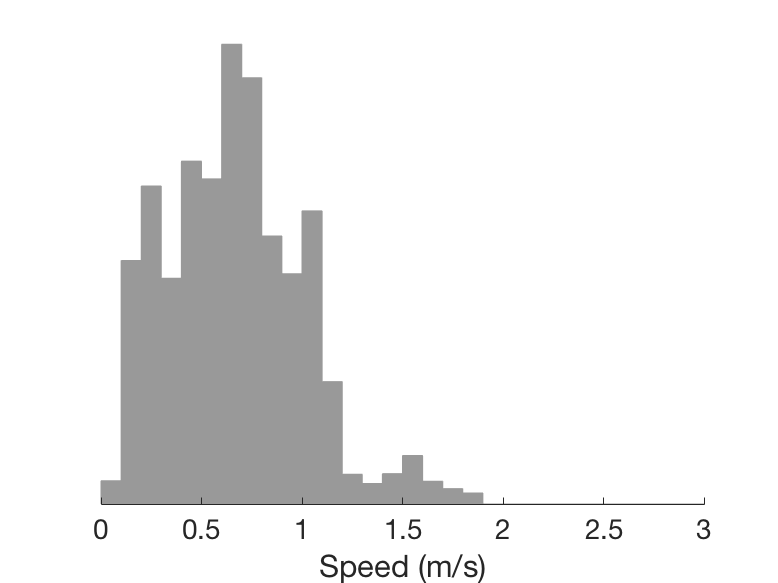}
  {14}
\end{minipage}
\hspace*{\fill}
\begin{minipage}{0.23\textwidth}
  \centering\footnotesize
  \includegraphics[width=\textwidth]{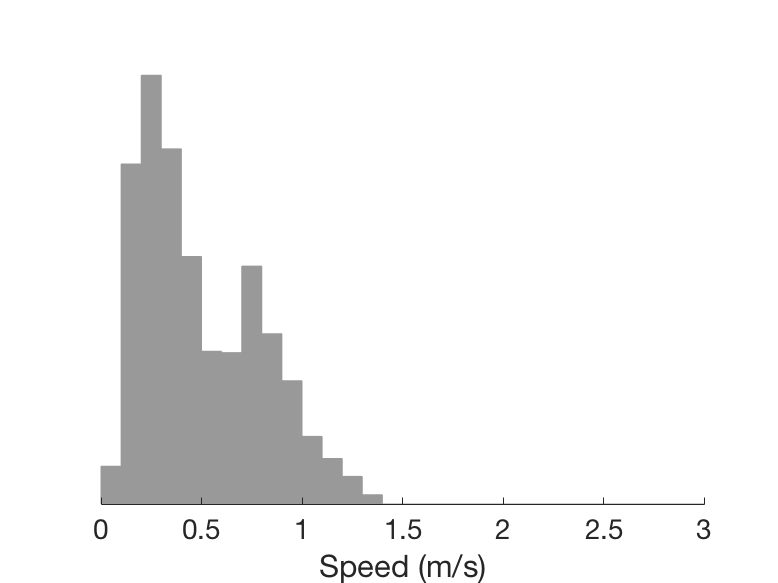}
  {15}
\end{minipage}
\hspace*{\fill}
\begin{minipage}{0.23\textwidth}
  \centering\footnotesize
  \includegraphics[width=\textwidth]{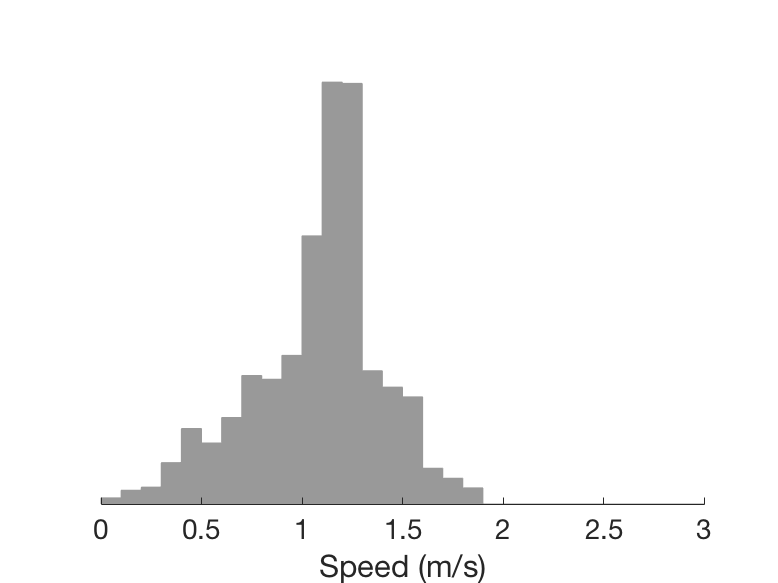}
  {16}
\end{minipage}

\noindent
\begin{minipage}{0.23\textwidth}
  \centering\footnotesize
  \includegraphics[width=\textwidth]{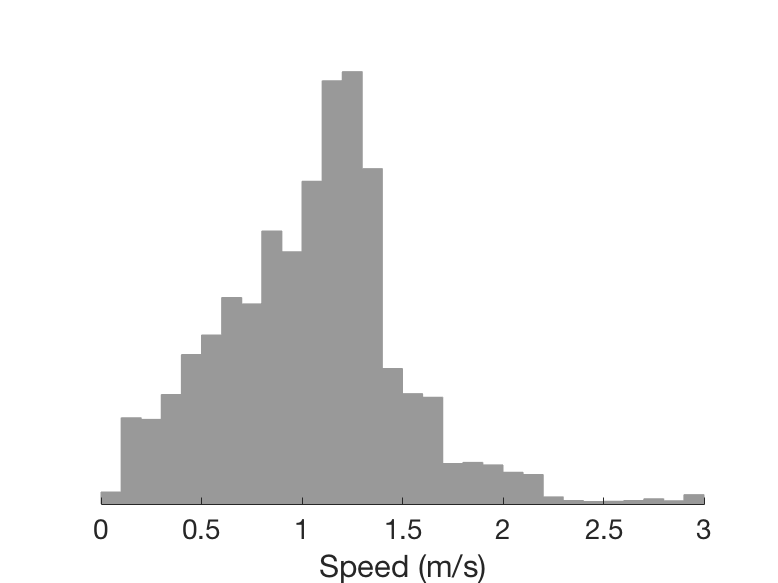}
  {17}
\end{minipage}
\hspace*{\fill}
\begin{minipage}{0.23\textwidth}
  \centering\footnotesize
  \includegraphics[width=\textwidth]{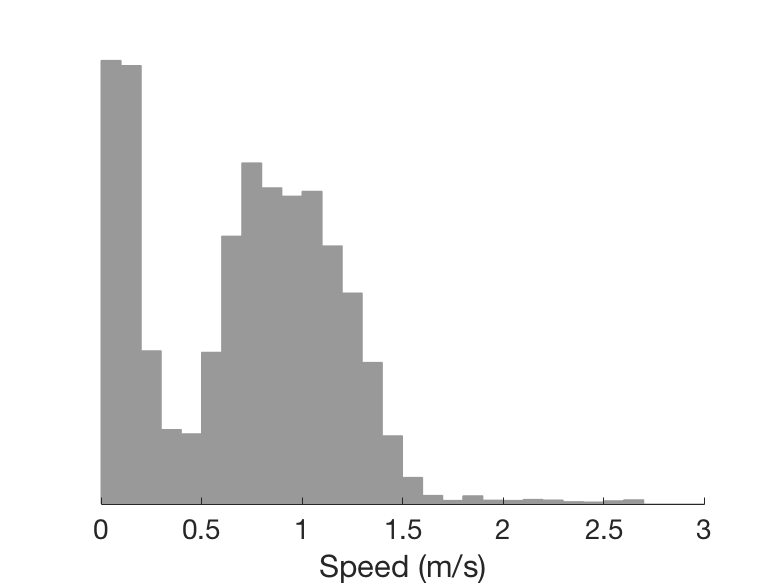}
  {18}
\end{minipage}
\hspace*{\fill}
\begin{minipage}{0.23\textwidth}
  \centering\footnotesize
  \includegraphics[width=\textwidth]{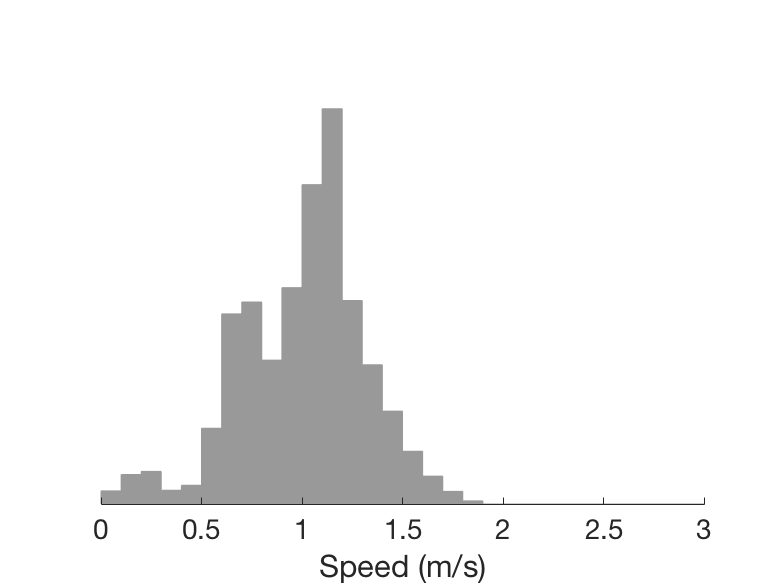}
  {19}
\end{minipage}
\hspace*{\fill}
\begin{minipage}{0.23\textwidth}
  \centering\footnotesize
  \includegraphics[width=\textwidth]{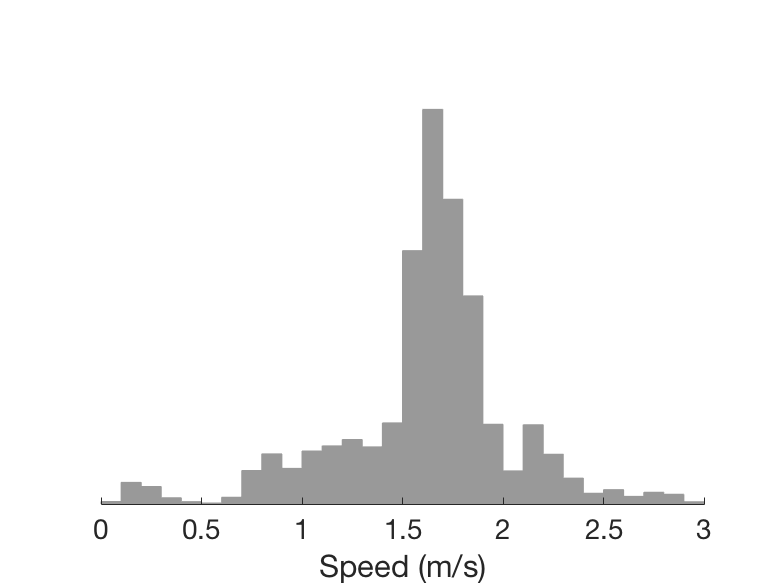}
  {20}
\end{minipage}

\noindent
\begin{minipage}{0.23\textwidth}
  \centering\footnotesize
  \includegraphics[width=\textwidth]{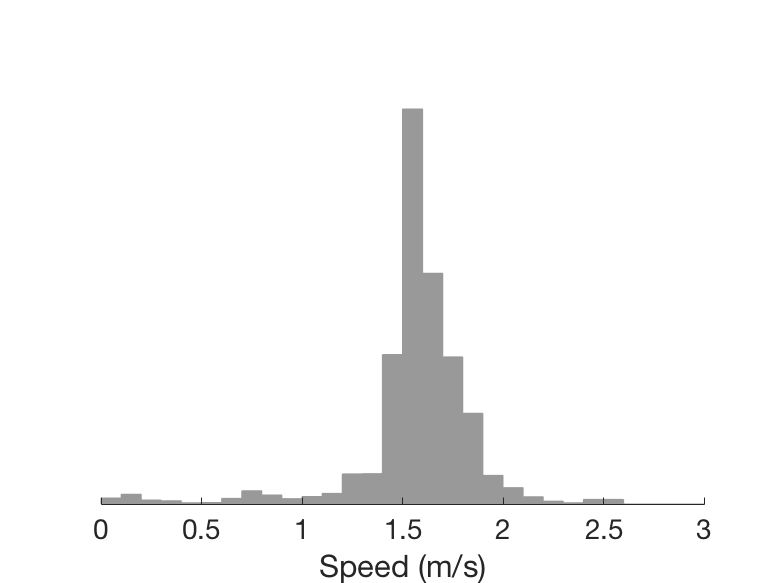}
  {21}
\end{minipage}
\hspace*{\fill}
\begin{minipage}{0.23\textwidth}
  \centering\footnotesize
  \includegraphics[width=\textwidth]{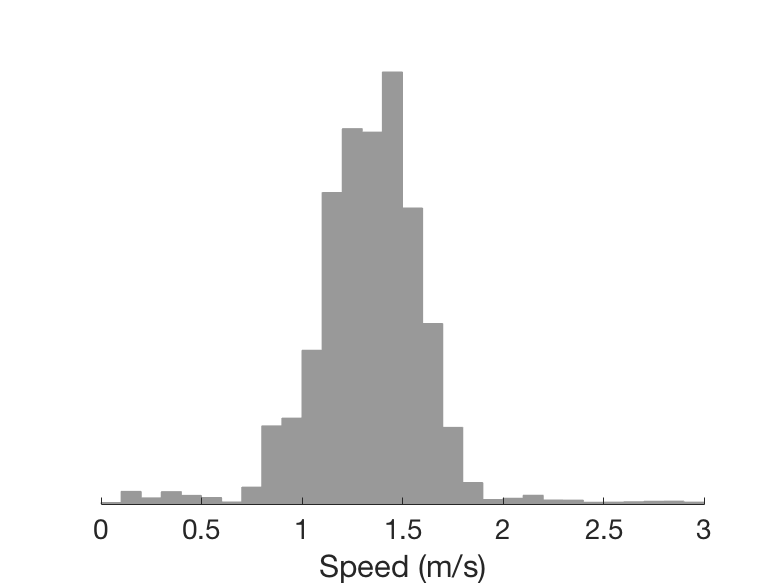}
  {22}
\end{minipage}
\hspace*{\fill}
\begin{minipage}{0.23\textwidth}
  \centering\footnotesize
  \includegraphics[width=\textwidth]{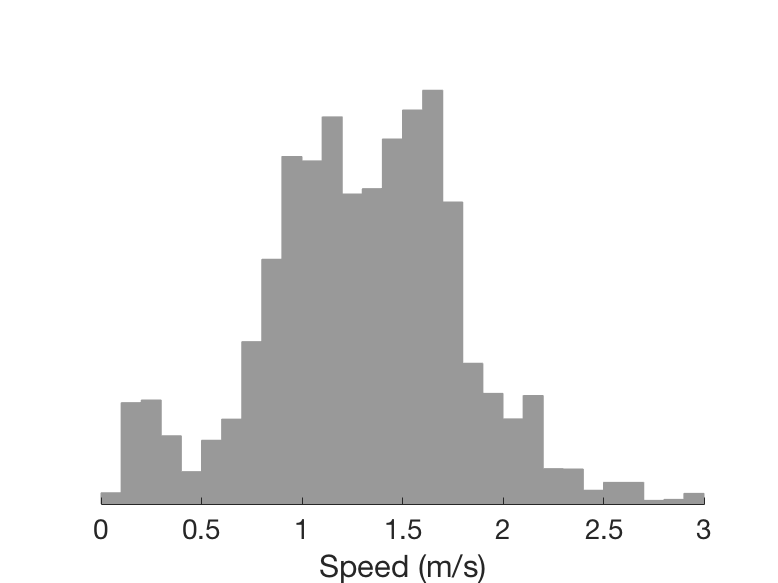}
  {23}
\end{minipage}
\hspace*{\fill}
\begin{minipage}{0.23\textwidth}
  \centering\footnotesize

  {~}
\end{minipage}

  \label{LastPage}

\end{document}